TEXT2INSIGHT: TRANSFORM NATURAL LANGUAGE TEXT INTO INSIGHTS SEAMLESSLY USING MULTI MODEL ARCHITECTURE

PRADEEP SAIN

Final Thesis Report

JUNE 2024

# DEDICATION

This thesis is dedicated to my loving family, whose unwavering support and encouragement have been my guiding light throughout this journey. Thanks to my mom for always praying for my well-being. And thanks to my dad for always believing in me and motivating me, your belief in my abilities has given me the strength to overcome every obstacle. To my wife Riya, thank you for your patience and understanding during the countless hours spent studying. I am grateful for your sacrifices and the love that binds us together. This accomplishment would not have been possible without each of you. Thank you for being my pillars of strength.

Finally, I would like to dedicate this thesis to future generations of researchers and scholars who strive to push the boundaries of knowledge and make a positive impact on the world. May this work contribute in some small way to the collective pursuit of truth and understanding.

# ACKNOWLEDGEMENTS

I would like to express my deepest gratitude to all those who have contributed to the successful completion of this research paper.

First and foremost, I would like to thank my advisor, Dr. Praveen D Chougale for his invaluable guidance, support, and encouragement throughout the research process. His insights and expertise were instrumental in shaping this work.

Special thanks go to Dr. Manoj J for patiently clarifying all the doubts during classroom sessions and providing detailed answers to queries. His live classroom sessions were instrumental in helping to complete this research.

I am also deeply grateful to Harshita Khanna, my student mentor, for her continuous support. Additionally, I would like to thank Dr. Harika Vuppala for her thoughtful feedback and constructive criticism, which greatly enhanced the quality of this research.

Lastly, I would like to thank UpGrad and all the co-learners who were always there to resolve my queries.

This research would not have been possible without the collective efforts of all these individuals, and to them, I am deeply indebted.

# ABSTRACT

The increasing need for dynamic and user-centric data analysis and visualization solutions is evident in various domain including healthcare, finance, research and other. Traditional data visualization systems, while valuable, often fall short in meeting user expectations as they are static and predefined by the system, not aligning with the unique requirements of individual users.

This research introduces Text2Insight, a novel solution designed to address this limitation by delivering data analysis and visualizations tailored to users' specific needs. The developed approach employs a multi-model architecture that provides data analysis and visualization from natural language text, specifying the user's analysis requirements.

The methodology involves an initial data analysis of the input file to extract relevant information such as shape, columns, and related values. Subsequently, a pre-trained Llama3 model is employed to convert the user-provided natural language text into an SQL query. This generated query is refined with the provided input data to formulate an accurate SQL query using NER model, ensuring accurate results. The methodology then utilizes a chart predictor method to determine the most fitting chart type based on the inputted text. Then, the methodology uses the Llama3 model to generate insights from the data resulting from the execution of the SQL query. Ultimately, the developed methodology generates visually informative charts as the output for user-friendly data visualization. Additionally, to provide comprehensive data analysis, the study developed a question-answering model and a predictive model. These models, utilizing the BERT model, can offer insights into past and future outcomes.

The performance of each inner model has assessed individually, followed by an overall evaluation of the complete model. The evaluation using accuracy, precision, recall, F1-score, syntactical evaluation, and BLEU score are found to be 99, 100, 99, 99, and 0.5 respectively for the Text2Insight model. For the question-answering mode, the accuracy is 89, precision 72, recall 69, and F1-score 69. For the predictive model, the accuracy is 70, precision 70, recall 68, and F1-score 68. These results determine the viability and effectiveness of Text2Insight as a promising solution for transforming natural language text into dynamic and user-specific data analysis and visualizations.

# TABLE OF CONTENTS

# LIST OF TABLES

# LIST OF FIGURES

# LIST OF ABBREVIATIONS

| | |
|---|---|
| AUC………………… | Area Under the Curve |
| BERT……………….  | Bidirectional Encoder Representations from Transformers |
| BLEU……………….  | Bilingual Evaluation Understudy |
| CNN………………… | Convolutional Neural Networks |
| CSV…………………. | Comma Separated Values |
| GPT…………………. | Generative Pre-trained Transformer |
| IPL…………………... | Indian Premier League |
| JSON………………... | JavaScript Object Notation |
| LLM………………… | Large Language Model |
| LSTM………………. | Long Short-Term Memory |
| ML4VIS……………. | Machine Learning for Visualization |
| NER…………………. | Named Entity Recognition |
| NLP………………… | Natural Language Processing |
| NL2VIS……………… | Natural Language to Visualization |
| NL4DV……………… | Natural Language for Data Visualization |
| OCR………………… | Optical Character Recognition |
| POS…………………. | Parts of Speech |
| QA…………………... | Question Answering |
| RDBMS…………….. | Relational Database Management System |
| RNN…………………. | Recurrent Neural Networks |
| ROC…………………. | Receiver Operating Characteristic |
| SQL…………………. | Structured Query Language |
| SVM………………… | Support Vector Machine |
| VQL………………… | Visual Query Language |
| YAML……………… | Yet Another Markup Language |

# CHAPTER 1

# INTRODUCTION

## 1.1 Background

In today's world, we're flooded with data from all directions, making it essential to find ways to make sense of it all. Data visualization steps in as a handy tool that transforms complex information into easy-to-understand pictures like charts or graphs. Imagine trying to read a giant spreadsheet versus seeing a clear graph – that's the power of data visualization. It's not just about making things look pretty, it helps us spot trends, connections, and important details in a way our brains can grasp effortlessly.

In recent years, the intersection of Natural Language Processing (NLP) and data analysis and visualization has emerged as a dynamic field with the potential to revolutionize information communication and comprehension. One notable application within this domain is the generation charts from textual input, providing a more dynamic and user-friendly presentation of data. The synthesis of text and charts holds promise across diverse domains, including business analytics, scientific research, healthcare, and journalism.

By leveraging NLP techniques to convert natural language to charts and then extract insights from charts, Decision-making processes can be enhanced and facilitate deeper understanding of complex datasets. This integration opens new opportunity for exploring and interpreting data, ultimately enabling more effective communication and knowledge.

Since the arrival of Large Language Models (LLMs), it has become common practice to use them to improve results in many different areas. Nowadays, LLMs are being applied extensively across various fields to achieve better outcomes and create systems that interact more like humans.

Adding large language models to data analysis and visualization not only makes complex data easier to understand but also can easy the process of decision making. Using these advanced language models helps create better visualizations that understand human language better. This mix of language and visualization could change how data can be understood and share information. It might help to make smarter choices and see things more clearly in lots of different areas.

## 1.2 Problem Statement

The current landscape of data analysis and visualization systems is characterized by a static nature, where visualizations are generated based on predefined data and chart types. This rigidity becomes a significant drawback when user requirements deviate from the predetermined visualization techniques, making it difficult to obtain customized visual representations from the existing systems. The problem at hand is the lack of adaptability and flexibility in current data visualization tools, which restricts their capacity to effectively cater to varying user needs and preferences, as per (Wu et al., 2022).

Furthermore, the interactive aspect in current data visualization presents a substantial obstacle. When data visualization lacks interactivity, it fails to provide meaningful insights or convey narratives associated with the depicted data. Moreover, extracting insights from charts remains a predominantly manual process, necessitating a comprehensive understanding of diverse chart types and the methodologies for deriving insights from them, thus presenting additional challenges.

As technology grows and data gets more complex, visualization tools that can change with the times are needed. Imagine if charts and graphs could adjust automatically as data changes or as per user's ask for different questions. That's the idea behind adaptive visualization. Instead of being stuck with the same old static graphs, new tools would be smart enough to adapt to new data and user needs. This could make exploring data easier and more insightful, helping user to uncover hidden patterns and tell better stories with data. It's like having a dynamic assistant that helps to make sense of information in real-time.

Furthermore, after generating a chart for data visualization, it becomes easier to understand the insights it offers. This reduces the need for manually analyzing complex charts to uncover insights and allows for making informed business decisions based on the data presented in the chart.

## 1.3 Research Questions

The following research questions are proposed for the study, as highlighted as follows:

1. How can user-friendly data analysis and visualization problems solved and make it easier to create interactive chart?
2. How can Natural Language Processing techniques or specifically LLMs can be used to enhance user-friendly data visualization?
3. How can data analysis tool effectively give useful information from the charts they generate, helping to make better business decisions?

## 1.4 Aim and Objectives

The research aims to propose a multi-model architecture designed for data analysis by generating charts dedicated to visualization from input data as per user's natural language inputs, with insights from the resultant visual representations. Also, to make the data visualization tool adaptable to different fields and datasets. It focuses on improving the accuracy of data visualizations to reduce the chances of making incorrect decisions. Furthermore, this research aims to develop question-answering model capable of deriving both historical and futuristic insights from contextual data within the input document, by leveraging the capabilities of the BERT model.

The research aim is divided into objectives, which are as follows:

- To develop a model dedicated to converting natural language input into data visualizations and their associated insights, leveraging the capabilities of Llama3.
- To develop a model that provides natural language responses to user queries based on the input document by leveraging capabilities of BERT model.
- To develop a model for providing future predictions based on available historical data, leveraging BERT's capabilities.

## 1.5 Significance of the Study

Data analysis nowadays requires a flexible approach to meet the different needs of various industries. Presently, most data insights are shown through pre-defined charts and static analyses, which don't give users much flexibility. Organizations usually provide set visualizations based only on specific data sets. So, when users come across situations where they need different analyses for various inputs, they face significant difficulties.

The significance of this research lies in its potential to address this inherent limitation. By developing a model facilitating the transformation of natural language input, into personalized and context-specific data analysis and visualizations, the study seeks to empower users with the capability to conduct data analysis and visualization according to their distinct preferences. This novel approach aims to enhance the adaptability and user-friendliness of data visualization, contributing to a more dynamic and intuitive exploration of diverse analytical scenarios.

## 1.6 Scope of the Study

This research paper is delimited to the analysis, visualization and predictive modelling of data within the context of cricket dataset, primarily due to time constraints. The chosen scope ensures focused and in-depth exploration of the proposed methodology within the specified timeframe. The dataset used for experimentation and validation is exclusively related to cricket.

Furthermore, the research predominantly concentrates on the English language, considering the challenges and resource-intensive nature of training models in multiple languages. Given the complexity and time requirements associated with multilingual model training, the decision to narrow the scope to English facilitates a more efficient and targeted investigation.

However, it is important to note that the designed model, once established and validated on the cricket dataset in English, holds the potential for broader applicability. Future endeavors can extend the use of the developed approach to diverse datasets beyond cricket and languages other than English. This adaptability allows for the scalability and versatility of the proposed model, opening opportunities for cross-domain and multilingual applications beyond the initial cricket-focused and English-centric scope.

## 1.7 Structure of the Study

The structure of the research is as follows.

Chapter 1 provides the Introduction for the study, which includes the background of the current Data Visualization techniques in section 1.1, followed by an examination of the problem statement in section 1.2. Section 1.3 talks about the various research question which needs to be answered. Section 1.4 defines the aims and objective for the research. The significance of the study is explored in section 1.5, chapter 1 concludes with section 1.6, which defines the study's scope.

Chapter 2 serves as a comprehensive Literature Review encompassing recent research within the domain of data visualization from natural language text. Within this chapter, Introduction of Literature Review is in section 2.1, followed by History for data visualization in section 2.2. Section 2.3 provides an overview of available dataset within this domain. Subsequently, Section 2.4 examines various approaches for data visualization, including Text 2 Visualization, Document Question Answering, Tabular Document Question Answering, Text 2 SQL and the utilization of Named Entity Recognition in Section 2.4.1, 2.4.2, 2.4.3, 2.4.4, 2.4.5 respectively. Recent surveys in this domain are synthesized in section 2.5. Section 2.6 critically analyses the limitations and challenges encountered in the reviewed literature. Finally, Section 2.7 provides a concise summary of the contents of Chapter 2.

Chapter 3 defines the Research Methodology implemented for this study. It begins with an introduction to the Research Methodology in section 3.1, followed by the presentation of a multi-model architecture in section 3.2. Within section 3.2, the analysis of input CSV file data is addressed in section 3.2.1, while section 3.2.2 covers the transformation of input text into SQL queries. Section 3.2.3 elaborates on the fine-tuning of SQL, and section 3.2.4 discusses data subset generation. The prediction of chart types is detailed in section 3.2.5, followed by the exposition of chart generation techniques in section 3.2.6. Insight generation from the final model output is explored in section 3.2.7. Section 3.3 discusses the research methodology for the question-answering model. Section 3.4 outlines the research methodology employed for the predictive model, with Section 3.4.1 focusing on the binary class predictive model and Section 3.4.2 on the multi-class predictive model. Section 3.5 covers the evaluation methods for the developed models, with a summary provided in Section 3.6.

Chapter 4 talks about the analysis and experiments conducted for this study. It begins with an introduction to Data Analysis in Section 4.1, followed by the Dataset Description in Section 4.2. Section 4.3 covers Dataset Preparation. Data analysis is addressed in Section 4.4, with its subsections as follows: Section 4.4.1 discusses the statistical analysis of the data, Section 4.4.2 covers exploratory data analysis, Section 4.4.3 focuses on time series analysis, and Section 4.4.4 addresses predictive analysis. The next section 4.5 is about question-answering model. The next section, 4.6, is about Predictive Modelling, with Section 4.6.1 detailing binary class predictive modelling and Section 4.6.2 focusing on multi-class predictive modelling. Finally, Section 4.7 provides a summary of the analysis and experiment chapter.

Chapter 5 discusses the Results and Discussion of the study. It begins with an introduction in Section 5.1, followed by the Evaluation of the Text2Insight Model in Section 5.2. Within Section 5.2, the Evaluation of the Text2SQL Model is covered in Section 5.2.1, the Evaluation of the Chart Type Prediction Method is in Section 5.2.2, and the Evaluation of the Insights Generation Model is in Section 5.2.3. Section 5.3 addresses the Generated Charts from Natural Language. Next section 5.4 is about evaluation of question-answering model. Section 5.5 focuses on the Evaluation of the Winner Prediction Model, with the Evaluation of the Binary Classification Model in Section 5.5.1 and the Evaluation of the Multi-Class Classification Model in Section 5.5.2. Finally, the summary of Chapter 5 is discussed in section 5.6.

The last chapter is Conclusions and Recommendations in Chapter 6. It begins with an introduction in Section 6.1, followed by Discussion and Conclusions in Section 6.2. Contributions are covered in Section 6.3, and finally, Section 6.4 presents Future Recommendations.

# CHAPTER 2

# LITERATURE REVIEW

## 2.1 Introduction

In today's digital age, numerous studies cover various topics. Before attempting to solve a problem, it's advisable to review existing research in the specific area. This can save time and prevent duplication of efforts. If a method is found to be ineffective, alternative approaches can be explored.

This chapter aims to examine existing research on analyzing or visualizing data from natural language text. It focuses on available methods, models, and datasets, as well as how large language models can enhance data visualization performance. Survey papers also reviewed to understand the current state of research in this field. Additionally, challenges and limitations in current data visualization systems is identified and discussed.

## 2.2 Evolution of Data Visualization: A Historical Perspective

As discussed in survey paper (Zhang et al., 2023b), Methods for data visualizations have changed a lot to match the increasing need for making decisions based on data. At first, tools like Vega-Lite and D3.js were created to help people show data visually. These tools allowed for a wide range of visualizations, from simple graphs to more complex heat maps. Later, Prolog and Datalog were introduced for querying databases using logic. Prolog focused on artificial intelligence and symbolic reasoning.

FunQL, a Functional Query Language, was developed to make it easier for people to ask questions about data using natural language. Moving from traditional methods to using neural networks for turning text into visualizations was a big step forward. It made interpreting user requests for visualizations more accurate and efficient. Figure 2.1 show how people are always working to make it easier for anyone to interact with data.

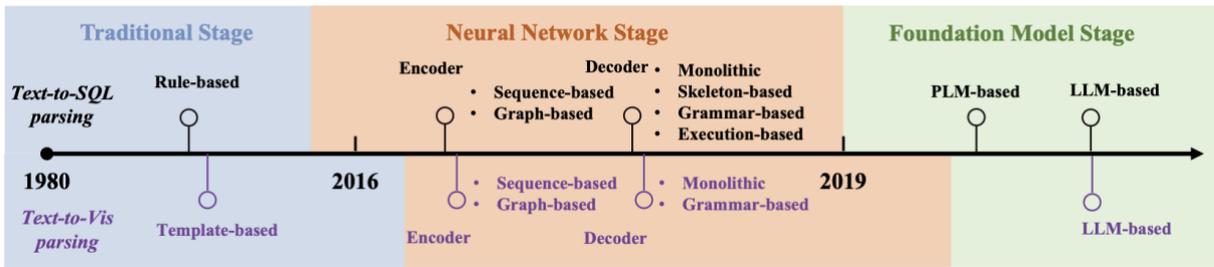

Figure 2.1: Evolution of Text-to-Vis approaches over time, as per paper (Zhang et al., 2023b)

## 2.3 Exploring Datasets for Natural Language-Based Data Visualization

There are three primary datasets available for converting visualization from natural language text. These datasets contain natural language samples along with corresponding charts. Researchers can utilize these datasets to accomplish tasks such as Natural Language to Visualization (NL2VIS).

The one dataset introduced is nvBench (Luo et al., 2021). It comprises JSON-based objects containing natural language queries and their corresponding SQL, VQL, chart types, and x and y-axis. This dataset aims to assist in the Natural Language to Visualization task by providing a large and varied set of training data. It includes 25,750 pairs of natural language queries and visualizations taken from 750 tables across 105 different domains. The intention behind this dataset is to address the limitations of previous datasets by offering a more extensive and realistic collection for NL2VIS tasks.

The method described in the paper for creating the nvBench dataset significantly reduces the time required compared to manual creation. It is suggested that manual creation would take 17.5 times longer than the proposed synthesis method, demonstrating the efficiency and effectiveness of this approach in generating high-quality NL2VIS benchmarks.

However, when adjustments are needed to NL queries for visualization purposes, user involvement becomes necessary, potentially causing delays and reliance on user input. While verification processes by experts and crowd-workers help ensure quality, subjective judgments may affect the accuracy of the synthesized (NL, VIS) pairs. Additionally, dealing with complex operations, such as data joining from multiple tables, may require extra effort and expertise, potentially impacting the scalability of the synthesis process.

Another available dataset is, CoVis (Haque et al., 2022). Which aims to create visual representations from conversations, which can be used in interactive systems for discussing data visualizations. This dataset, based on the nvBench dataset but modified for CoVis tasks, contains dialogues between users and chatbots, along with queries and responses related to creating visualizations from text. It includes various statistics such as the number of dialogue sessions, different types of queries, and the number of datasets involved. It enhances CoVis research and development by improving the dataset with more diverse samples.

However, it's important to note that the CoVis dataset derived from nvBench has limitations, as it may not cover all possible dialogue scenarios and query types.

Another dataset called CHARTDIALOGS (Shao and Nakashole, n.d.) is available, which comprises 3,284 conversations, 15,754 individual exchanges, and a total of 141,876 words. It was specifically designed to help train computer programs to update charts based on spoken instructions. In comparison to similar datasets used for training computer programs with specific tasks, CHARTDIALOGS stands out due to its large size and complexity.

One of the challenges faced is accurately describing certain aspects of the data in spoken language, which can result in errors in the predictions made by computer programs. Moreover, these programs encounter difficulties when handling types of charts that contain limited data points or are unfamiliar to them. This makes it challenging for the programs to learn effectively, especially when provided with minimal or no examples.

**2.4  Diverse Methodologies: Exploring Various Approaches**

There are several methods for transforming natural language into data analysis or visually appealing charts. One common approach involves directly converting natural language into charts using various large language models. Another method is to first convert the natural language into Visual Query Language (VQL), then apply this VQL to a dataset to create the chart. Alternatively, one can convert the natural language into Structured Query Language (SQL) and execute it on the dataset to generate charts from the resulting data. Another approach involves treating the dataset as an image and using natural language text to extract charts from the image. Additionally, specific entities can be extracted from natural language using Named Entity Recognition (NER) models, and then charts can be retrieved based on the refined query using NER. Finally, visualization can be obtained from tabular images, where tabular data is extracted from the image and converted into charts.

### 2.4.1 Text to Visualization

A notable contribution in this field is Text2Chart (Rashid et al., 2021), a multi-staged chart generator in the realm of natural language processing. It used 717 text samples from various sources like Wikipedia and statistical websites. These samples were chosen to help Text2Chart recognize information needed to create different types of charts, such as pie charts, line charts, and bar charts. Different models like Bidirectional Encoder Representations from Transformers (BERT), Bidirectional Long Short-Term Memory (LSTM), Support Vector Machines (SVM), Random Forest, and Feed Forward Neural Networks were tested. It used neural networks with various word embeddings and sequence representations.

The research explored to find ways to identify the x-axis and y-axis information in the text and connect them to create accurate charts. BERT embeddings and Bidirectional LSTM were found useful for this task. The provided system could predict which type of chart (pie, line, or bar) was suitable based on the given text and achieved good results in recognizing the x-axis and y-axis information. Precision, Recall, F1-score, and AUC ROC were used to measure its performance, which was satisfactory.

However, the system's effectiveness depended on the quality and variety of the training data. It faced challenges when dealing with complicated texts containing unclear information. It also had limitations in handling unusual cases that didn't fit into standard chart types. Also, the task of mapping x and y entities is noted to be highly imbalanced, resulting in a bias towards negative mappings in the model. Furthermore, its ability to work with new or different types of data was limited because it was trained on specific datasets.

Another research for chart generation have been made with the introduction of ChartLlama (Han et al., 2023), a multimodal large language model for creating charts. Its goal is to improve the variety and quality of the charts produced by using new methods to generate data with GPT-4. The model's performance is tested on tasks like answering questions about charts, turning charts into text, and extracting information from charts. The dataset used is a standard set created for training and testing ChartLlama. This dataset covers many types of charts and tasks, allowing for more varied data to train the model according to specific needs. It's created using a new process that relies on GPT-4 to generate data specifically for chart-making tasks.

ChartLlama uses these new methods to improve the variety and quality of its generated charts. It's evaluated based on how well it performs tasks like answering questions about charts, turning charts into text, and extracting information from them. The focus is on using one model to handle the entire process of understanding charts, aiming to make the generated charts better and more varied.

ChartLlama performs well compared to other models in tasks like precision and F1 score, beating models like Pix2struct, Matcha, and Unichart in some areas. Its new data generation methods greatly improve its ability to create accurate charts, making the results more varied and controllable.

However, ChartLlama, which can understand both text and images, have difficulties with complex chart data. The datasets mainly support simple tasks like answering questions or writing captions, lacking thorough support for advanced chart tasks. If these models are trained too much on visual data, they might lose their ability to generate code, affecting their performance on tasks that need complex chart interpretation.

Another notable research is Chat2VIS (Maddigan and Susnjak, 2023), for Natural Language to Visualization solutions, with a particular focus on addressing challenges in generating data visualizations. Chat2VIS leverages the functionalities of pre-trained, large-scale language models like GPT-3, Codex, and ChatGPT, demonstrate a robust integration of natural language processing with chart generation. Also, Chat2VIS has ability to effectively interpret and respond to diverse, multilingual user queries.

It used datasets called nlvUtterance and nvBench to provide examples of queries and the visualizations they produce. In the evaluation of Chat2VIS, it used the Codex "code-davinci-002" model and compared its performance against nvBench. And compared how well Codex,

GPT-3, and ChatGPT generate visualizations from natural language queries. The research showed how Chat2VIS can improve visualizations by refining charting elements based on additional instructions given in queries. Additionally, research demonstrated how these language models can respond to additional instructions to enhance visualization outcomes and handle requests in multiple languages.

Chat2VIS research has found that unclear queries could lead to different interpretations by the models and benchmarks, resulting in varying visualizations. Also, there is challenges in ensuring the accuracy of visualizations when queries are ambiguous or open to interpretation. Also, issues such as inconsistencies in syntax generation affecting the rendering of desired plots, especially those involving mathematical computations. And noted mismatches in the types of charts generated by the models, such as histograms being plotted as bar charts or scatter charts lacking proper color coding. Lastly, there are concerns about the consistency between test case data and visualization outcomes, which affects the reliability of benchmark examples.

Another research for text to graph generation is CycleGT (Guo et al., 2020), that addresses the challenge of training on non-parallel graph and text data. CycleGT adopts an iterative back-translation strategy between graphs and text, aiming to bridge the gap between these modalities without the need for labelled data. It utilizes different datasets like the WebNLG 2017 Dataset, WebNLG 2020 Dataset (WebNLG2020, n.d.), and the GenWiki Dataset (Jin et al., n.d.). For text-to-graph tasks, it employs BiLSTM, while for graph-to-text tasks, it uses T5. Additionally, RuleBased and GT-BT models are used for graph-to-text tasks. The document implements relation extraction through Convolutional Neural Networks (CNNs) and Recurrent Neural Networks (RNNs).

It evaluates the performance of the CycleGT model against other supervised and unsupervised models on the GenWiki dataset, where CycleGT achieves a Bilingual Evaluation Understudy (BLEU) score of 55.5 in graph-to-text tasks, indicating strong performance. Supervised models also perform well in graph-to-text tasks, underlining the significance of model supervision for accuracy.

However, it is important to note that the model faces challenges like the large-scale datasets used for training to achieve human-level performance. Aligning text and graph representations is difficult due to the absence of direct correspondence. Furthermore, models trained on specific datasets may struggle to generalize to new data or different domains. Lastly, optimizing models within the cycle training framework encounters difficulties with non-differentiable intermediate outputs.

Another notable research is ChartQA (Masry et al., 2022), this research introduces two transformer-based models specifically designed for the generation of charts and question, answers from charts. A pipeline approach is employed, integrating visual features with extracted data from charts to enhance overall performance. The paper proposes a new standard for assessing visual and logical reasoning skills using questions created by humans. It also provides specific settings for experiments to ensure that they can be replicated by others. Also, paper used dataset, called ChartQA Benchmark (Masry et al., 2022), designed for tasks involving visual and logical reasoning. The dataset consists of charts from various sources to cover a wide range of topics and styles. These charts were then used to enhance the ChartQA Benchmark with positional embeddings for tabular data.

The used model in the paper is fine-tuned models like T5 over multiple epochs on different datasets to optimize their performance. Performance evaluation of these models based on their accuracy and how well they performed on various types of charts and question categories. It founds that providing the correct data table significantly improved accuracy. While these fine-tuned models performed competitively, they did not show much improvement over T5, especially in handling questions with multiple references. Some models, like VisionTaPas and VL-T5, performed better on bar charts.

However, the limited options for answers in the ChartQA datasets made it difficult for models to handle complex reasoning questions, especially those requiring mathematical operations. Models also struggled with nested mathematical and logical operations, which affected their performance on more challenging questions. Finally, existing evaluation metrics did not adequately account for noisy textual tokens extracted from real-world figures, which impacted the accuracy assessment of these models.

Another research NL4DV (Narechania et al., 2021), Natural language for Data Visualization which is focused on the generation of data visualization using natural language processing. NL4DV, is a toolkit which serves as a powerful tool for facilitating natural language-driven data visualization by proficiently interpreting natural language queries and generating JSON-based analytic specifications. Within the generated JSON object, NL4DV encapsulates essential elements, including data attributes, analytic tasks, and Vega-Lite specifications.

The NL4DV toolkit aims to help visualization developers understand natural language queries and make visualizations based on the identified attributes and tasks. It wants to make the process of creating visualizations easier by using natural language input, which can sometimes be unclear or not detailed enough. NL4DV wants to make it easier for developers to work with data by using natural language interfaces. It started with a dataset about IMDb movies, including information like IMDB Rating. NL4DV uses a method called heuristics to understand natural language queries and figure out the data attributes and tasks involved. It follows a process of analyzing the query, parsing taskMap and visList, and adjusting view mappings accordingly. The toolkit has been able to create visualizations based on natural language input, showing that it can understand attributes, tasks, and make relevant charts.

However, because it uses heuristics, there might be some uncertainties in its output, which could lead to inaccuracies in the visualizations. Another challenge NL4DV faces is incorporating interface context, like active encodings and selections, into a format that can be understood by different types of toolkits, which might limit its ability to work in different visualization situations.

Another study (Wang and Crespo-Quinones, 2023) which aims to investigate how large language models such as BERT and T5 can be used to predict visualization commands from natural language queries. The main objective of this research is to improve the efficiency of data analysis by automatically converting queries into visual representations. The dataset used in the paper is nvBench dataset. The nvBench dataset contains pairs of natural language queries and data visualization commands in vega zero syntax, which helps in training and testing models for translating queries into commands.

This study uses the basic model, employing a sequence-to-sequence transformer architecture and fine-tuning with a BERT encoder to predict visualization commands. T5 models fine-tuned on the nvBench dataset are also experimented with for better prediction accuracy. Additionally, the study examines the impact of BERT embeddings on the baseline model by combining ncNet and BERT encoders. Results indicate that CodeT5 models consistently outperform other models, achieving high accuracy in predicting query structures and labels.

However, most errors observed were related to meaning or abbreviations, suggesting areas for enhancing model training and fine-tuning. Some prediction errors were found to be systematic, highlighting the need for improved error correction mechanisms. It's noted that the syntax rules for visualization queries may not support complex requests or queries requiring table joins, which may limit the effectiveness of model training and testing.

### 2.4.2 Document Question Answering Models

A notable contribution in the field of document question answering is CFRet-DVQA (Zhang et al., 2024), to improve document visual question answering (DVQA) using a new method. This method involves multiple steps to find relevant information in documents and fine-tune how the system works based on the results. The goal of this research was to overcome problems with existing models and make DVQA better.

The researchers tested their method on five different datasets commonly used for DVQA research. It used specific software to help with the complex tasks involved. The experiments were run on a powerful computer system. The CFRet-DVQA framework has two main stages: first, it selects larger pieces of text, then it refines the selection to smaller, more relevant parts. The study found that this method was more accurate than simpler methods at finding the right information. The researchers also showed that their method worked well at understanding complex documents and solving problems.

However, the method didn't consider certain aspects of text, like how it's laid out or if there are images, which could be important in real-world situations. The study didn't discuss any difficulties that might arise when using CFRet-DVQA in real-life situations.

Another research paper introduces a new framework called DocGraphLM (Wang et al., 2023b). It combines language models with graph semantics to extract information from visually complex documents. The goal of the research is to improve tasks like finding entities and answering questions about images by using graph features and a special way of processing information called a joint encoder architecture.

In this study, the framework was tested on three common datasets: FUNSD, CORD, and a modified version of DocVQA. DocGraphLM integrates pre-trained language models with graph neural networks (GNN) and uses a joint encoder to understand documents better. It also suggests a method for predicting connections between parts of a document to help understand its structure. The results shows that DocGraphLM is effective at improving how documents are represented and performs well on tasks like extracting information and answering questions.

However, it's important to note that the evaluation focused mainly on tasks involving visually complex documents, so it can't be sure how well it would work in other areas. Additionally, the presence of errors in the datasets, especially in the OCR (optical character recognition) outputs, made it difficult to find accurate answers. The modified DocVQA dataset used for training and validation was also smaller than usual because of these errors.

Another research paper which introduces a new dataset called DCQA (Wu et al., 2023), which stands for Document-level Chart Question Answering. The aim of this research is to help understand visual representations, like graphs and charts, found in documents, especially in cases where understanding them requires complex thinking. The study also presents a tool for generating questions and answers, as well to understand charts without needing Optical Character Recognition (OCR), focusing on common-sense reasoning.

This dataset contains various types of charts and questions, allowing researchers to explore different aspects of understanding charts. The study also introduces a model called TOT-Doctor, which has two main parts: analyzing how documents are laid out and answering questions about charts. This model uses a Swin Transformer to understand document layouts and BERT to answer chart questions. TOT-Doctor consistently performs better than other models in terms of accuracy on both validation and test datasets, showing that it works well. The study also shows that analysis part of TOT-Doctor significantly improves its ability to answer questions about charts.

However, there are some limitations in this research. These include having a limited variety of chart types, a limited number of question templates, and relying on OCR for some tasks.

### 2.4.3 Tabular Document Question Answering Models

A notable contribution in the field of tabular question answering is TabIQA (Nguyen et al., 2023), uses advanced deep learning methods to analyze tables and answer questions about numerical data, text information, and complex queries. The research addresses challenges in understanding table structures, referring to information across different documents, and performing complex calculations beyond simple searches.

The dataset utilized is called VQAonBD (Agrawal et al., 2015), which contains images of documents from the FinTabNet dataset along with relevant questions about these images. The developed model in this research comprises a shared encoder and decoder, along with separate decoders for specific tasks such as recognizing table structures, detecting cells, and recognizing cell contents. It treats table recognition as a multi-task learning challenge and achieves high accuracies on datasets like PubTabNet and FinTabNet. TabIQA surpasses baseline models like TAPAS, TAPEX, OmniTab, and the Zero model.

However, there are limitations. The current TabIQA system may struggle with complex queries that involve reasoning across multiple tables or non-tabular sections of documents. Additionally, its performance might be limited when dealing with tables from various domains due to insufficient domain-specific training data.

Another research which introduces a method called Table-to-Text Generation (TTGen) (Li et al., 2021) that improves Machine Reading Comprehension (MRC) models when answering questions based on tables. The method utilizes the GeoTSQA dataset, specifically designed for Tabular Scenario based Question Answering (TSQA), containing tables with numeric content essential for answering questions. Additionally, it mentions the use of the GeoSQA dataset (Huang et al., 2019) for gathering questions and code reuse to expand the dataset.

The model developed in this research generates top-k sentences from tables using templates to manipulate tabular data and incorporates K-BERT, a knowledge-enabled language model, for question and knowledge-aware ranking. TTGen demonstrates better performance than RE2 in terms of Mean Average Precision (MAP) and Mean Reciprocal Rank (MRR), resulting in improved sentence ranking accuracy.

However, the method developed in this research heavily relies on domain-specific knowledge for accurate question answering, which limits its use across different domains. Moreover, the large number of generated sentences may overwhelm typical Machine Reading Comprehension (MRC) models, leading to noise and reduced accuracy. Lastly, its performance may vary across different datasets and domains, indicating challenges in generalizing its effectiveness.

Another research in the field of tabular question answering is, TAT-QA dataset and TAGOP model (Zhu et al., 2021). Which is designed to help in creating question-answering systems for mixed data, particularly in finance. The introduced dataset contains questions that needs numerical thinking, such as addition, subtraction, multiplication, and division. It's built from 182 financial reports.

The introduced model uses sequence tagging and aggregation methods to find answers from tables and text. It offers ten operators like Span, Sum, Count, and others to do numerical and symbolic reasoning on the data. TAGOP works better on questions with answers in tables rather than text.

However, it finds arithmetic questions challenging, which shows the difficulty of handling numerical tasks in finance. Different operators in TAGOP affect its performance differently. Some, like Difference and Change ratio, are more helpful than others. The model struggles with arithmetic questions, indicating a need for improvement in handling various numerical calculations effectively.

Another research (Khalid et al., 2007) whose purpose is to explain how machine learning can be used to answer questions based on tabular data in Question Answering (QA) systems. The study discusses using a standard set of questions and answers from the TREC QA tasks, and it employs techniques like vector space modeling and language modeling for retrieving relevant information. Additionally, it explains how question features are generated using part-of-speech tagging and named entity tagging. The study also explores different methods for representing text, retrieval models, and normalization techniques to enhance performance.

The model used in this study achieved moderate success with an Mean Reciprocal Rank of 0.234 and accuracy rates ranging from 11.09% to 28.7% at various ranks. It showed that using machine learning models and data-driven approaches can improve the accuracy and efficiency of answering questions.

However, the study deals with various challenges such as dealing with noisy data, combining information from different sources, and using a data-driven approach to answer questions from tabular data. And normalizing named entities to their standard forms may still encounter issues with ambiguity, affecting entity recognition accuracy. Moreover, the model's reliance on keyword-based relevance functions may limit its ability to understand complex semantic relationships within the data. Lastly, the quality and diversity of training data used for machine learning models can significantly affect the system's performance and its ability to generalize to new situations.

### 2.4.4 Text to SQL

A notable contribution in the field of natural language text to SQL conversion is MAC-SQL (Wang et al., 2023a), which introduces a framework designed to tackle difficulties encountered in tasks involving converting text into SQL queries. The introduced model does this by making database structures simpler, breaking down complex user queries, and checking and fixing SQL errors. This framework uses three main components: Selector, Decomposer, and Refiner. These components work together to make the conversion process smoother and to enhance the accuracy and efficiency of generating SQL queries from text.

Two datasets, Spider Dataset (Yu et al., 2018) and BIRD Dataset, are used for training and testing the framework. The Spider Dataset consists of 7,000 question-query pairs in the training set and 1,034 pairs in the development set. On the other hand, the BIRD Dataset includes 95 large-scale real databases with high-quality text-SQL pairs. Similarly, like the MAC-SQL framework, the BIRD Dataset also employs three agents: Selector, Decomposer, and Refiner for Text-to-SQL tasks.

Another pre-defined model called SQL-Llama, which is based on Code Llama, is fine-tuned using instructions from the MAC-SQL framework. This fine-tuning process involves using the agent instruction data from MAC-SQL to enhance performance. The framework effectively addresses challenges related to simplifying database structures, breaking down queries, and correcting SQL errors. The study has shown better performance in these areas compared to other methods. Also, the effectiveness of the multi-agent collaboration approach employed by the MAC-SQL framework in improving Text-to-SQL tasks has been validated.

However, study has limitation which is lack of detailed information on the specific evaluation metrics used to measure model performance. Additionally, challenges in dealing with extensive databases, complex user queries, and incorrect SQL results may persist despite improvements made to the framework.

Another research is, DBCopilot (Wang et al., 2023c). Which is designed to create a system that can effectively translate natural language queries into SQL queries, regardless of the structure of the database being queried. It uses a two-step process: first, it identifies the relevant parts of the database schema, and then it generates the corresponding SQL query. DBCopilot has been tested using various datasets, including Spider, Bird, Fiben, Spidersyn, and Spiderreal. Study employs a neural network to automatically match questions to the appropriate database schema. Additionally, system creates a visual representation of the relationships within the database. By using Large Language Models and schema-aware prompts, DBCopilot can generate accurate SQL queries. It has shown better performance compared to other methods in terms of accurately retrieving information from databases.

However, continuously updating the LLMs with schema information can be resource intensive. Additionally, methods that rely on retrieving information directly from the database can struggle with the diversity of natural language queries and differences in vocabulary.

Another research in this area is, CRUSH (Kothyari et al., 2023), which combines language model hallucination and dense retrieval to pinpoint a small yet highly recallable subset of schema elements for a Text-to-SQL process. The datasets used for this study are SpiderUnion and BirdUnion. The process involves converting the user's query into a simplified form using a Language Model. The focus is on identifying a subset of schema elements from the database that match with the imagined schema derived from the user's question. Language Models are employed to create a basic schema based on user queries and limited examples. A new optimization approach is introduced to select a small-sized schema subset with high recall.

CRUSH outperforms other methods based on Single DPR and token-level techniques on datasets like SpiderUnion and SocialDB. It also shows better results even in scenarios where no examples are available, especially for simpler schema elements found in the SpiderUnion dataset.

However, a limitation of CRUSH is that its current hallucination lacks guidance from the client DB schema, potentially reducing the accuracy of schema generation. Research is underway to find ways to incorporate information from the client DB schema into the prompt text, which could improve the performance of the Language Model Schema Hallucinator.

Another notable research around this area is, SQLformer (Bazaga et al., 2023). Which is used for translating text into SQL queries. It highlights the difficulties in understanding queries when dealing with new databases and stresses the importance of accurately representing both natural language queries and the structures of the databases they query. The paper usages the Spider dataset to test the SQLformer model's performance.

The model itself employs a Transformer encoder-decoder setup, with added features such as learnable tokens for tables and columns in the encoder, which help in selecting the relevant parts of the database schema. Additionally, the Transformer decoder is extended to incorporate node adjacency and type embeddings to generate SQL queries. The paper also compares the performance of the SQLformer model with other models like RAT-SQL, T5-3B, LGESQL, and GRAPHIX-T5-3B, showing that it performs competitively.

However, it notes that the model relies on external libraries for tasks like query transformations, which could lead to compatibility issues and maintenance difficulties.

Another research is, ACT-SQL (Zhang et al., 2023a). Which aims to explore different methods used in text-to-SQL research, particularly focusing on improving how well Large Language Models can generate SQL queries from natural language questions. It covers topics such as creating database structures, using Auto-CoT to connect schemas, employing the ACT-SQL method to boost LLM performance, and presenting detailed experimental findings comparing various approaches. The dataset which is being used in this research is Spider dataset.

The ACT-SQL approach incorporates the CoT method to guide LLMs in constructing the correct SQL query, thus aiding in logical reasoning. Additionally, it combines grammar-based text-to-SQL models with graph encoding to encode both the database schema and the question together. The study provides examples of SQL queries generated for different questions within the database schema, illustrating how effective the ACT-SQL approach is in achieving top-notch performance.

However, the hybrid strategy for selecting examples in the ACT-SQL approach requires manual tweaking of hyperparameters for both static and dynamic examples, indicating the need for further refinement. Furthermore, the two-phase method used in multi-turn text-to-SQL datasets could result in performance challenges, stressing the importance of devising better strategies to manage context dependencies and schema connections.

Another research (Gao et al., 2023) whose purpose is to discuss and evaluate different methods for quickly creating instructions in the context of turning text into SQL queries, using Large Language Models. To do this, study looked at different ways of representing questions, selecting examples, and organizing them. Research also explored if training the LLMs with specific examples can improve their performance.

Study uses two datasets called Spider and Spider-Realistic. It also sees if training LLMs with examples from these datasets can help. And tried different approaches, like Basic Prompt and Alpaca SFT Prompt, to make LLMs perform better. The results shows that these strategies can indeed improve how well LLMs handle Text-to-SQL tasks.

However, it's important to note that the effectiveness of these strategies might depend on which LLMs being used. Also, other factors like biases in the datasets or the models themselves, as well as changes in the environment, could affect its findings.

Another research for Text2SQL tasks is, SeqGenSQL (Li et al., n.d.). Which discussed about different ways to make SQL statements more accurate using the T5 model. The study looked at things like trying out different methods, using a reversed trainer model, and a technique called execution guided inference. Study also discussed the difficulties in generating SQL statements and how the T5 model can help.

Research usage a dataset called WikiSQL, which has examples of SQL queries and data for each table. It uses T5 model to generate SQL statements from sequences. SeqGenSQL model, along with execution guidance, achieved a 90.5% accuracy on the test dataset, which is better than previous models.

However, there are differences between what the correct answer should be and what the question suggests, which can cause the model to make mistakes. Dealing with these issues, like mistaken information and unclear questions, is still a challenge for the study. For future work, it has suggested to investigate bigger base models, improving how it extract information, and using more data to make the model even more accurate.

### 2.4.5 NER Models

A notable contribution in the field of Name Entity Recognition is, TOPRO (Ma et al., 2024) method. A new approach for labeling sequences of tokens. It examines how effective TOPRO is in improving performance in tasks like Named Entity Recognition (NER) and Part-of-Speech (POS) tagging, especially when dealing with different languages. The study uses two main datasets: PAN-X and UDPOS. It also tests Multilingual Pre-trained Language Models (MPLMs) in two configurations: Encoder-Only Models and Encoder-Decoder Models.

In the experiments of this study, TOPRO employs prompt patterns and verbalizers to help the model predict labels for each token in a sequence. This aligns with the typical process in text classification, where the goal is to assign a label to each piece of input text. TOPRO shows improved accuracy and efficiency, particularly in situations involving multiple languages.

The study evaluates various MPLMs with TOPRO fine-tuning, but notes that TOPRO sometimes generates identical prompts for tokens that appear more than once in a sentence, which can lead to errors. Additionally, the authors manually designed the prompt patterns and verbalizers used in the study.

Another research in this field is, GLiNER (Zaratiana et al., 2023). The purpose of this research paper is to discuss the Multiconer model and its performance, including metrics, results, methodology, and the evolution of Named Entity Recognition tasks. It offers insights into the GLiNER model's components, such as entity type and span matching, and the shift from rule-based to machine learning approaches in NER. The study mentions the use of the Pile-NER dataset to train the GLiNER model, which is derived from the Pile corpus, a collection of diverse text sources commonly used for pretraining large language models.

The GLiNER model uses a pre-trained textual encoder and a span representation module for NER tasks. Study aims to improve upon existing models by utilizing smaller-scale Bidirectional Language Models and diverse training data, resulting in enhanced NER capabilities.

The results of the study indicate improvements in precision, recall, and F1 scores achieved through in-domain supervised fine-tuning and ablations such as negative entity sampling. The GLiNER model exhibits promising performance across various domains and languages, demonstrating its adaptability and efficiency in the NER task.

However, there are limitations, including reliance on large autoregressive language models, which can lead to slow token-by-token generation, and the inability to predict multiple entity types simultaneously.

Another research (Gupta et al., 2023) which aims to compare three different methods of extracting relationships from text related to rare diseases. These methods are evaluated using a dataset called RareDis, which contains information about rare diseases, focusing on cases where the diseases have complex structures. The methods assessed include traditional NER to RE pipelines, joint sequence-to-sequence models, and advanced generative pre-trained transformer models.

The researchers used a specific model called an encoder-decoder model, employing AllenNLP with a PubMedBERT encoder and LSTM decoder. This model considers relationships as a list without a specific order.

The findings of the study revealed that the traditional pipeline approach performed better than the sequence-to-sequence and generative pre-trained transformers models in extracting relationships from biomedical texts related to rare diseases. The pipeline model showed higher precision, recall, and F1-score compared to the other methods.

The study also observed common errors in the models, particularly regarding partial matches where the model predicts a part of or a larger section than the actual relationship. This issue can result in both false positives and false negatives. Additionally, the models struggled with longer object entities, often making errors in recognizing them. Even missing a single token in these longer entities could lead to errors in relationship extraction.

Another research (Hulth, n.d.) explores the use of automatic keyword assignment to summarize documents, improve information retrieval, and help navigate document collections. The study focuses on finding ways to automate this process, especially using supervised machine learning techniques. The study analyzed 2,000 English abstracts from journals in Computers and Control, and Information Technology.

To automate keyword assignment, the researchers used a rule induction model created from examples, employing recursive partitioning to maximize classification accuracy. Study built different classifiers using various term selection methods and features. Specifically, study experimented with three approaches: n-grams, noun phrases chunks, and terms matching POS tag sequences.

The findings of the research reveal the distribution of stemmed keywords in the test set, the average number of terms per document, and the variability in uncontrolled terms. The paper also compares the performance of different algorithms in terms of precision and the average number of correctly identified terms.

It highlights the challenges of existing methods in handling keywords with more than three tokens and suggests further research to improve accuracy. Another limitation identified is the lack of correlation between different POS tag feature values, which restricts the semantic categorization of terms. Additionally, relying solely on manually assigned keywords for evaluation may overlook relevant terms. Therefore, the study underscores the need for more comprehensive evaluation methods.

Another research (Payak Avinash, 2023) which aims to explain the significance and process of summarizing text and extracting keywords through natural language processing techniques. It emphasizes the importance of efficiently extracting valuable information from extensive documents or multiple sources. The focus of study lies on the methods, processes, and system architecture involved in these tasks, rather than on a particular dataset.

Techniques such as BERT and pre-trained encoders are utilized, with an emphasis on data preprocessing, keyword extraction, and text summarization algorithms. While specific results or outcomes are not provided, the document underscores the importance of these techniques for efficiently extracting useful insights from large documents. The approach involves leveraging NLP and transformer techniques to improve accuracy and reduce time in generating summaries and extracting keywords. Additionally, managing large volumes of data presents a challenge.

**2.5    Assessing the Landscape: Surveys on Text-to-Visualization Techniques**

A recent notable survey (Zhang et al., 2023b) for how natural language interfaces is used in querying and visualizing tabular data. Survey discusses the role of Large Language Models, such as ChatGPT, in systems that convert text into SQL queries and visual representations. It highlights the potential of LLMs in understanding complex user queries but also stresses the importance of customization for best results. Various evaluation metrics are mentioned for assessing the accuracy of generated queries and visualizations.

The survey paper also discussed several datasets in relation to Text-to-SQL and Text-to-Vis tasks. For Text-to-SQL, datasets like WikiSQL, Spider, KaggleDBQA, SParC, CoSQL, CHASE, and others are mentioned. In the Text-to-Vis domain, datasets such as nvBench, ChartDialogs, and Dial-NVBench are mentioned in this survey paper, focusing on pairs of natural language and visualizations.

The survey delves into how LLMs, like ChatGPT, are utilized in natural language interfaces for databases, particularly in Text-to-SQL and Text-to-Vis systems. It helps capture context, understand nuances, and generalize from limited examples. Strategies like zero-shot prompting and few-shot prompting are discussed to guide LLMs in generating accurate SQL queries.

Survey also highlighted recent approaches involving encoder-decoder models like T5 and BERT for sequence-to-sequence tasks, capturing both SQL structure and natural language nuances. In this survey paper user studies are conducted to evaluate Text-to-Vis models practically, focusing on aspects like user experience and model effectiveness. Various evaluation metrics are used to assess the accuracy of generated SQL queries, especially in multi-turn scenarios.

The survey emphasizes the importance of dataset creation, model customization, and system refinement to improve natural language interfaces for data processing. It also mentioned that developed models need to be resistant to adversarial attacks and unexpected inputs without sacrificing accuracy. Model should effectively combine known concepts to handle composite queries and adapt across different domains while understanding nuances. Survey discovers that understanding the relationships between consecutive queries in multi-turn conversations is crucial for providing accurate responses.

Also, survey highlighted that the current dataset landscape is mostly focused on English, potentially neglecting diverse global users. And deploying Large Language Models without customization may not optimize performance for specific tasks. Lastly, integrating natural language interfaces with other functionalities can present challenges in system design and performance optimization.

Another survey (Wang et al., 2022) aims to explore specifically how Machine Learning (ML) techniques can be used to enhance Data Visualization processes. It specifically looks at how ML can make visualization creation more efficient, improve how humans perceive visualizations, and tackle problems related to visualization. It discusses how ML can help in creating, interacting with, and evaluating visualizations, with the goal of improving data visualization practices.

The survey mentions several datasets such as FigureSeer, AI2D, Visually29K, DVQA, FigureQA, and Color Mapping can be used for ML4VIS tasks. ML4VIS study also focus on solving visualization challenges by using ML models like Graph-LSTM, reinforcement learning agents, and Markov models to enhance visualization techniques and user interactions.

Survey mentioned some of the challenges in ML4VIS which are as follows, the effectiveness of ML models in VIS tasks depends on the quality and quantity of training data available. If the training data is insufficient or biased, it can result in suboptimal model performance. And, ML models may struggle with complex visualization problems that have ambiguous or subjective criteria, leading to unsatisfactory outcomes and reduced user trust in ML4VIS systems. It suggests that systematic cognitive studies are necessary to understand user behaviors and expectations in ML4VIS contexts, which can help developers to improve system usability and user experience.

Another survey (Wu et al., 2022) aims to explore how artificial intelligence can be used with visualization data. It talks about the difficulties, methods, and future research opportunities in this area. The focus of this survey paper is on using different types of features together to make chart detection and classification better. It also looks for how AI can help create, improve, analyze, and assess visualizations.

It discusses combining data to improve visualization tools like DataVizard, DataSite, Profiler, Calliope, Voder, and DataShot. It talks about turning recommendation tasks into problems that can be solved, as seen in Voyager, DeepEye, and similar tools. Also, the use of VQL to turn visualizations into a formal way of representing them is seen as a good way to judge how good a visualization is. The paper also talks about breaking down visual elements into groups to better understand visual data structures. However, it points out that the way this breakdown is done might not cover everything and there could still be questions that haven't been looked at yet.

It highlights the importance of making visualizations for people and creating a bigger system where AI and people can work together. It mentions the challenges of making visualizations that both people and machines can understand easily and talks about the difficulties of finding better solutions. It also looks at the limitations of using AI alone and suggests combining human and AI approaches to make better visualizations. Finally, it suggests ways to improve the breakdown of visualizations by looking at different tasks and addressing questions that haven't been asked yet.

## 2.6 Discussion

The domain of converting natural language text into data analysis or visualization remains relatively underexplored within current academic research. Despite its potential significance, this area has received limited attention, leaving substantial room for further exploration and improvement. The lack of research activity can be attributed partly to the complexities associated with data visualization. Given that visualized data often underpins critical business decisions, inaccuracies in visualization may lead to consequential errors.

While existing studies exist in this realm, they are often constrained by limitations such as reliance on specific datasets for model training, thereby hindering their ability to generalize effectively across diverse domains. Moreover, employing Language Models for text-to-chart conversion introduces inherent noise and biases into the process.

Additionally, determining the most appropriate chart type for a given dataset can pose challenges, and models may struggle with accurately predicting the x-axis and y-axis variables. Furthermore, the correct identification of syntax for data retrieval to generate charts is prone to errors. Lastly, the absence of comprehensive, cross-domain datasets presents a significant barrier to achieving high levels of accuracy in chart generation.

## 2.7 Summary

In summary, there's not much research available on visualizing data from natural language text. Different methods have been explored, but each has its limitations. For example, Chart2Vis shows promise in predicting axis but lacks generalizability due to its small dataset. Chartllma is decent at generating charts but struggles with complex designs and focuses more on basic questions. CycleGT gets good scores but is limited by its specific datasets and small sample size. NL4DV seems accurate but lacks transparency in how its toolkit was developed. Overall, most research deals with small or specific datasets and simpler chart types, rather than tackling more complex queries.

# CHAPTER 3

# RESEARCH METHODOLOGY

## 3.1 Introduction

This chapter will discuss the research methodology proposed for data analysis and visualization of data derived from input text. Through a thorough examination of existing literature in this field, suitable approaches or models for data visualization have been identified. It has been observed that distinct pre-trained models, each designed for specific tasks, can effectively address the challenges present in current data visualization practices. These methods can subsequently be implemented sequentially to achieve the desired outcomes. Furthermore, the predictive modeling to predict cricket match outcomes based on contextual data is discussed.

The research methodology employed for transforming natural language to data visualization and their insights encompasses several crucial steps. These include the analysis of data from input file, the transformation of natural language input text into SQL queries, refinement of the SQL query using the data analysis information from the input file, identification of chart types based on either the input text or the generated output of the SQL query, followed by the visualization of the chart.

Another, research methodology employed for providing answers based on the user's natural language questions from the provided dataset using BERT model. Which consist of context-based data preparation then model training on these datasets to develop a model which can answers the user's questions from the provided input documents.

Additionally, a BERT-based classification model is employed to predict future outcomes by analysing historical data. This phase of the methodology involves creating a suitable dataset and training it using the BERT model. Ultimately, the methodology establishes evaluation metrics for all proposed models in the study.

**3.2  Dataset Selection**

The dataset utilized in this study related to cricket match details, encompassing various formats such as Test matches, One Day Internationals (ODIs), Twenty20 (T20) matches, League matches, Premier leagues, and Domestic matches for both men and women. The data is organized into folders corresponding to each match format, with each folder containing files in YAML or JSON format that document the matches specific to that format. This collection provides a rich source of information for detailed analysis and insights into the game of cricket.

The study uses Indian Premier League (IPL) dataset due to its manageable size and current relevance, given that the IPL season is ongoing. The IPL dataset comprises all matches played in the league from its inception in 2007 up to 2024. Each file within this dataset represents a single match and includes extensive details about that match. These files are structured to provide both metadata and detailed match information, ensuring a thorough representation of each game's events and outcomes.

The dataset found to be clean, and the missing values are significant, as they might indicate that a match was not played due to rain or other reasons. Therefore, the dataset is used as it is, with no pre-processing done. The study focuses on analysing the data as it is, without altering their values.

**3.3  Text2Insights: A Multi Model Architecture**

The study introduces a novel multi-model architecture, Text2Insights. As shown in Figure 3.1, this architecture operates on two primary inputs. The first input comprises tabular data formatted such as CSV, serving as a foundational database from which users can pose various inquiries concerning data visualization. The second input consists of natural language text, utilized in the creation of visual representations.

Upon receiving the tabular data, Text2Insights initiates data analysis, extracting insights from the data without altering its contents. This approach mitigates the risk of erroneous data visualization that may arise from the manipulation or deletion of missing or null values. The model's functionality depends on the nature of the input data it receives.

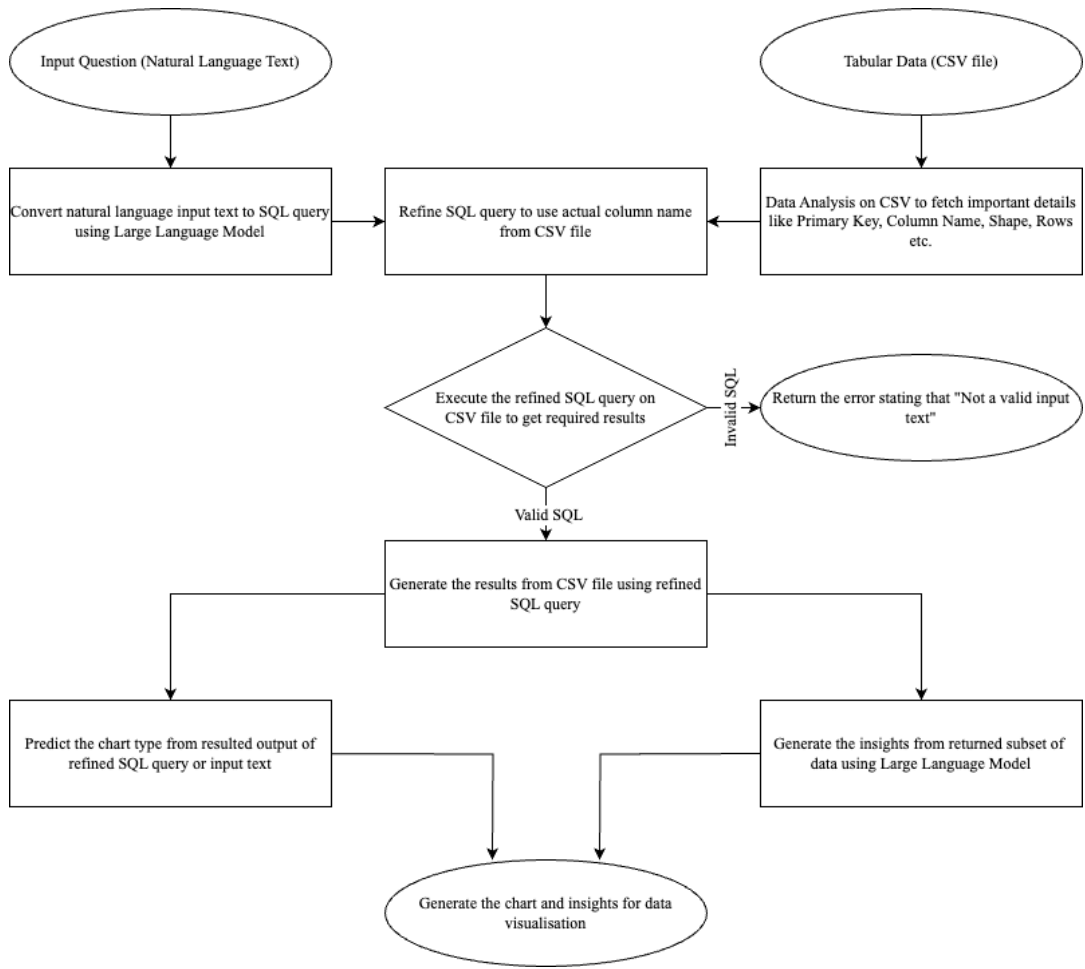

Figure 3.1: Multi-Model Architecture for Data Visualization

Figure 3.1 shows the Text2Insight model's architecture, where input text undergoes transformation into SQL statements via Large Language Model. Given SQL's efficiency in retrieving precise subsets of data from tabular formats, this conversion enables the direct execution of queries on the input file, aligning with user requirements. However, LLM-generated SQL queries may occasionally be mislabeled, necessitating refinement through Spacy's similarity model for English language to ascertain the correct keywords for accurate querying.

Upon obtaining validated SQL queries, Text2Insights executes them on the input dataset, generating the requisite data subsets for visualization. Any syntactical errors encountered during SQL execution prompt the model to request corrected input from the user. Successful execution provides data subsets used for visualization.

Furthermore, the model checks specific chart preferences outlined in user queries. If no explicit chart type is specified, the model analyses the returned data subset to determine the most suitable chart type for visualization.

To derive detailed insights from the data subset, a pre-trained Language Model is employed. This LLM processes the subset and provides comprehensive insights. Finally, the model leverages libraries such as Matplotlib or Seaborn to generate the requisite visualizations.

### 3.3.1 Comprehensive Data Analysis of Input file

The proposed methodology is designed to accept two inputs: firstly, an input file designated by the user for data visualization, and secondly, natural language text specifying the criteria for the desired data visualization. In the initial phase of input data analysis, upon receiving the input file, the model initiates an exhaustive examination, extracting important details such as the total number of columns, rows, numeric and string columns, column names, primary key column, and other relevant attributes associated with the input file. These values are systematically stored utilizing the Pandas and NumPy libraries in Python. Additionally, the dataset columns will be adjusted into their correct types. Subsequently, the input file is transformed into a relational database management system (RDMS) using the SQLite library in Python.

Importantly, this process does not fill in missing data or removing any incomplete entries. Likewise, it does not involve modifying any existing values within the dataset provided. This cautious approach is necessary changing the dataset could lead to inaccuracies in the results. Additionally, given that the dataset is not specific to any field and could originate from various domains, it would not be good idea to engage in any form of data cleaning as part of the methodology. Therefore, this study assumes that the dataset provided is sufficiently prepared for visualization purposes without the need for further preprocessing.

The shape of the dataset, indicating the number of rows and columns, can be obtained using libraries like pandas in Python. Numeric columns can be identified by performing statistical analyses to assess the distribution and characteristics of numerical data. Extracting column names provides insight into the variables present in the dataset. Identifying the primary key involves examining unique identifiers that uniquely identify each record in the dataset.

It is possible that certain columns may contain numerical or string values while their actual data type is object, thus potentially leading to errors or inaccuracies in arithmetic operations. Consequently, rectifying this discrepancy by converting such columns to their correct data types is the most effective approach for addressing these potential errors. This process involves scrutinizing the values within the dataset columns and adjusting their data types accordingly. Employing these methodologies facilitates a thorough analysis of input data.

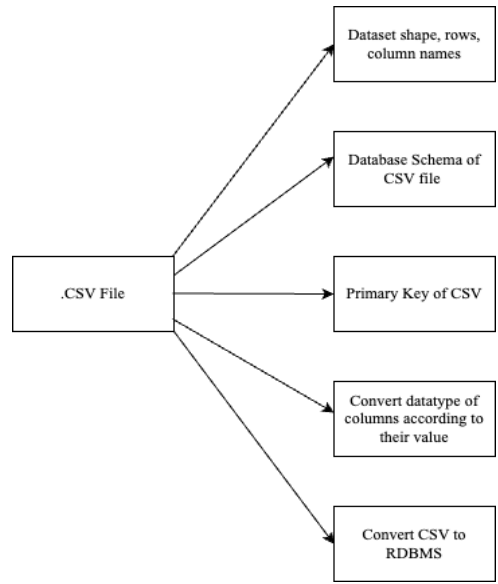

Figure 3.2: Data Analysis of input file

The schema of the input file is generated using details obtained from data analysis to ensure accurate SQL query formation by LLMs. This schema is created by using the column names, their datatypes, and the primary key.

Figure 3.2 shows the data analysis step of the input file in the Text2Insights architecture. In this step, the input file is analyzed to extract details such as the number of rows and columns, primary key, and schema. The data types are then converted to their correct forms, and finally, the input file is converted into an RDBMS.

### 3.3.2 Transformation: From Natural Language Text to SQL Query

The subsequent phase in the proposed methodology involves the transformation of the second input, consisting of natural language text requesting the desired outcome, into an SQL query utilizing an open-source pre-trained decoder only transformer-based model architecture Llama3.

The reason behind conversion of input text to SQL is imperative due to the critical nature of ensuring accuracy in the results generated by the proposed model. Given the potential impact of these results on business decision-making based on charts, the model is designed to either furnish accurate outputs or prompt the user to refine their query, thereby ensuring the provision of valid input and preventing the delivery of inaccurate results.

Pre-trained Llama3 (llama3-70b-8192) model is employed to convert natural language text into SQL queries by leveraging its ability to capture contextualized semantic representations. Initially trained on a 15 trillion parameters, pre-trained Llama3 model inherently grasp the contextual nuances and relationships between words.

Figure 3.3 shows the architecture of the SQL transformation step, where natural language text is transformed into SQL queries using the Llama3 model.

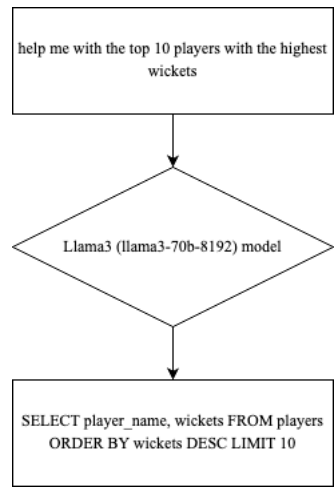

Figure 3.3: Transformation from text to SQL

### 3.3.3 Refinement of SQL Query

The subsequent stage in the proposed study involves the refinement of the generated SQL query, utilizing insights derived from the initial step of data analysis applied to the input file. This refinement process incorporates the application of Spacy's 'en_core_web_sm' model and similarity index.

The rationale behind fine-tuning the SQL query lies in its direct applicability to the converted input file turned database, enabling the provision of results. In the event of an invalid query, SQLite would raise an error, prompting the proposed methodology to guide the user towards updating their input text requirements for accuracy and coherence.

By utilizing Spacy's similarity index, proposed model can process a given text and extract the relevant keywords, which can help to fine-tune the resulted SQL query.

The output from preceding stages serves as the input for this step, where it undergoes refinement. Previous stages yield schema of input file, column names and a SQL query. In this step, the SQL query is refined by using schema of the input file as mentioned in figure 3.4. For instance, the initial SQL query may employ keywords such as 'player_name,' 'wickets,' and 'players' to execute the query on a RDBMS. However, these cannot be used directly as column names in the RDMS may differ. Therefore, the refinement of generated queries is necessary to provide accurate results.

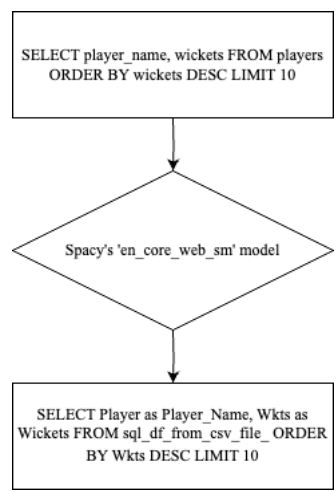

Figure 3.4: Refined SQL Output

### 3.3.4  Data Subset Generation

In the subsequent stage of the proposed methodology, the refined query is applied to the generated database to generate outcomes, thereby constituting a unique subset of the initial dataset. The SQLite library is utilized for executing SQL commands, given that the input data has been transformed into a relational database management system. This process enables the retrieval of specific information aligned with the user's predetermined criteria.

This stage provides a subset of the dataset, generated from the execution of refined SQL query derived from the previous step on the converted database. The resultant table is utilized for crafting visual representations of the data and extracting insights.

### 3.3.5  Chart Type Prediction

In the subsequent phase of the proposed methodology, the attention goes to predicting the appropriate chart type based on the resulting subset of the dataset obtained from prior steps or through user-specified input text.

Initially, the model checks if user has mentioned any types of charts they prefer, if they mentioned, the model gives those preferences more importance. If the user hasn't specified any preferences, the proposed model leverages characteristics of the derived dataset subset. This includes considerations such as the number of categorical, continuous, univariate, bivariate, multivariate, and time-series columns, as well as the presence of specific named columns. For instance, if the subset comprises two continuous columns and one categorical column, the model strategically determines the most fitting chart type. Furthermore, the model provides insights into the optimal placement of columns on the x-axis and y-axis, guided by the unique attributes of the dataset subset.

The predictive capabilities of the proposed model enhance its adaptability, ensuring that chart types align with user expectations or data characteristics. To achieve this, the proposed methodology has employed a chart predictor method. By utilizing this Chart Predictor method, the methodology gains the ability to predict specific chart types based on given input parameters.

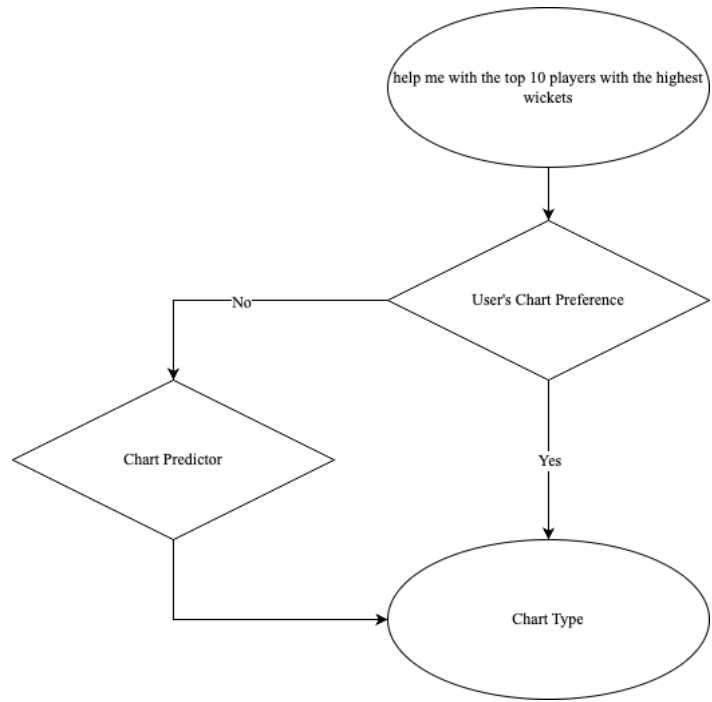

Figure 3.5: Chart Predication

As shown in figure 3.5, Chart Predictor method is used when user does not specify any specific chart type. This method takes input as subset of the dataset generated from user's query. The developed method can predict widely used ten different kinds of charts based on the data provided. These chart types are Bar chart, Box plot, Line chart, Pie chart, Scatter plot, Histogram, Area chart, Bubble chart, Radar chart and Heatmap. Also, the method predicts chart type in the same order which is mentioned here.

Box plot is returned when data is continuous and univariate, and requirement is to show the distribution of data and detect outliers. Line chart is returned when data type is continuous or time-series, with the purpose of showing trend over time. Pie chart is returned when dataset has categorical and quantitative value and requirement is to compare quantitative data across categories. Scatter plot is returned when data type is continuous and bivariate, and requirement is to show relationship between two variables.

Histogram is returned when data is of continues and univariate, and requirement is to show the data distribution. Area chart is returned when data is continuous or time-series, and requirement is to show the volume or magnitude of data over time. Bubble chart is returned when data is continuous and multivariate, and requirement is to show the relationship between more than two variables. Radar chart is returned when data is continuous and multivariate, and requirement is to show the performance of different categories across multiple dimensions. Heatmap is returned when data is continuous and multivariate, and requirement is to show the relationship between more than two variables using color intensity.

As shown in Figure 3.6, the Chart Predictor method checks the dataset. If the dataset is empty, the method returns an error stating that the provided dataset is empty. If the dataset is not empty, the method proceeds to generate a bar chart if the dataset contains both categorical and quantitative values, and if the requirement is to compare quantitative data across categories. If the method does not find the necessary categorical and quantitative columns, or if an exception is raised while creating the bar chart, the Chart Predictor method then moves on to the next chart type, which is a box plot.

If the method does not find the continuous and univariate columns, or if an exception is raised while creating the box chart, the Chart Predictor method then moves on to the next chart type, which is a Line Chart. If the method does not find the continuous and time-series columns, or if an exception is raised while creating the line chart, the Chart Predictor method then moves on to the next chart type, which is a Pie Chart.

If the method does not find the categorical and quantitative columns, or if an exception is raised while creating the pie chart, the Chart Predictor method then moves on to the next chart type, which is a Scatter Plot. If the method does not find the continuous and bivariate columns, or if an exception is raised while creating the scatter plot, the Chart Predictor method then moves on to the next chart type, which is a Histogram. If the method does not find the continuous and univariate columns, or if an exception is raised while creating the histogram, the Chart Predictor method then moves on to the next chart type, which is an Area Chart.

If the method does not find the continuous and time-series columns, or if an exception is raised while creating the area chart, the Chart Predictor method then moves on to the next chart type, which is a Bubble Chart. If the method does not find the continuous and multivariate columns, or if an exception is raised while creating the bubble chart, the Chart Predictor method then moves on to the next chart type, which is a Radar Chart.

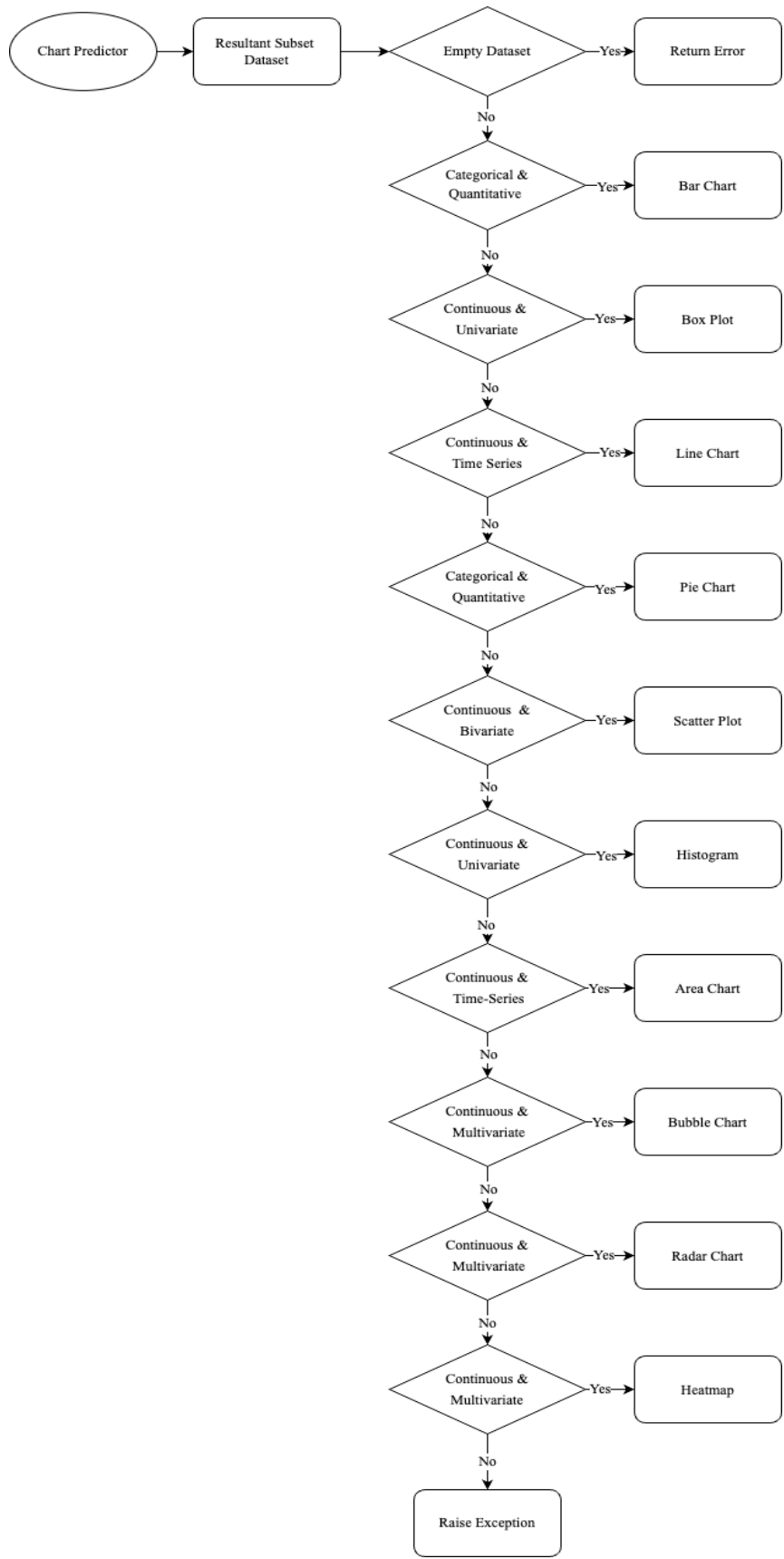

Figure 3.6: Chart Prediction's Preferences

If the method does not find the continuous and multivariate columns, or if an exception is raised while creating the radar chart, the Chart Predictor method then moves on to the next chart type, which is a Heatmap. If the method does not find the continuous and multivariate columns, or if an exception is raised while creating the heatmap. Finally, Chart Predictor method raises an exception stating that no suitable chart type could be generated from the dataset.

### 3.3.6 Chart Generation

In the subsequent phase of the proposed methodology, the generation of charts for user-oriented data visualization is undertaken, leveraging the outputs obtained from previous steps. This process is accomplished through the utilization of Python's Matplotlib and Seaborn libraries. The integration of these libraries facilitates the creation of visually informative charts, aligning with user-specified criteria and preferences as determined in earlier stages of the methodology. The utilization of established Python libraries ensures a robust and standardized approach to chart generation, enhancing the overall coherence and reliability of the data visualization outcomes within the proposed framework.

Matplotlib and Seaborn are powerful Python libraries widely employed for creating diverse and visually appealing charts and plots. Upon acquiring the complete output from the preceding stages, the chart can be generated using Matplotlib or Seaborn.

### 3.3.7 Insights Generation

The following step in proposed methodology, involves deriving insights from the generated chart. Open-source pre-trained LLM model Llama3 is employed for this step. The Llam3 'llam3-70b-8192' will analyze a subset of the data obtained in the previous step and generate insights. These insights can then be utilized by businesses to inform important decisions regarding the data. This stage aims to provide a concise summary of the insights derived from the user's data, limited to 500 words. By imposing this word limit, it has been ensured that the model does not oversimplify the data or introduce any irrelevant information.

## 3.4 Question-Answering Model

To enhance data analysis, it is beneficial to enable users to ask natural language questions based on contextual data and receive responses in natural language. This approach allows users to analyse data more effectively and intuitively. To achieve this the study also developed a document question-answering model that allows users to ask natural language questions about an input file and receive accurate answers. The model employs the 'distilbert-base-uncased' pre-trained model to correctly respond to the input queries.

The reason for using BERT is its ability to understand the context of a word in all its surroundings (bidirectionally), rather than just the context that precedes or follows it. This bidirectional understanding makes BERT particularly effective for question-answering tasks, where grasping the nuances of the context is crucial for generating accurate answers.

The study created a dataset for training the model, consisting of textual context, various types of questions, and their corresponding answers. These datasets were derived from the Indian Premier League's (IPL) cricket data. The context includes detailed information about each match, such as the teams playing, the toss winner, the match winner, the player of the match, the match location, individual player scores, and other related information. Based on this context, specific questions and answers were formulated, which were then used for model training. The created dataset is then divided into training and testing sets, with 80% of the data allocated for training and 20% reserved for testing.

The BERT model is trained on the train dataset for question-answering task to provide responses to user queries. The training parameters include a learning rate of approximately 5e-05, a batch size of 8, five epochs, an adam epsilon of 1e-06, a weight decay of 0.0, and 500 warmup steps. The remaining 20% of the dataset is used to evaluate the model's performance on accuracy, precision, recall, and F1-score.

After training the model, the study implemented a context retrieval system using TF-IDF vectorization to identify the most relevant context paragraphs for a given question. When a question is asked, the system retrieves the top matching context from the dataset, which is then fed into the BERT model to generate the answer. The developed model is useful for detailed analysis of the provided data, allowing users to ask natural language questions and receive resultant natural language responses.

## 3.5 Predictive Model

When analysing data, obtaining future predictions based on historical contextual data is invaluable. This approach provides users with insights into potential future outcomes, allowing them to take necessary actions based on these predictive insights. To achieve this, the study developed a predictive model that analyses historical contextual data to generate future outcomes in response to user queries. This model empowers users to make informed decisions by leveraging the patterns and trends present in past data.

The model utilizes Bert for Sequence Classification, employing the 'bert-base-uncased' pre-trained model. The reason behind using 'bert-base-uncased' is its exceptional performance and proficiency in sequence classification tasks like sentiment analysis and others. This pre-trained model demonstrates significantly high performance in classifying natural language into multiple labels.

The study created a dataset for training the model, consisting of textual context, and the label. This dataset is derived from the Indian Premier League's cricket data. The contextual input data for the classification model includes all matches details such as venue, participating teams, playing eleven, match dates represented in textual contextual data, and the ultimate match winner (label). The model is trained on IPL match data from its inception in 2007 up to the current year. This data is in natural language and includes the winner of each match corresponding to each context column.

The dataset is divided into an 80% training dataset and a 20% test dataset. The training dataset, comprising 80% of the data, is utilized for training the predictive model. Subsequently, the remaining 20% of the data is reserved for evaluating the performance of the developed model. The historical data includes information on ten different cricket teams currently playing Indian Premier League. However, each record only contains details about the two teams playing in a specific match.

The study has developed two different approaches: binary classification and multiclass classification. Binary classification predicts the match outcome with two labels: Team 1 and Team 2. Multiclass classification uses ten different labels, each representing one of the ten teams currently playing in the IPL.

### 3.5.1 Binary Classification

The binary classification model uses BERT for sequence classification with two labels for the teams playing the match. If Team 1 wins, the label is 0; if Team 2 wins, the label is 1. Team 1 and Team 2 are determined by their order of appearance in the context: the first team mentioned is Team 1, and the second team is Team 2. Therefore, if the first mentioned team wins, the label is 0; otherwise, it is 1.

The training data, which comprises 80% of the total dataset, is used to train the binary classification model. The training parameters include a learning rate of approximately 3.5e-5, a batch size of 8, five epochs, a weight decay of 3.8e-5, and 466 warmup steps. The remaining 20% of the dataset is used to evaluate the model's performance on accuracy, precision, recall, and F1-score.

### 3.5.2 Multi-Class Classification

The multiclass classification model uses BERT for sequence classification with ten different labels, each representing one of the ten teams currently playing in the IPL. The labels are as follows: SRH-0, CSK-1, DC-2, MI-3, KXIP-4, RCB-5, GT-6, KKR-7, RR-8, and LSG-9. The model predicts the results based on these labels, assigning the highest weight to the label derived from the context.

The training data, comprising 80% of the total dataset, is used to train the multiclass classification model. The training parameters include a learning rate of approximately 1.43e-5, a batch size of 16, four epochs, a weight decay of 0.02, and 447 warmup steps. The remaining 20% of the dataset is used to evaluate the model's performance on accuracy, precision, recall, and F1-score.

## 3.6  Model Evaluation Matric

The concluding step in the research methodology involves the comprehensive evaluation of the employed machine learning models. As mentioned in the proposed methodology, various types of models are employed within the study framework. Each of these models is carefully evaluated, examining various factors to assess how well they perform and how effective they are. This thorough evaluation process guarantees a rigorous assessment of the models' abilities, providing the study with detailed insights into their strengths and weaknesses.

Initially, the Llama3 model is used to translate natural language text into SQL queries. The generated SQL queries are evaluated for both syntactical correctness and overall accuracy. Syntactical evaluation, conducted using a SQL parser, checks how well the SQL query adheres to syntax rules. The correctness of the SQL query is analysed using the BLEU score, which measures how closely the generated query matches the actual SQL queries. The BLEU score proves instrumental in assessing the quality of SQL query generation, as it involves comparing the n-grams of the generated query with those of the reference query. This methodological approach enhances precision in evaluating the effectiveness of the Llama3 model's ability to seamlessly transform natural language inputs into coherent and accurate SQL queries.

BLEU score can be calculated with the formula (3,1):

$$BLEU = BP \times e^{(1/n \sum_{i=1}^{n} \log(precesion\ i))} \qquad (3,1)$$

Subsequently, a second model is used to refine SQL queries through Spacy's Named Entity Recognition model. This step refines the SQL query generated by Llama3. The refined query undergoes the same syntactical evaluation and BLEU score evaluation as the initial query.

In addition to the internal evaluation of individual models, the study has conducted a holistic assessment of the entire data visualization model. This comprehensive evaluation process involves initially gathering sample inputs from multiple users for cricket dataset. Subsequently, a small dataset is curated using these inputs, and charts are manually generated for each sample input text. This sample dataset, not previously encountered by the model during its training phase, will then be inputted into the developed model for evaluation.

Accuracy, Precision, Recall and F1-Score, is calculated by the formula (3,2), (3,3), (3,4) and (3,5):

TP =Number of True Positive Predictions
TN =Number of True Negative Predictions

FP = Number of False Positive Predictions

FN = Number of False Negative Predictions

$$Accuracy = \frac{TP+TN}{(TP+FP+TN+FN)} \qquad (3, 2)$$

$$Precision = \frac{TP}{(TP+FP)} \qquad (3, 3)$$

$$Recall = \frac{TP}{(TP+FN)} \qquad (3, 4)$$

$$F1 - Score = \frac{2*Precision*Recall}{(Precision+Recall)} \qquad (3, 5)$$

The question-answering model is evaluated using the created 20% test dataset. Similarly, the predictive model is also evaluated on the 20% test dataset.

## 3.7 Summary

This study introduces a novel approach Text2Insights, for data visualization within a multi-model framework. This approach adheres to a sequential pipeline structure wherein the output of one phase serves as the input for the subsequent phase. The development of this architecture is done by a comprehensive analysis of existing research gaps and constraints within the data visualization domain. The choice of the models depends on its ability to provide users with precise data visualization or to indicate errors. It is imperative that the introduced model avoids generating erroneous visualizations, as such inaccuracies could significantly impact crucial business decisions.

The introduced framework requires two inputs: an input query and a dataset corresponding to the user's query. Initially, the model conducts a comprehensive analysis of the supplied dataset, yielding essential outputs such as the dataset's row count, shape, column names, and primary key, which are pivotal for subsequent procedures. Parallelly, the second query input is inputted into another model for processing. Specifically, the utilized model in this stage is pre-trained

llama3 model, designed to process natural language text inputs and generate SQL queries as outputs.

Given the potential for Llama3 Model to introduce noise into the output, relying on the generated SQL queries as valid may not be advisable. Variations in column names within the CSV dataset could lead to errors during SQL execution. Hence, an alternative spacy's similarity index model will be employed to address this issue. Through the word similarity approach of 'en_core_web_sm' model, words representing column names will be aligned with those generated by the SQL, enabling refinement of the SQL query accordingly. This refined SQL query can then be directly applied to the CSV dataset.

Parallelly, the input query will be examined to determine if any preferences regarding chart types are specified. In cases where the user specifies a particular chart type, the model will prioritize generating that specific type. Conversely, if no specific chart type is stipulated, the process proceeds to the subsequent stage, wherein the refined SQL query is executed on the dataset. This stage facilitates the extraction of a user-specific subset of data. Subsequently, if the user has not indicated a preferred chart type, the chart type will be inferred based on the characteristics of the resulting subset data. To achieve this, an additional Chart Predictor method is employed to predict the appropriate chart type, considering the data types present in the user-specific subset.

The derived subset of data will serve as the basis for generating charts to visualize the data, employing various Python libraries such as Matplotlib and Seaborn. Simultaneously, Large Language Model Llama3 will be used to get insights regarding the user's specified data requirements.

Additionally, the study has developed a data analysis model for the input data, comprising two different types of analysis models. The first model is designed to provide answers based on the input file for contextual questions from the user. The second developed model focuses on predicting future outcomes based on the available historical data.

The study created a contextual dataset from the available IPL data, which is utilized to train both data analysis models. The question-answering model takes natural language input text from the user and provides answers based on all available contextual data.

The predictive model that predicts the outcome of cricket matches from natural language text using a BERT sequence classification model. The model is trained on context-based historical data of all IPL matches. Two approaches were used to predict match outcomes: binary classification and multiclass classification.

In binary classification, the model assigns two labels to the context—Team 1 and Team 2—and predicts the winner between these two teams. In multiclass classification, the model assigns ten different labels, each representing one of the ten teams currently participating in the IPL, and predicts the winner from these ten labels.

Finally, different evaluation metrics have been defined for each model, which will undergo individual evaluation. The collective performance of all the models is assessed using various parameters and metrics.

# CHAPTER 4

# ANALYSIS AND EXPERIEMNTS

## 4.1  Introduction

The chapter will provide an in-depth examination of the Indian Premier League dataset, data preparation, covering a range of statistical, exploratory, time series, and predictive analyses. This chapter offers an understanding of the various facets of IPL matches, teams, and player performances over the seasons from 2007 to 2024.

Preparing datasets involves understanding the YAML file and then converting it to JSON and CSV formats for better data analysis. The context-based dataset for the document question-answering model includes context, questions, and their respective answers. Another dataset, prepared for predictive modelling, contains context with match information and the winner of each match.

The statistical analysis, where insights like the total number of matches played, the teams involved, and the venues used are presented. Following this, the exploratory data analysis section discusses about team performances, revealing critical statistics like match-winning percentages and toss-winning outcomes. This analysis helps to identify patterns and trends that distinguish the successful teams from the rest, providing a detailed understanding of what factors contribute to a team's success.

Then the study progresses to time series analysis, which examines team performance across different IPL seasons. This section highlights how teams have improved or declined over time, offering insights into their scoring trends and consistency. The team and player performance analysis sheds light on individual and team achievements. By focusing on detailed metrics such as run totals and player contributions, this section emphasizes the importance of consistent performance and highlights standout players who have significantly impacted their teams' fortunes.

Predictive analysis demonstrates how the predicted match outcome evolves with each ball, depending on how the team plays ball by ball and their forecasted total runs.

The experiments conducted for the document question-answering model are discussed, focusing on the different hyperparameters used for training the initial and final models through hyperparameter tuning. Additionally, experiments for predictive modelling to forecast future outcomes are also discussed, detailing the various sets of hyperparameters tuned for optimal performance. Utilizing models like BERT for such tasks, this demonstrates how historical data can be leveraged to predict future performances.

Overall, the study provides a thorough exploration of the IPL dataset, offering valuable insights and practical applications. Provides a holistic understanding of the league's dynamics, the factors influencing team and player success, and the potential of predictive analytics in cricket.

## 4.2  Dataset Description

The dataset utilized in this study related to cricket match details, encompassing various formats such as Test matches, One Day Internationals (ODIs), Twenty20 (T20) matches, League matches, Premier leagues, and Domestic matches for both men and women. The data is organized into folders corresponding to each match format, with each folder containing files in YAML or JSON format that document the matches specific to that format. This collection provides a rich source of information for detailed analysis and insights into the game of cricket.

For the purposes of analysis, the Indian Premier League dataset has been used due to its manageable size and current relevance, given that the IPL season is ongoing. The IPL dataset comprises all matches played in the league from its inception in 2007 up to 2024. Each file within this dataset represents a single match and includes extensive details about that match. These files are structured to provide both metadata and detailed match information, ensuring a thorough representation of each game's events and outcomes.

The metadata within each file includes essential information such as the creation date of the file and its version, ensuring that the dataset's integrity and updates can be tracked. The "info" object provides comprehensive details about each match, including the date it was played, event specifics, the winning team and the margin of victory, the player of the match, participating teams, playing elevens for both teams, the season in which the match occurred, the toss winner and their decision, the city and venue of the match, and the names of officials such as umpires and referees.

Furthermore, the "innings" object captures the sequential play details, specifying which team batted in each inning. For both innings, the dataset provides ball-by-ball details, including the striker, non-striker, bowler, and the outcome of each delivery. This includes runs scored, whether from the bat or as extras such as wides or no-balls, and details of wickets taken, including the method of dismissal (e.g., caught, bowled, run out). For the second innings, the target runs to be achieved and the number of overs available are also documented. This granular level of detail is invaluable for analyzing player performance, game strategies, and match outcomes.

By focusing on the IPL dataset, this study leverages a well-structured and detailed collection of match data to explore various analytical dimensions, providing insights into player performances, team strategies, and the overall dynamics of the league. The dataset's comprehensive nature ensures that all critical aspects of the matches are covered, facilitating a robust analysis of the IPL's rich history.

## 4.3   Dataset Preparation

For analyzing the cricket dataset, the necessary data is extracted and stored in CSV files, enabling conversion into a panda Data Frame for further analysis. Three primary datasets are created from the IPL directory to facilitate different analytical techniques such as statistical analysis, exploratory data analysis, time series analysis, and predictive analysis.

The first dataset comprises all the innings-related data, crucial for detailed statistical and exploratory analysis. This dataset includes columns such as the match ID, creation date, city, match dates, outcome details including the winner, margin of victory by runs or wickets, player of the match, participating teams, season, toss winner and decision, venue, team batting, over number, ball number, batter, bowler, non-striker, runs scored by the batter, extra runs, total runs, player out, and the type of dismissal. This dataset provides a comprehensive view of each innings, capturing essential match details and individual performance metrics. It supports various types of analysis, including identifying trends over seasons, comparing team performances, and evaluating player contributions. The detailed ball-by-ball data allows for granular analysis, enabling insights into game dynamics and strategies.

The second dataset focuses on the contextual details of each match, including the match ID, context description, questions formatted from the context, and their answers. An example of the context description is:

*"Cricket match played on 2024-04-09 at city Mohali between Sunrisers Hyderabad (SH) and Punjab Kings (PK), toss is won by Punjab Kings, and they have decided to field Winner of the match is Sunrisers Hyderabad they won by 2 runs, and Player of the match is Nithish Kumar Reddy. First Inning is played by Sunrisers Hyderabad, and they have set the target of 183. First Inning batsman has scored runs as follows: TM Head has scored 21 runs, Abhishek Sharma has scored 16 runs, AK Markram has scored 0 runs, Nithish Kumar Reddy has scored 64 runs, RA Tripathi has scored 11 runs, H Klaasen has scored 9 runs, Abdul Samad has scored 25 runs, Shahbaz Ahmed has scored 14 runs, PJ Cummins has scored 3 runs, B Kumar has scored 6 runs, JD Unadkat has scored 6 runs, . Second Inning batsman has scored runs as follows: S Dhawan has scored 14 runs, JM Bairstow has scored 0 runs, P Simran Singh has scored 4 runs, SM Curran has scored 29 runs, Sikandar Raza has scored 28 runs, Shashank Singh has scored 46 runs, JM Sharma has scored 19 runs, Ashutosh Sharma has scored 33 runs."*

This dataset is specifically designed for the document question-answering model, aimed at providing answers to user questions based on historical data. By using context-based information, the model can provide specific answers to the user's input questions.

The last dataset focuses on the contextual details of each match, which includes the match ID, context description, and match winner. An example of the context description is:

*"Cricket match played on city Delhi on 2008-05-17 in the stadium Feroz Shah Kotla between Delhi Daredevils (DD) and Kings XI Punjab (KXP), toss is won by Delhi Daredevils and they have decided to bat. Players for Delhi Daredevils (DD) are G Gambhir, V Sehwag, S Dhawan, TM Dilshan, MF Maharoof, Shoaib Malik, PJ Sangwan, KD Karthik, A Mishra, R Bhatia, GD McGrath and players for Kings XI Punjab (KXP) are SE Marsh, JR Hopes, Yuvraj Singh, LA*

*Pomersbach, DPMD Jayawardene, IK Pathan, RR Powar, PP Chawla, U Kaul, S Sreesanth, VRV Singh."*

This dataset is specifically designed for predictive analysis, allowing the determination of match outcomes based on natural language queries. By using this context-rich information, predictive models can be trained to forecast match results, enhancing the understanding of factors that influence game outcomes. This approach leverages the detailed narrative of each match, providing a robust foundation for predictive analytics in cricket.

By organizing the data into these three datasets, the study ensures a structured and comprehensive approach to analyzing the IPL matches. This method not only facilitates various analytical techniques but also provides a clear pathway for deriving actionable insights from the rich cricket data.

## 4.4    Data Analysis

Data analysis focuses on statistical analysis, exploratory data analysis, time series analysis, and predictive analysis. Statistical analysis has identified the total number of matches played, the teams involved, and the venues used. Exploratory data analysis discovered team performances, critical statistics like match-winning percentages and toss-winning outcomes. Time series analysis examines team performance across different IPL seasons, highlighting how teams have improved or declined over time. Predictive analysis section introduces sophisticated techniques to forecast match outcomes and team scores. By leveraging historical data and advanced models BERT for such tasks.

### 4.4.1    Statistical Analysis

Statistical analysis is conducted on the first dataset, which contains innings-related data. This analysis focuses on several key aspects of the Indian Premier League, providing insights into various dimensions of the league's matches and team performances.

It is observed from analysis that there are a total of 1,093 matches played in the IPL as per the dataset available, starting from 2007 and continuing annually through 2024, with no interruptions even during the COVID-19 pandemic. Additionally, 12 different teams have participated in the league, and the matches have been hosted in 34 different cities across 58 different venues. That shows the popularity and distribution of matches across different locations and league expansion over the years.

From Figure 4.1, it can be seen that most of the matches (182) are played in Mumbai, followed by Bangalore (94), Kolkata (93), Delhi (90), Chennai (83), and Hyderabad (77). Additionally, it is observed that the majority of the matches are played in India, although a few matches have been held outside India in countries such as South Africa, the UAE, and the UK.

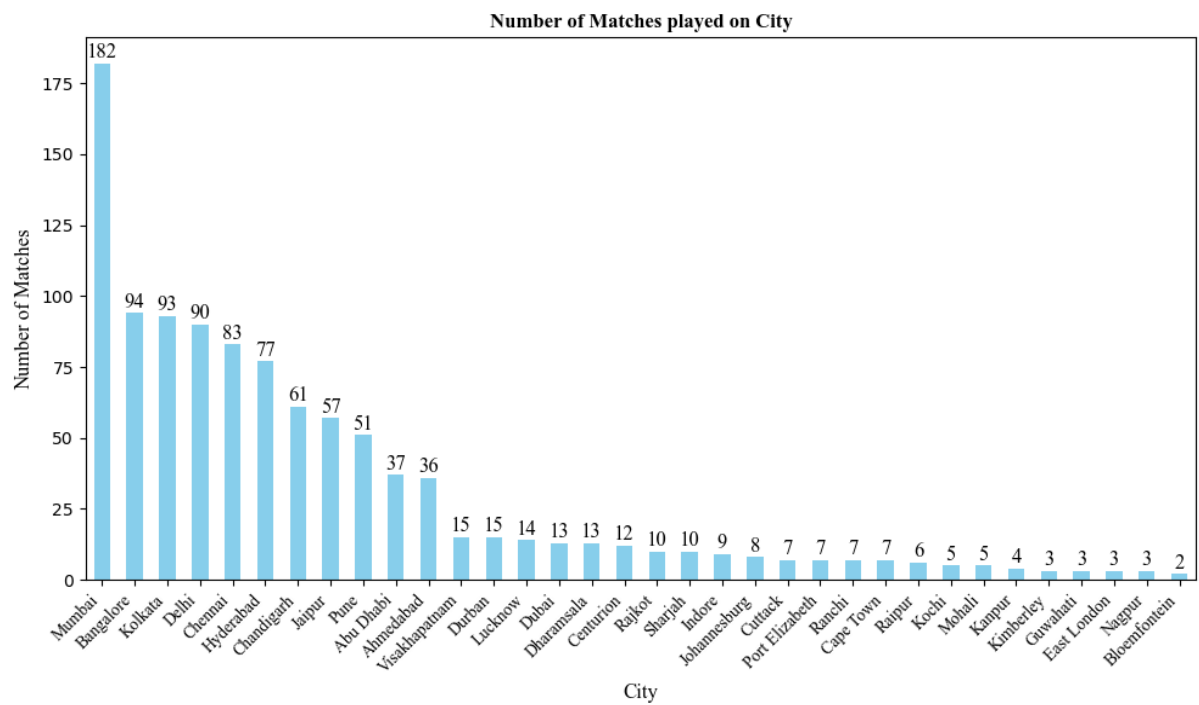

Figure 4.1: Matches Played on Different City

As shown in Figure 4.2, there are 58 different venues where IPL matches have been played. Eden Gardens (77) and Wankhede Stadium (73) have hosted the most matches. In contrast, stadiums outside India have hosted significantly fewer IPL matches.

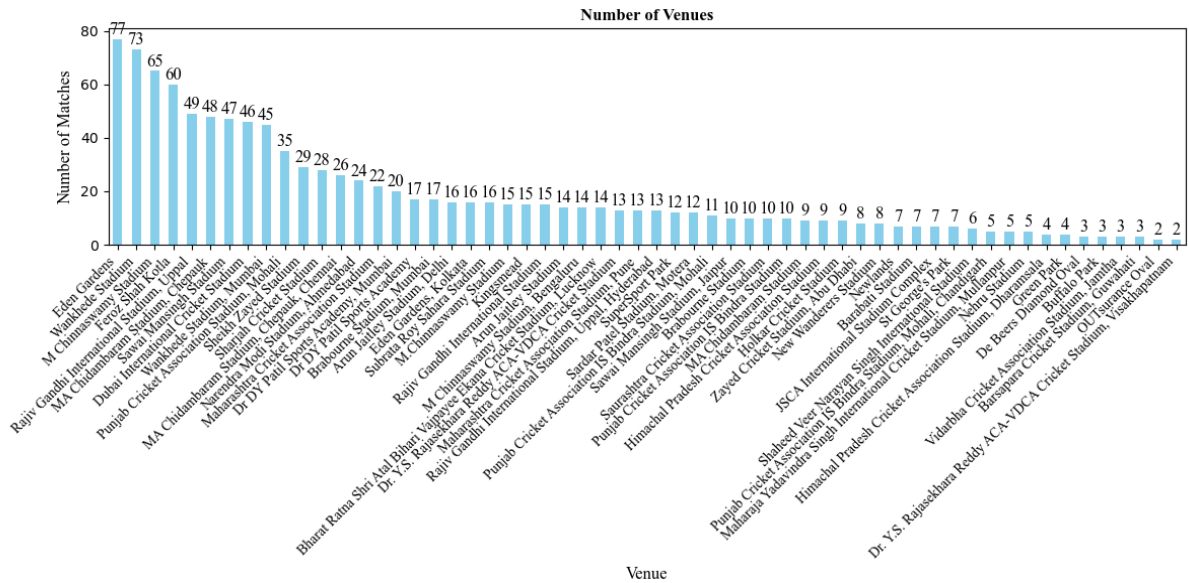

Figure 4.2: Number of Venues

When evaluating players' performances, it is best to examine who has won the Player of the Match awards most frequently. Winning the Player of the Match indicates consistent high performance throughout the season. Analyzing different Player of the Match winners reveals that 290 different players have received this award, signifying that not only a set group of players perform well, but various players excel at different times, often leading their teams to victory.

As per Figure 4.3, it appears that AB de Villiers (25), Chris Gayle (22), Rohit Sharma (19), Virat Kohli (18), and David Warner (18) top the leaderboard for the most Player of the Match awards. Their significant performances over the years highlight their crucial contributions to their teams' successes.

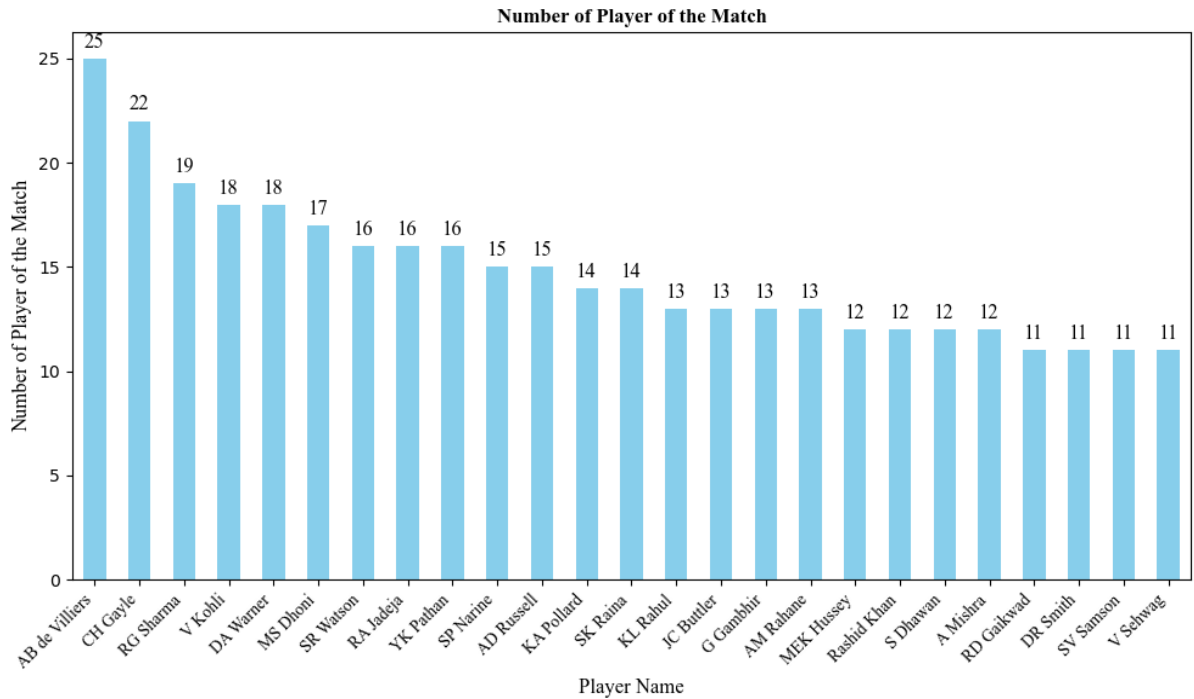

Figure 4.3: Player of the Match

### 4.4.2 Exploratory Data Analysis

Exploratory Data Analysis is also conducted on the first dataset, which contains innings-related data. Providing insights into team performance, match-winning percentages, and toss-winning percentages. EDA helps in understanding patterns and trends within the data, revealing which teams consistently perform well, their likelihood of winning matches based on various factors, and how often teams win the toss.

As shown in Figure 4.4, the Mumbai Indians have won the most matches (142), followed by the Chennai Super Kings (138), Kolkata Knight Riders (129), and Royal Challengers Bangalore (121). The Kochi Tuskers Kerala have the least number of wins (6) as they have not participated in all the IPL seasons.

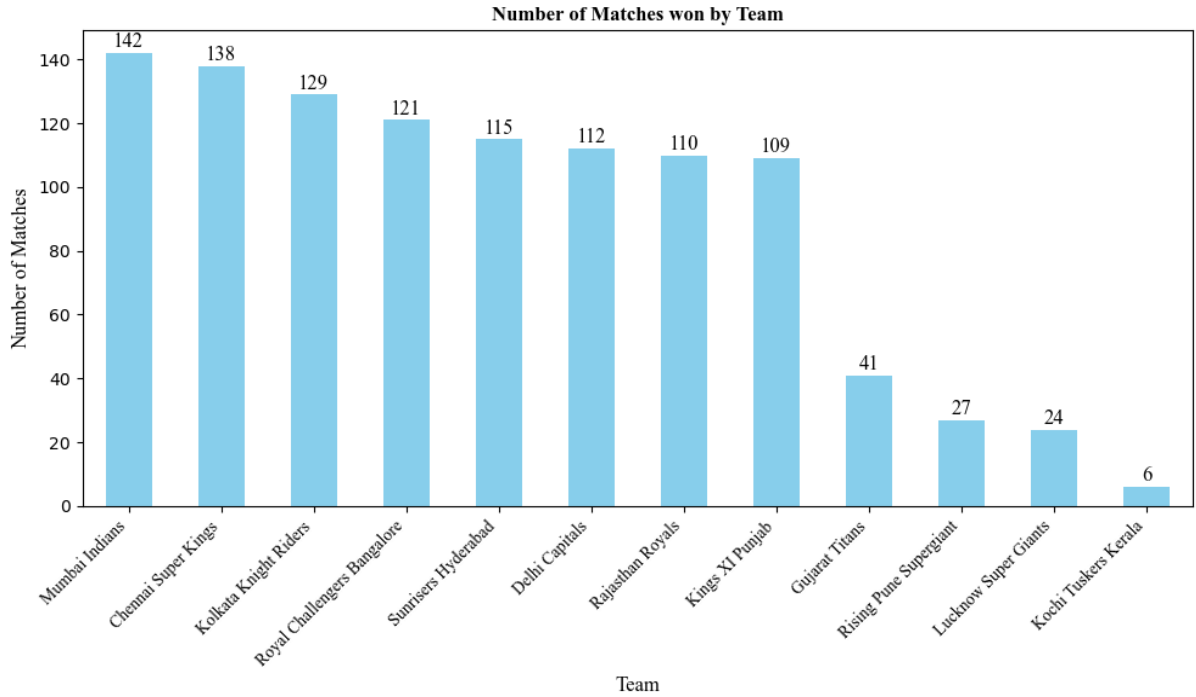

Figure 4.4: Matches won by each Team

Though Figure 4.4 shows that the Mumbai Indians have won the most matches, Figure 4.5 reveals that their win percentage is at 54.4%, placing them fourth among all teams. This is still a strong performance, although the team with the highest win percentage, Chennai Super Kings, stands at 58.2%. Notably, the second and third teams in terms of win percentage, Gujarat Titans and Lucknow Super Giants, have played roughly one-third of the number of matches that the Mumbai Indians have played.

However, it appears that there is no significant difference between the teams in terms of winning matches, as each team has a win percentage close to 50%. This indicates that all the teams are highly competitive, making it difficult to predict the outcome of any given match, as both teams have nearly equal chances of winning.

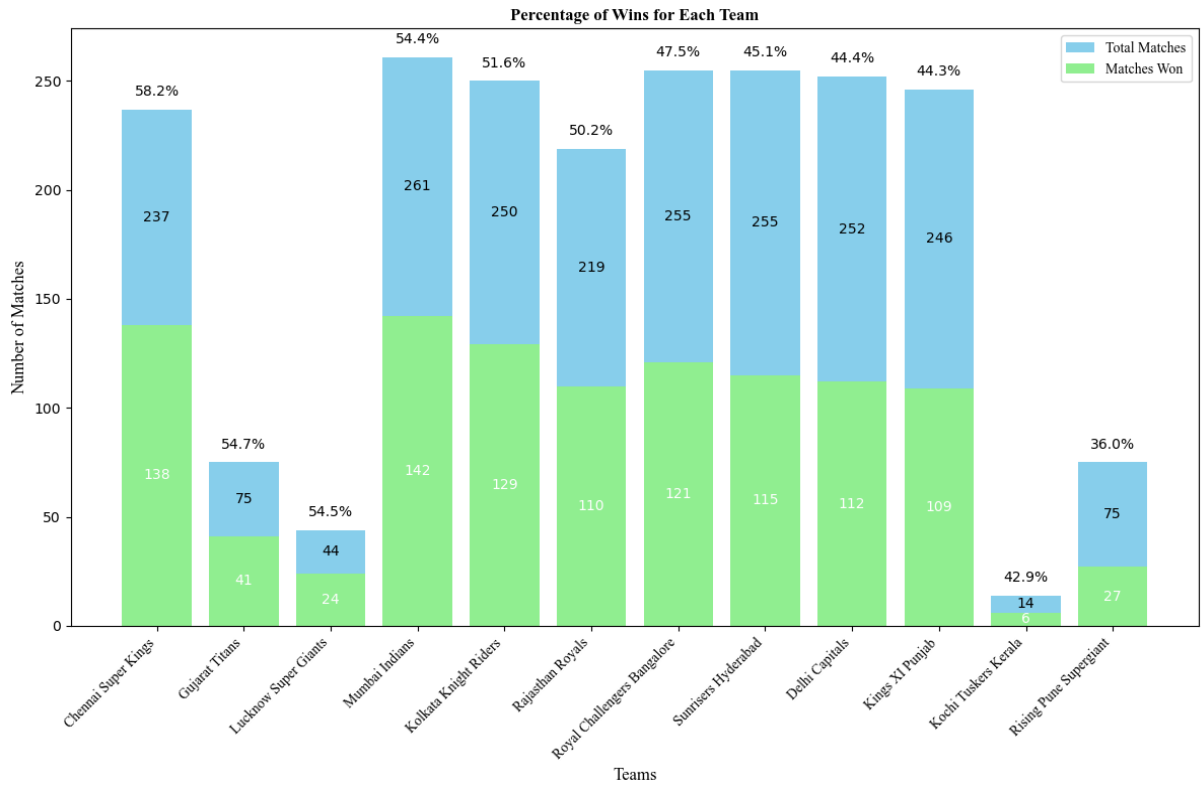

Figure 4.5: Winning Percentage of each Team

It is shown in Figure 4.6 that the Mumbai Indians rank first on the leaderboard in terms of winning the toss.

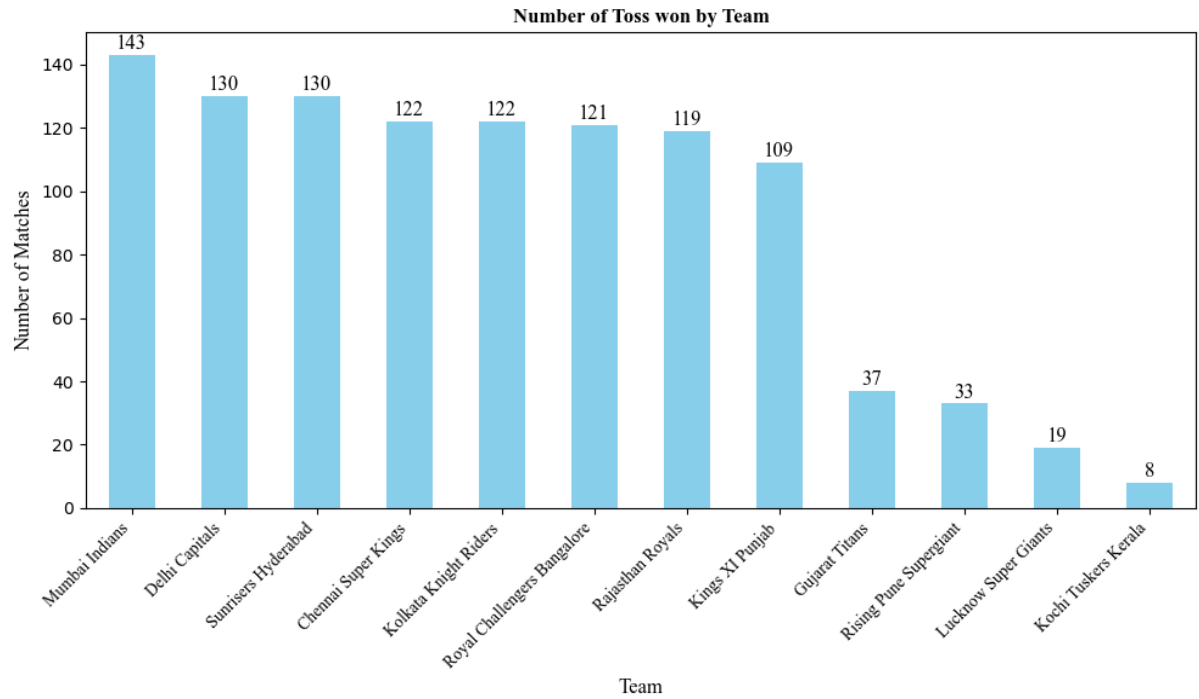

Figure 4.6: Toss Won by each Team

As shown in Figure 4.7, it can be observed that Kochi Tuskers Kerala has won the most tosses; however, they have played only 14 matches, making this data less relevant. Therefore, Mumbai Indians have the highest toss win percentage at 54.8%, followed by Rajasthan Royals at 54.3%. Additionally, it appears that there is no significant difference between the teams in terms of winning the toss, as every team has a toss win percentage close to 50%.

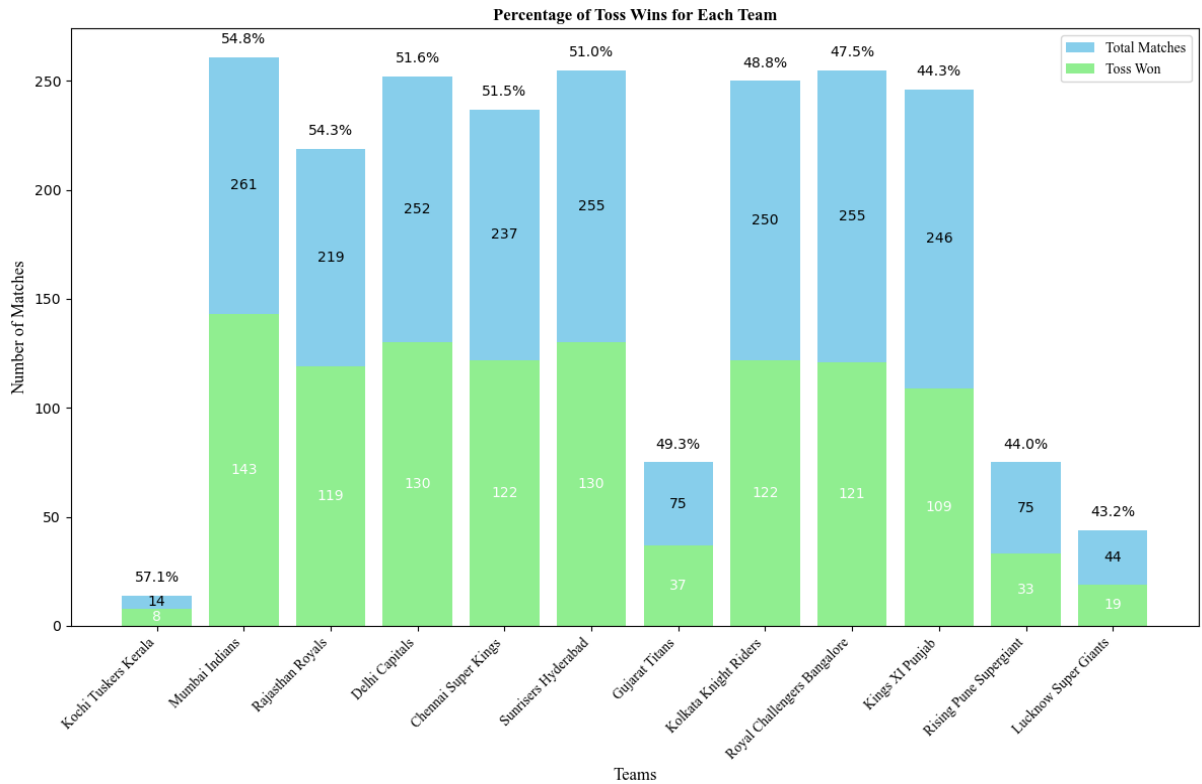

Figure 4.7: Toss Win Percentage by each Team

As shown in Figure 4.8, the total matches won by teams when they won the toss is 548, while the matches won by teams when they lost the toss is 545. This similarity indicates that winning the toss does not significantly impact the likelihood of winning the match. Therefore, it cannot be said that a team will necessarily win if they win the toss or lose if they lose the toss.

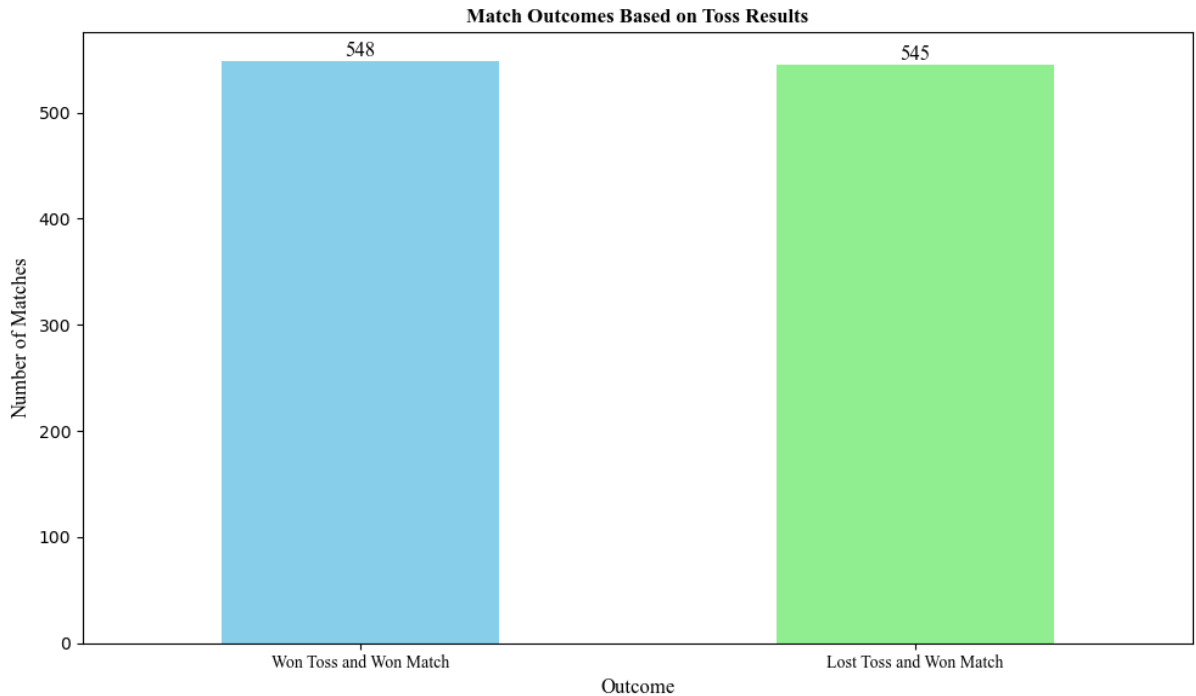

Figure 4.8: Match Outcome Based on Toss Result

However, it is insightful to examine team-wise match outcomes based on toss results. From Figure 4.9, it appears that Chennai Super Kings, Mumbai Indians, Delhi Capitals, Gujarat Titans, and Rajasthan Royals have significantly higher numbers of matches won when they win the toss. This observation suggests that these teams are particularly adept at assessing playing conditions and making strategic decisions, such as choosing to bat or field, after winning the toss.

Though Royal Challengers Bangalore and Sunrisers Hyderabad have nearly the same winning results whether they win or lose the toss, it indicates that these teams are highly competitive and that the toss decision does not significantly affect their chances of winning a match. They perform equally well whether they are chasing a target or defending a score.

On the other hand, Kings XI Punjab, Lucknow Super Giants, and Rising Pune Supergiants have a higher probability of winning matches when they lose the toss. This suggests that these teams may not effectively analyze the playing and opposition conditions to make optimal decisions when they win the toss.

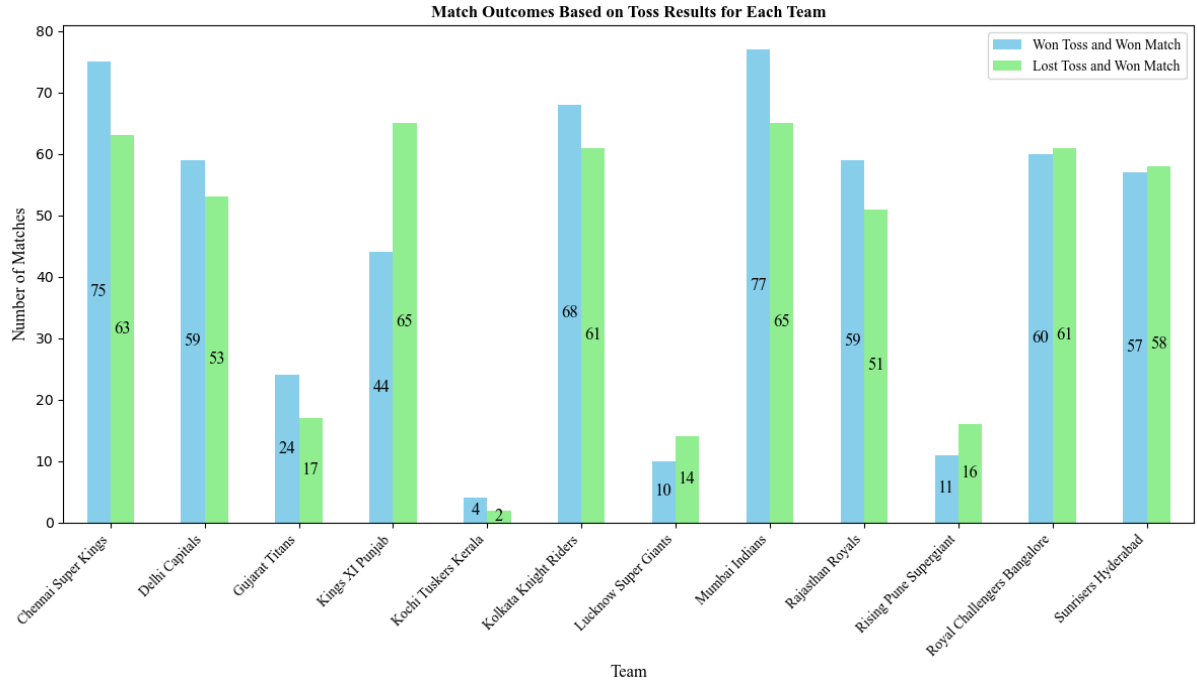

Figure 4.9: Winning matches based on Toss Outcome

As per Figure 4.10, it appears that only the Chennai Super Kings decide to bat or field almost equally when they win the toss. In contrast, every other team tends to prefer batting first to set a target and then defend it.

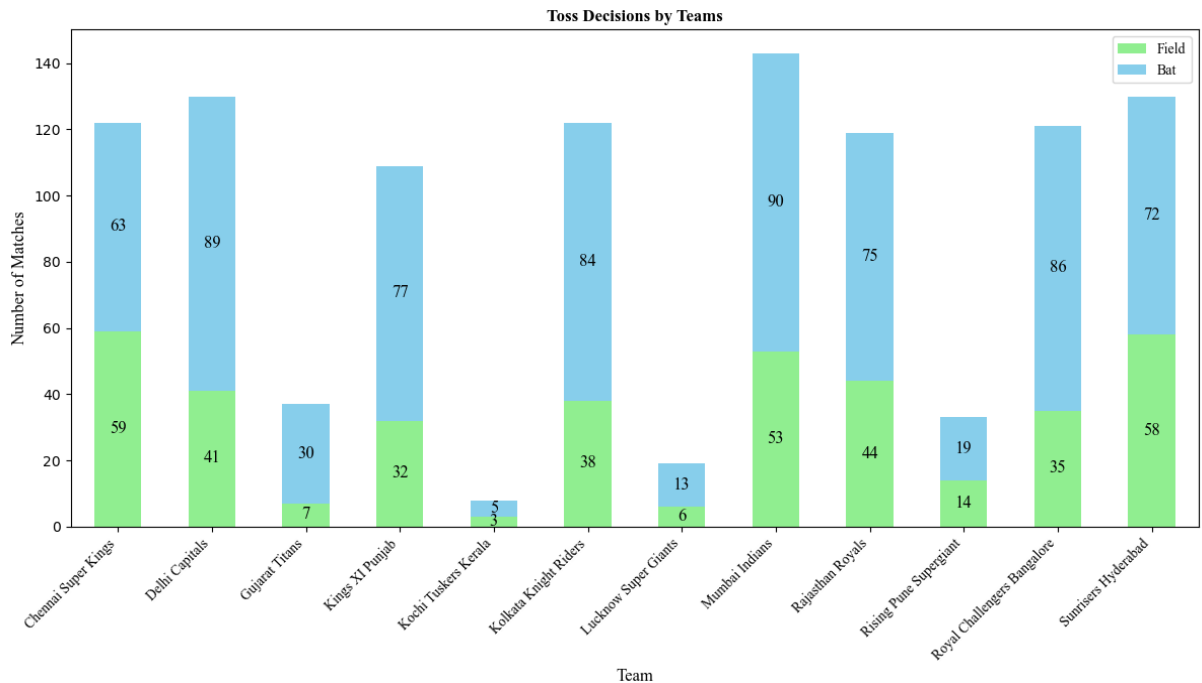

Figure 4.10: Toss Decisions by Teams

Analyzing ball-by-ball scoring runs for each team is beneficial as it identifies how each team approaches the match, highlighting their aggressiveness in scoring runs. The average run rate per ball is examined for all the matches each team has played. As per Figure 4.11, it is evident that all teams start well, typically reaching close to 50 runs within the first five overs. Some teams gradually increase their run rate, going beyond 150 runs within 15 overs, while others find it challenging to surpass 150 runs within the same period and end up scoring less than 200 runs on average within 20 overs.

Teams like Kolkata Knight Riders (KKR) and Royal Challengers Bangalore (RCB) are notable for increasing their run rate consistently from start to finish. In contrast, Delhi Capitals (DC), Lucknow Super Giants (LSG), Mumbai Indians (MI), Rajasthan Royals (RR), Chennai Super Kings (CSK), and Sunrisers Hyderabad (SRH) tend to maintain a steady run rate throughout their innings. On the other hand, Rising Pune Supergiants (RPS) and Kochi Tuskers Kerala (KTK) struggle to start well and fail to score close to 200 runs by the end of their innings. This analysis shows that a strong start is crucial for achieving a high total score.

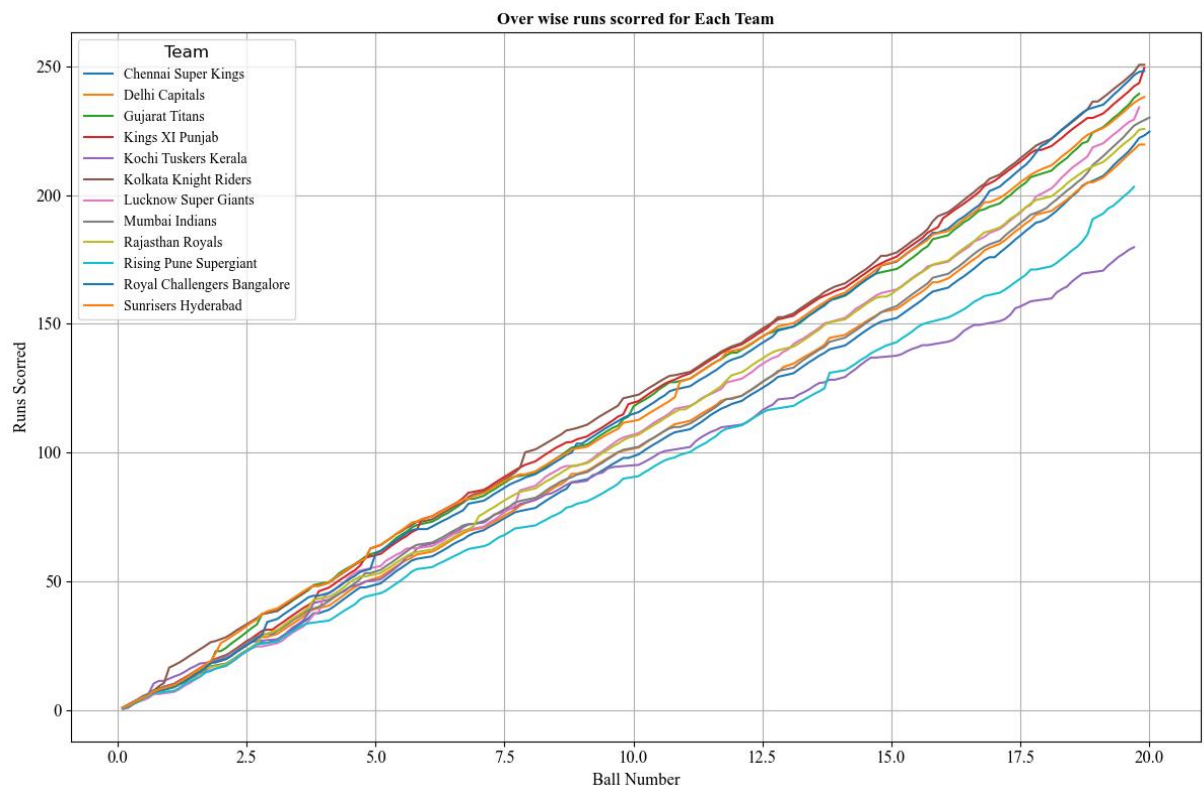

Figure 4.11: Over wise runs scored for each Team

Over-wise analysis can also be visualized using a heatmap, as shown in Figure 4.12, which illustrates how each team performs during different overs. It can be seen that Chennai Super Kings, Gujarat Titans, and Kings XI Punjab tend to start slowly, while Delhi Capitals, Kolkata Knight Riders, Mumbai Indians, Royal Challengers Bangalore, and Sunrisers Hyderabad start aggressively in scoring runs. The scoring rate from the 7th to the 10th over is generally steady and slow for almost every team. From the 11th to the 15th over, some teams maintain a steady run rate, while others increase their scoring during these overs.

Most teams begin to accelerate their run rate significantly during the death overs, from the 16th to the 20th over. KKR and SRH exhibit strong run rates in the starting overs (1-5), showcasing their powerful opening batsmen. CSK and KKR have the highest run rates in the 19th and 20th overs, indicating their strong death-over batting prowess. Conversely, RCB has the highest run rate in the 16th over, highlighting their robust middle-over batting capabilities. This heatmap analysis provides a detailed view of each team's scoring strategies and performance trends across different phases of the match.

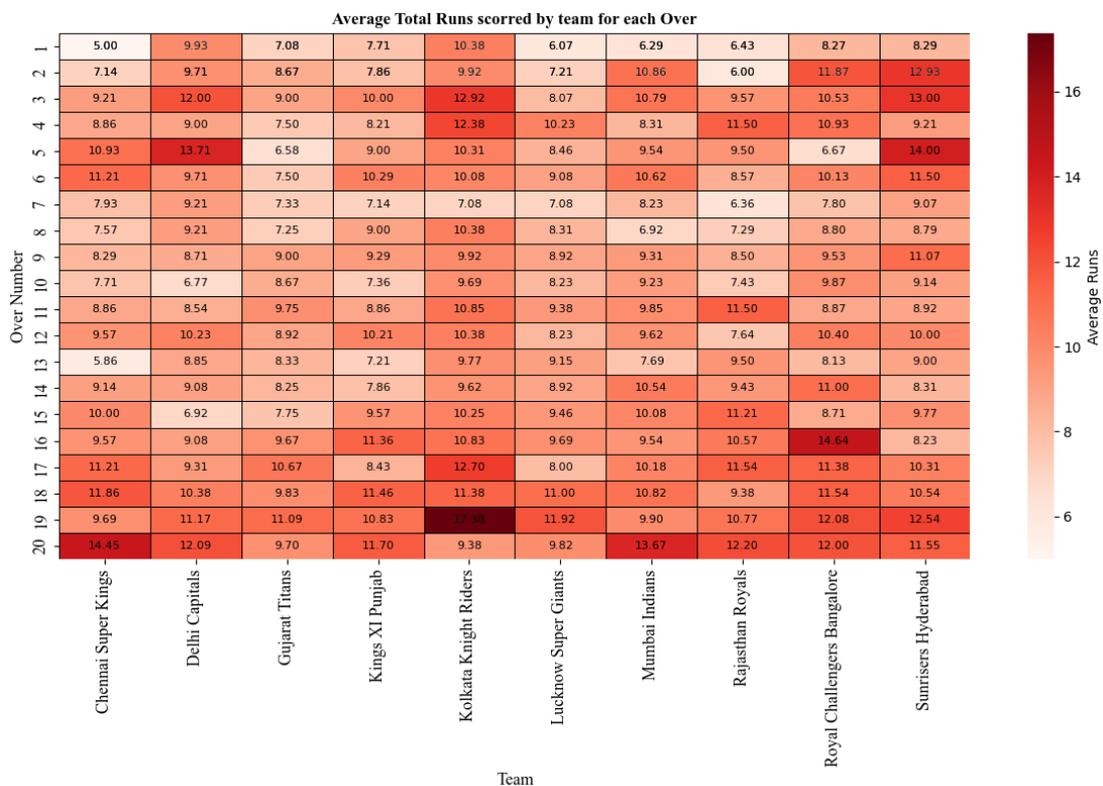

Figure 4.12: Team Wise Average Runs Scored Each Over

**4.4.3 Time Series Analysis**

Time series analysis examines into the performance of each team across different seasons, providing insights into which teams are gradually improving their performance and which teams may be experiencing a decline over time. By examining team performance over the years, time series analysis offers a detailed understanding of how teams have evolved.

As shown in Figure 4.13, it is evident that KXIP began the season with an average per inning score of 165. Throughout the seasons, they experienced fluctuations in their average scoring rate, eventually ending the 2024 season with an average score of 175 runs per match. In contrast, KKR started the season with an average score of 150 runs per match but gradually increased their scoring rate over time. By the end of the 2024 season, they achieved an impressive average score of 200 runs per match, showcasing significant growth. However, it is important to note that KTK has only played one season, making their data irrelevant for this analysis.

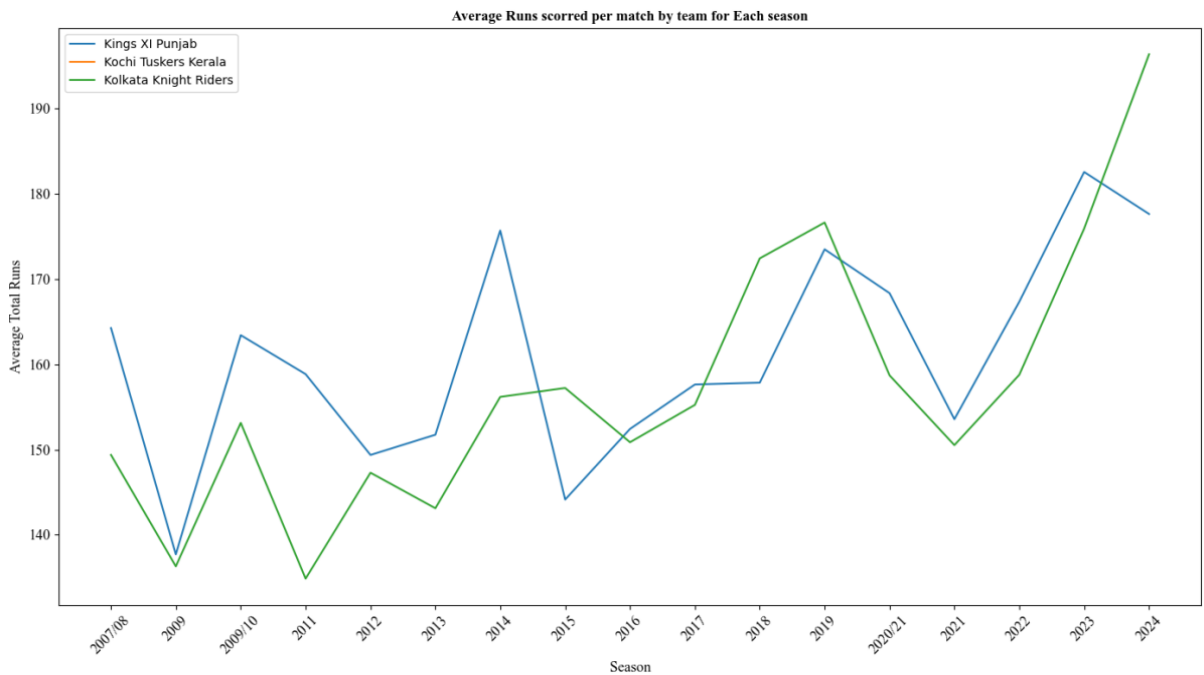

Figure 4.13: Teams Performance over the Year

As shown in Figure 4.14, it is apparent that DC have progressively increased their average scoring run rate over the years, starting from around 150 and reaching close to 190 by the end of the analysis period. Additionally, it appears that CSK did not participate in the 2016 and 2017 seasons, with GT competing in their place during those years. GT did not join the IPL at its inception in 2007; instead, they entered in 2016, participating in two seasons before returning again in 2022. Despite their late entry, GT has maintained an average run score close to 170, demonstrating competitive performance.

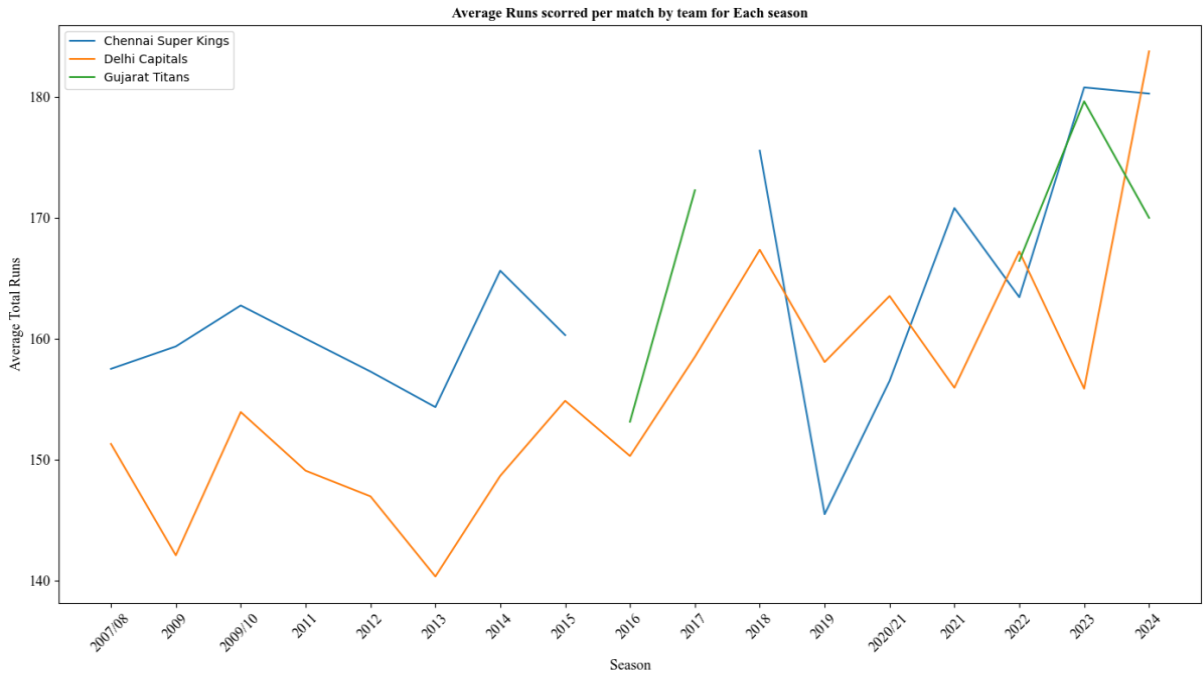

Figure 4.14: Teams Performance over the Year

As shown in Figure 4.15, RR experienced fluctuations in their performance during the initial seasons and did not participate in 2016 and 2017. Since their return, they have been gradually increasing their run rate each season. MI have shown steady growth over the years, starting with an average of 150 runs per match and reaching 170 currently. LSG appear to have started in 2022 with an average of close to 170 runs, but their run rate has decreased each season since then.

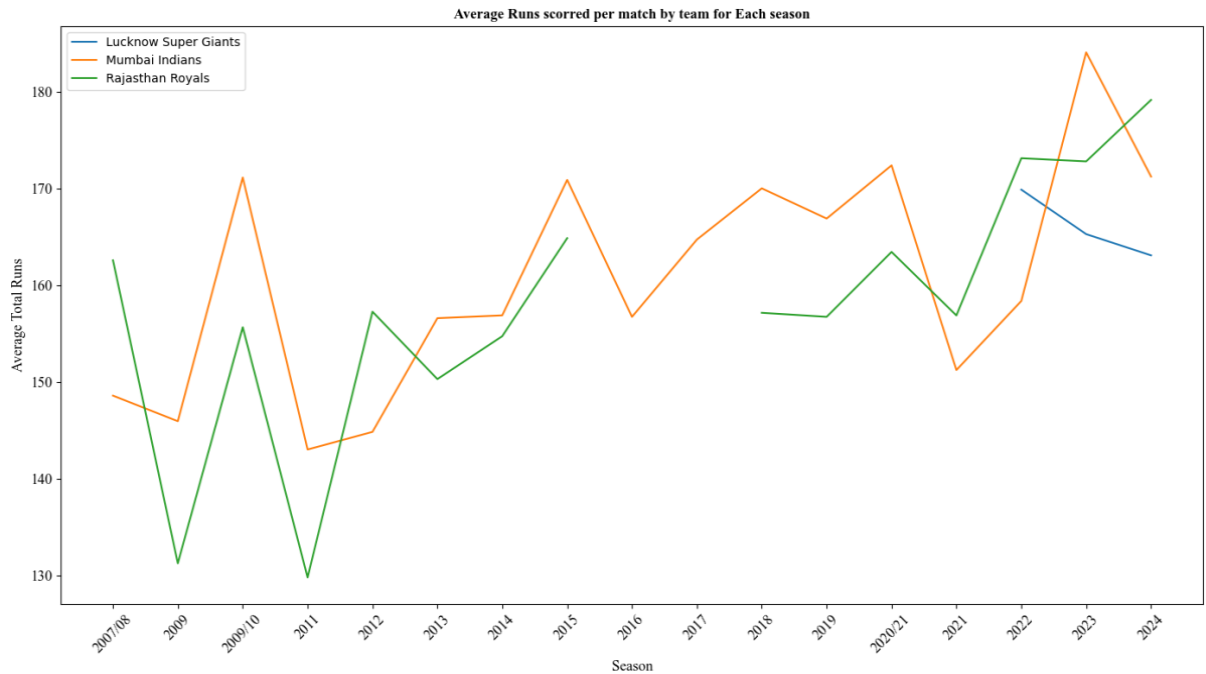

Figure 4.15: Teams Performance over the Year

As shown in Figure 4.16, it is observed that RCB began with an average of 140 runs per match and consistently increased their average each season. However, in 2014 and 2015, they experienced a drop in their average. Impressively, they achieved their highest average of close to 190 runs in 2016, indicating an extraordinary season for them. On the other hand, SRH started well with an average of 160 runs but saw a drop for a few years until 2013. Since then, they have steadily increased their average each season, with a significant rise in their run rate after 2021. Additionally, it appears that RPS have only played for a few seasons, making their data less comprehensive.

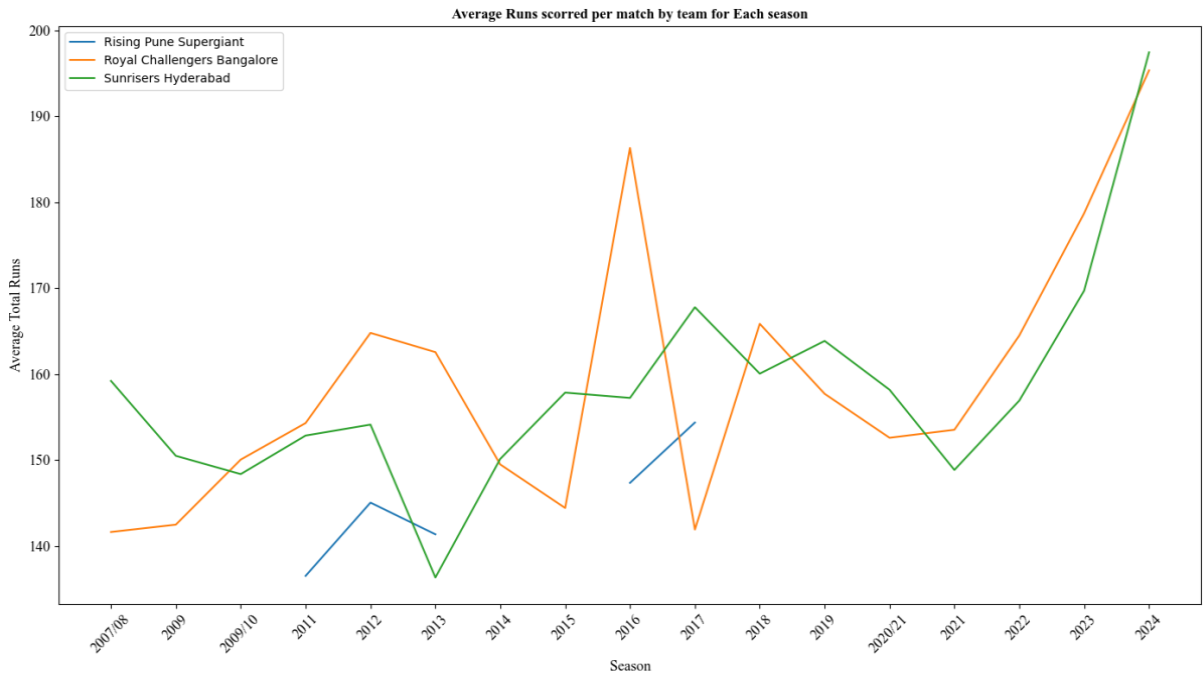

Figure 4.16: Teams Performance over the Year

Analyzing team performance over each match played throughout the years is an effective way to assess consistency, identify periods of low or high performance, and understand overall team dynamics. As shown in Figure 4.17, the total runs scored by RCB in each match are visualized. It is observed that RCB did not start the 2007 season well, scoring below 100 runs. In the following season, they experienced significant fluctuations, including three matches where they scored below 100 runs. In 2013, they achieved their all-time highest score, surpassing 250 runs.

However, in 2015, RCB recorded their lowest score, which was less than 50 runs—a notable deviation from their average scoring performance for that season. The 2016 season appears to have been their best, with consistently high performances. The current 2024 season also seems promising, as they have scored close to or more than 200 runs in several matches. This analysis highlights RCB's journey of fluctuating performances and notable achievements over the years.

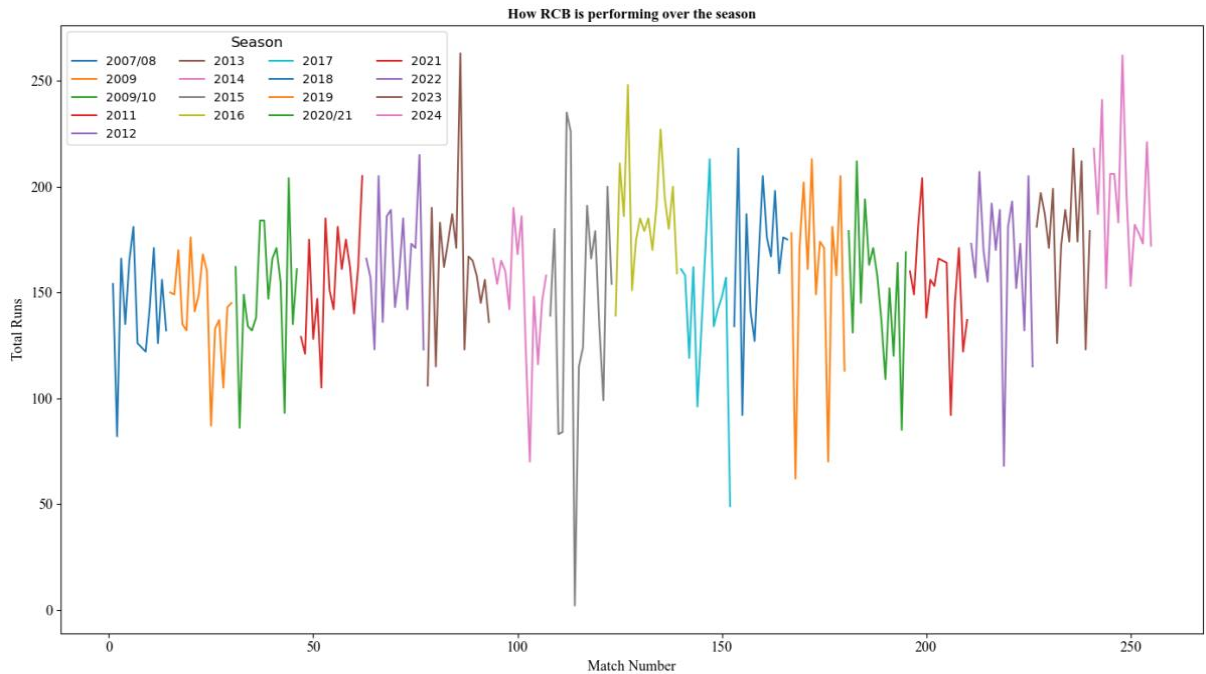

Figure 4.17: Team's Performance over the Year

It is also very important to analyze how different players perform over the years, as this provides insights into their individual contributions and development. When examining Virat Kohli's performance over the years, as shown in Figure 4.18, it appears that in the early seasons, his highest score in an innings was around 40 runs. Each season, he steadily increased his highest score. In 2013, he scored his first century in the league.

The 2016 season was particularly remarkable for Kohli, as he frequently surpassed 100 runs in an innings. In the 2024 season, he achieved his all-time highest score, nearing 120 runs. Additionally, it appears that most often, Kohli scores between 40 to 60 runs in an innings, indicating a consistent and reliable performance as a key player for his team.

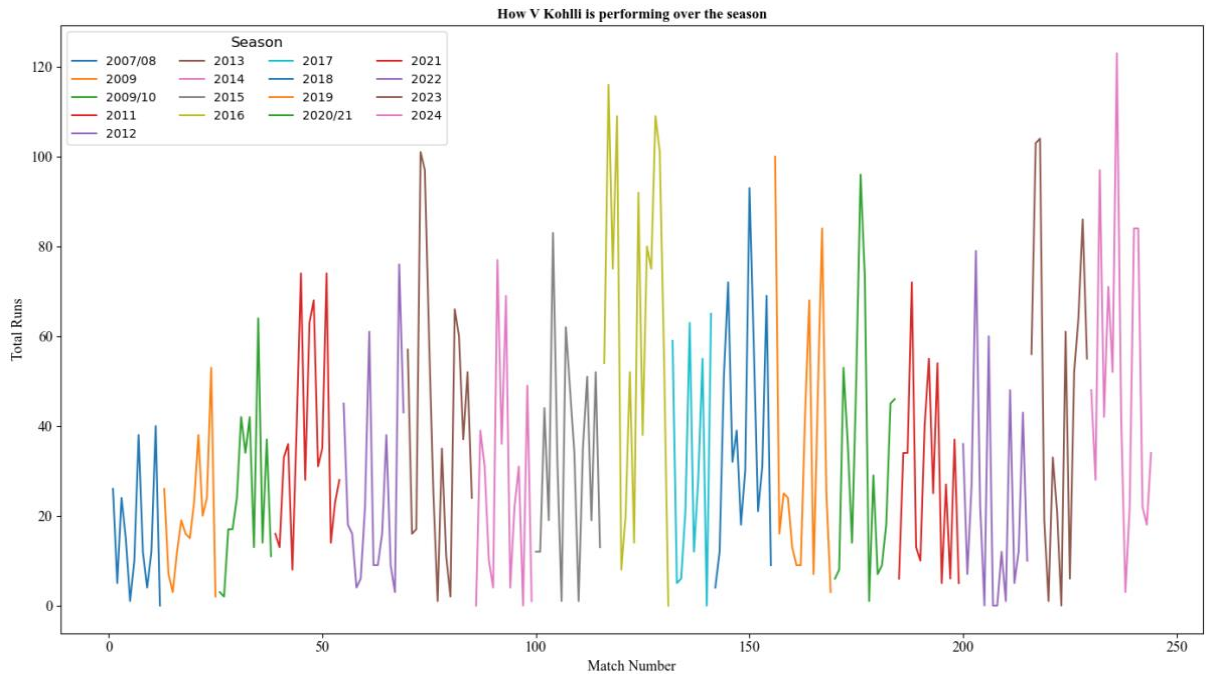

Figure 4.18: Player's Performance over the Year

### 4.4.4 Predictive Analysis

Predictive analysis is beneficial for determining how many runs a team is likely to score or which team is likely to win, based on historical data. To predict a team's total score, several factors are analyzed, such as the average runs scored per ball, changes in run rate when wickets fall, the average runs scored during a season, the venue of the match, the opposition team, and whether the team is batting first or second. By considering all these statistical details, the predicted total score for a team is calculated.

As shown in Figure 4.19, one recent match is taken as an example, and the first inning's predicted total runs are calculated on a ball-by-ball basis. This predictive analysis helps in forecasting the likely outcomes of the match and the performance metrics for each team, thereby providing valuable insights for strategy and planning.

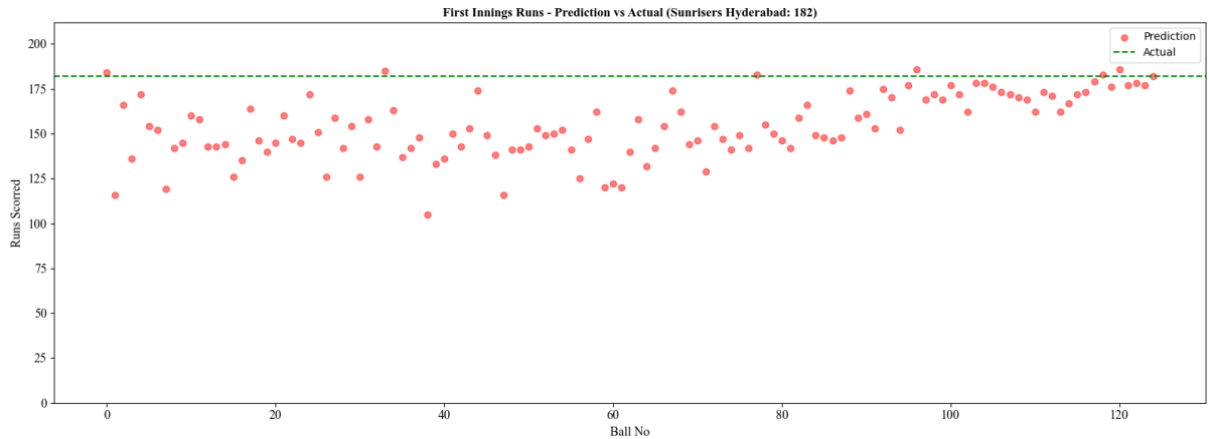

Figure 4.19: First Inning Score Prediction

Figure 4.20 illustrates the forecasted runs scored for the second inning, determining whether the opposition team will achieve the target from the first ball onwards. The analysis predicts not only the likelihood of reaching the target but also the specific over in which they are expected to achieve it. This predictive model provides a comprehensive view of the second inning's dynamics, offering strategic insights into the team's performance and potential match outcomes.

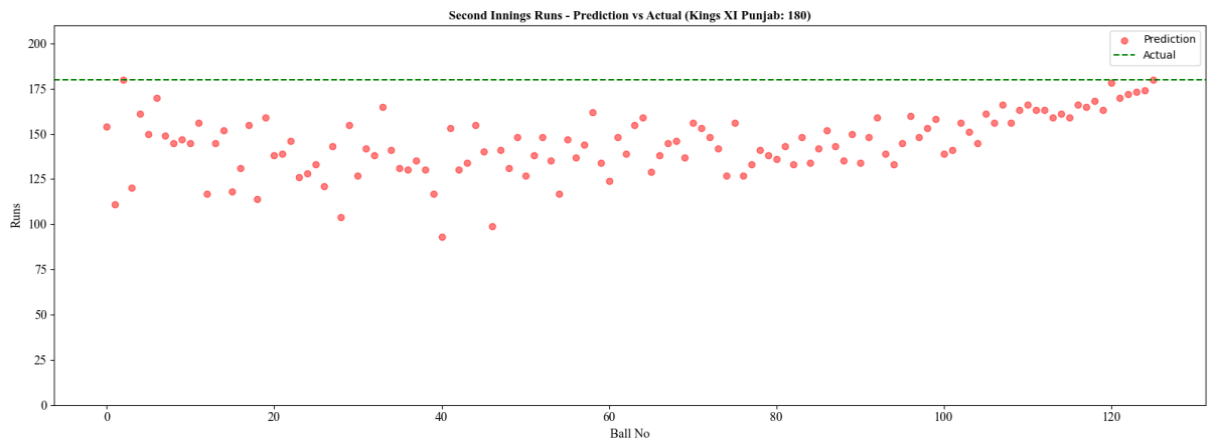

Figure 4.20: Second Inning Score Prediction

Figure 4.21 illustrates the likelihood of each team winning the match from the first ball to the last. It shows the winning percentage chance of Team 1 based on their first innings score and the winning percentage chance of Team 2, who is batting second. The graph tracks how both teams progress through their innings, with their winning chances calculated at each stage. This dynamic visualization provides a clear understanding of how the probability of winning shifts throughout the match, offering valuable insights into the teams' performance and strategic decisions.

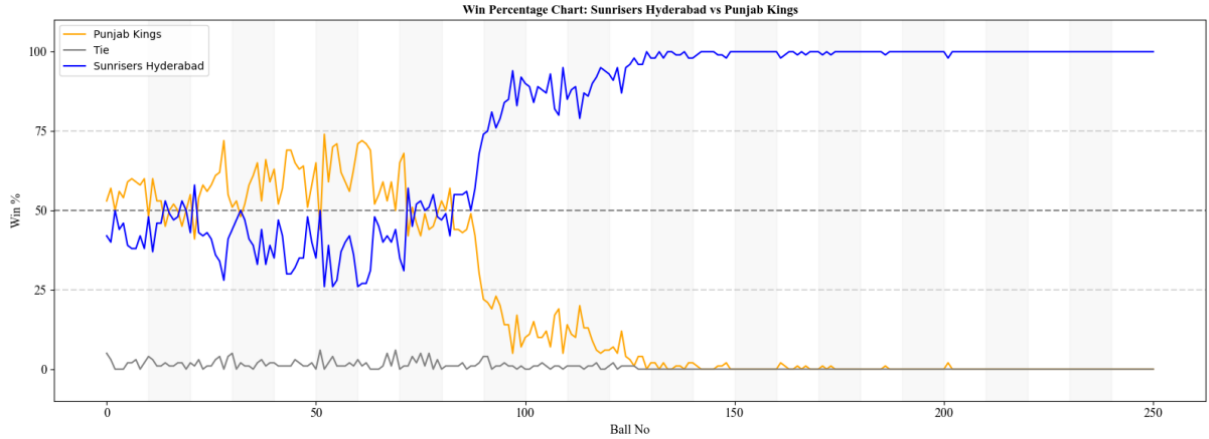

Figure 4.21: Winning Prediction

## 4.5 Experiments: Question-Answering Model

Document question-answering is performed on the second dataset, which contains context information for all historical matches along with formulated questions and answers from that context. This model provides natural language responses to user questions based on the contextual data. The context includes details such as the two teams playing, the city and venue of the match, the toss winner and the decision made after winning the toss, the match winner, and the individual scores of players. The BERT base pre-trained model is utilized to develop this model.

The dataset is split into training and testing sets with an 80:20 ratio. The initial model is trained with the following parameters: number of train epochs is chosen as three, per device train and eval batch size is chosen as 16, a learning rate of 2e-5, warm up step of 500, a weight decay of 0.01 and gradient accumulation step is chosen as four. However, during evaluation based on accuracy, precision, recall, and F1-score, the initial results were not satisfactory with F1-score as 0.16, indicating the need for hyperparameter tuning.

Hyperparameter tuning is performed using the following values: learning rate is suggested to be between 1e-5 and 7e-4, the per device train batch size is chosen from 8, 16, 32 or 64, weight decay is chosen between 0.0, 0.01 and 0.1, warmup steps are chosen between 0 and 500, adam epsilon are chosen from 1e-8, 1e-7 and 1e-6, and the number of training epochs is suggested between three, four and five.

After hyperparameter tuning, the optimal parameters for the document question answering model are found to be a learning rate of approximately 5e-05, a per device train batch size of eight, a weight decay of 0.0, an adam epsilon of 1e-06, 500 warmup steps, and five training epochs.

## 4.6 Experiments: Predictive Model

Predictive modeling is performed on the third dataset, which contains context information for all historical matches along with the corresponding winners. This model aims to predict the winner of a match based on various factors. It includes details such as the two teams playing, the city and venue of the match, the toss winner, and the decision made after winning the toss, as well as the playing XI for both teams. These factors are crucial in predicting the match outcome, providing a comprehensive approach to forecasting the winner based on historical patterns and contextual variables.

There are only ten teams participating in the IPL currently, so the dataset is filtered to include only these teams. The study identifies two different approaches for designing predictive models to predict the winner: Binary Classification and Multi-Class Classification. Both approaches leverage the rich contextual information available in the dataset to make accurate predictions about the match outcomes.

### 4.6.1 Binary Class Predictive Modelling

In this approach, the context is used to extract the two teams (Team 1 and Team 2) playing the match. This information is then fed into a binary classification model, which predicts the winner based on the comparison between Team 1 and Team 2.

To predict the winner between Team 1 and Team 2, the BERT Tokenizer and BERT For Sequence Classification model are employed, leveraging their capability to classify contextual text into different categories. These models are particularly effective for tasks such as sentiment analysis and various classification tasks.

The dataset is split into training and testing sets with an 80:20 ratio. The model uses two labels, representing the binary winning outcome (Team 1/Team 2). The initial model is trained with the following parameters: number of train epochs is chosen as one and per device train batch size is chosen as eight. However, during evaluation based on accuracy, precision, recall, and F1-score, the initial results were not satisfactory with F1-score as 0.34, indicating the need for hyperparameter tuning.

Hyperparameter tuning is performed using the following values: learning rate is suggested to be between 1e-5 and 5e-4, the per device train batch size is chosen from 8, 16, or 32, weight decay is suggested between 1e-6 and 1e-2, warmup steps are suggested between 0 and 500, and the number of training epochs is suggested between one and five.

After hyperparameter tuning, the optimal parameters for the binary classification model are found to be a learning rate of approximately 3.5e-5, a per device train batch size of 8, a weight decay of approximately 3.8e-5, 466 warmup steps, and five training epochs.

### 4.6.2 Multi Class Predictive Modelling

Multi-Class Classification: Here, the model uses ten different labels, one for each team. The context, including details such as the teams, city, venue, toss outcome, and playing XI, is fed into a multi-class classification model. The model then predicts the winner by determining which of the ten team labels is the most likely outcome.

The multi-class prediction model is used for predicting the winner out of ten different teams. For this analysis, the BERT Tokenizer and BERT For Sequence Classification model are utilized.

The data is split into training and testing sets with an 80:20 ratio. The model uses ten labels, representing the ten different teams. The initial model is trained with three training epochs, a batch size of 16 for both training and evaluation, four gradient accumulation steps, 500 warmup steps, a weight decay of 0.01, and a learning rate of 5e-5. Additionally, the logging is set to every ten steps, the evaluation strategy is "steps", and the model is saved every 500 steps with a limit of three total saves. Despite these settings, the evaluation results based on accuracy, precision, recall, and F1-score were not satisfactory with F1-score as 0.04, indicating the need for hyperparameter tuning.

Hyperparameter tuning is performed with the following values: the learning rate is suggested to be between 1e-5 and 5e-5, the number of training epochs between two and five, the per device train batch size chosen from 8, 16, or 32, warmup steps between 0 and 500, weight decay between 0.01 and 0.1, and logging steps between 10 and 50.

After hyperparameter tuning, the optimal parameters for the multi-class classification model are found to be a learning rate of approximately 1.43e-5, four training epochs, a per device train batch size of 16, 447 warmup steps, a weight decay of approximately 0.02, and logging every 11 steps.

## 4.7 Summary

The study provides a comprehensive analysis of the Indian Premier League dataset, offering valuable insights into various aspects of the matches, teams, and players. The dataset encompasses detailed information from all IPL seasons, ranging from 2007 to 2024, including match metadata, team details, and ball-by-ball actions.

The statistical analysis reveals critical findings about the IPL. The total number of matches played is 1093, with the league commencing in 2007 and continuing uninterrupted, including during the COVID-19 pandemic. The league has seen participation from 19 different teams, played across 36 cities and 58 venues. Notably, Mumbai has hosted the most matches, followed by Bangalore, Kolkata, Delhi, and Chennai. In terms of team performance, Mumbai Indians have won the most matches, although Chennai Super Kings have the highest win percentage. The analysis of toss outcomes indicates no significant advantage in winning the match, as the match results are nearly evenly split between teams winning and losing the toss.

The EDA focuses on innings-related data, providing insights into team performance metrics such as match-winning percentages and toss-winning percentages. It identifies consistent patterns and trends that highlight team strengths and weaknesses. The analysis shows that teams like Chennai Super Kings and Mumbai Indians excel in making strategic decisions after winning the toss, while others like Kings XI Punjab and Lucknow Super Giants tend to perform better when they lose the toss.

Time series analysis tracks the performance of teams across different IPL seasons. It highlights trends such as the steady improvement of Kolkata Knight Riders and Delhi Capitals in their scoring rates over the years. Conversely, teams like Royal Challengers Bangalore exhibit fluctuations, with some extraordinary seasons like 2016. The analysis also points out the participation and performance of teams like Gujarat Titans and Rising Pune Supergiants.

Detailed examination of team performance over the years provides insights into consistency and variations in team performance. For instance, Royal Challengers Bangalore's match-by-match run totals reveal significant improvements and occasional setbacks. Player performance analysis, focusing on key players like Virat Kohli, illustrates their growth and consistency, highlighting standout seasons and high-scoring innings.

Predictive analysis leverages historical data to forecast team scores and match outcomes. The analysis uses context-specific data to predict match winners, taking into account factors such as venue, opposition, and playing conditions. Predictive models are refined through hyperparameter tuning to improve accuracy, with BERT models being employed for both binary and multi-class classification tasks.

All the experiments conducted for both the document question-answering model and the predictive model are discussed. For the document question-answering model, various hyperparameters were tuned during training to optimize performance, leveraging the BERT base pre-trained model. This model was evaluated on its ability to provide accurate natural language responses based on historical match contexts.

Similarly, the predictive model experiments involved hyperparameter tuning to forecast future match outcomes based on historical data. The predictive modelling focused on how the predicted match outcome evolves ball by ball, depending on the team's performance and forecasted total runs. These experiments aimed to enhance the model's accuracy and reliability in predicting future outcomes.

Overall, the analysis chapter underscores the competitive nature of the IPL, with teams showing varied performances across different metrics. The findings highlight the importance of strategic decisions, consistent player performance, and the impact of historical trends on predicting future outcomes. This comprehensive analysis not only enhances understanding of team dynamics and performance but also provides a robust foundation for predictive modeling in the cricket.

# CHAPTER 5

# RESULTS AND DISCUSSIONS

## 5.1 Introduction

As discussed in Chapter 3, the study developed a multi-model architecture for Text2Insights, encompassing natural language text to data visualization, a document question-answering model, and predictive modelling using various approaches. This comprehensive architecture aims to convert text into actionable insights through different analytical models. Each model undergoes with different evaluation, considering multiple factors to define its effectiveness and performance. Subsequently, the end-to-end model is evaluated using a diverse range of queries, spanning from simple to complex. Lastly, both the document question-answering model and the predictive model are assessed using test data, providing comprehensive insights into their capabilities.

## 5.2 Evaluation of Text2Insights Model

The study developed a Text2Insights model designed to convert natural language text into insightful visualizations. The developed methodology utilizes a multi-model architecture comprising distinct models: Text2SQL, SQL refinement, Chart Type Prediction, Insights Generation, and Chart Generation. The results of each model are discussed individually in this section.

### 5.2.1 Evaluation of Text2SQL Model

The Text2SQL model is responsible for translating natural language queries into SQL queries. This model is a critical component of the Text2Insights system, as it enables the extraction of relevant data from input file based on user's input natural language query.

The Text2SQL model is evaluated using the 'spider-clean-text-to-sql' dataset, which contains a diverse set of natural language queries and their corresponding SQL queries. The dataset is divided into train, dev, and test sets, containing 6,016, 665, and 1,929 records, respectively. Due to the smaller size of the train dataset, the study utilizes only the train component.

As detailed in Chapter 3, the Lllama3 model is used to convert natural language text into SQL queries. For all the natural language queries in the train dataset, SQL queries are generated using Lllama3. A new dataset is then formed by pairing the actual SQL queries from the original dataset with the SQL queries generated by Lllama3.

These generated SQL queries from Llama3 are then refined using Spacy's similarity index by utilizing the schema of the actual query mentioned in the dataset to produce refined SQL queries. The dataset created with actual and Llama3 model-generated queries is then updated with refined SQL queries in place of the generated queries.

Firstly, a syntax-based evaluation is conducted on this dataset to verify the validity of the refined SQL queries using an SQL parser. It was found that out of 665 generated SQL queries, five were invalid or syntactically incorrect. These records indicate the resultant matrix of this model. As shown in Figure 5.1, the confusion matrix illustrates the syntactical results of this model.

The syntactical results show that the model's accuracy is 99.25%, with a precision of 100%, a recall of 99.25%, and an F1-score of 99.62%. These metrics indicate the model's strong performance across different cross-domain natural language queries.

Table 5.1: Text2SQL Syntactical Results

| Measure | Results |
|---|---|
| Accuracy | 0.9925 |
| Precision | 1.0000 |
| Recall | 0.9925 |
| F1 Score | 0.9962 |

As shown in Table 5.1, these parameters demonstrate the model's performance in generating syntactically correct SQL queries.

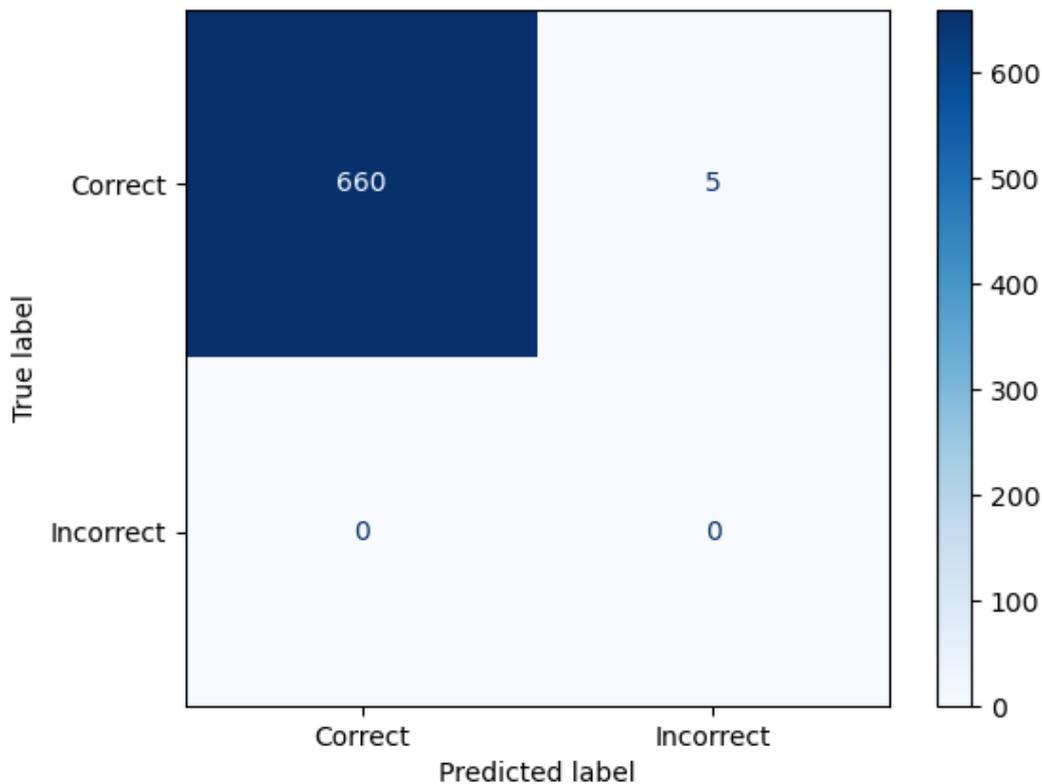

Figure 5.1: Text2SQL Syntactical Results

Secondly, the correctness of the refined SQL queries from LLaMA3 is analyzed using the BLEU score. The BLEU score assesses the quality of the generated SQL queries relative to the actual SQL queries. It measures the n-gram overlap between the generated SQL and the actual SQL. BLEU scores are calculated for all the generated queries using the NLTK's translate library.

After obtaining BLEU scores for all the generated queries, the study conducts threshold analysis with different values of BLEU. BLEU score ranging from 0.5 to 1.0 are consistently effective for determining the similarity between two SQL queries. A higher BLEU value indicates a higher similarity between the two queries. When the threshold value is set to 0.5, it is observed that all the generated queries are mostly similar to the actual ones. Increasing the threshold value of BLEU tends to decrease the performance, although a BLEU score of 0.5 remains a good indicator of similarity between two queries.

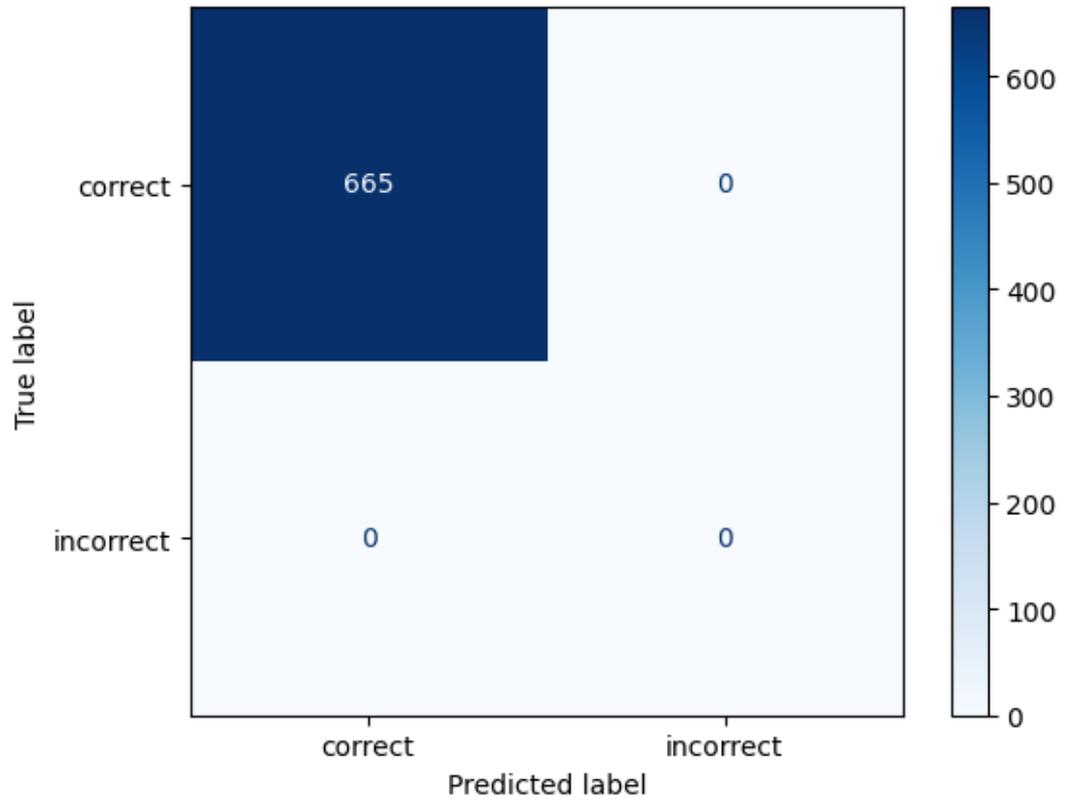

Figure 5.2: Confusion Matrix for BLEU Score

When a threshold of 0.5 for the BLEU score is selected, it can be observed from Figure 5.2 that the model correctly generates SQL queries, indicating its performance. Additionally, the accuracy, precision, recall, and F1-score are observed to be 1.0, as shown in Table 5.2.

Table 5.2: Text2SQL BLEU Results

| Measure | Results |
| --- | --- |
| Accuracy | 1.0000 |
| Precision | 1.0000 |
| Recall | 1.0000 |
| F1 Score | 1.0000 |

### 5.2.2 Evaluation of Chart Type Prediction Method

The study introduces a chart type prediction method, as discussed in Chapter 3, which predicts the chart type based on the user's input query. The model initially identifies any specified chart type in the user's query. If a preferred chart type is available in the user's query, that particular chart type is given preference. Otherwise, the chart type prediction method is employed.

In this method, the model observes the dataset generated from the SQL query and defines all the data types within that particular subset of the dataset. These data types are then utilized to predict the chart type.

The study encountered challenges in finding a suitable dataset containing chart types and their corresponding natural language descriptions. Hence, the study conducted thorough examinations of the developed method with various types of input.

It was observed that when users provide a preference for a specific chart type, the method accurately identifies it. However, when users express their preferences in a negative manner, such as "not a bar chart" or "don't need scatter plot," the model does not perform well and fails to identify the negative context of the input text.

When the chart type is not specified in the input query, the observed results vary depending on the number of columns available in the subset of the dataset. If the subset of the dataset contains univariate, bivariate, categorical, and continuous columns, the method can identify the chart type as shown in Table 5.3.

For instance, if there is one numeric datatype and one categorical datatype column, the resulting chart type is a bar chart. If the dataset consists of only univariate data, the chart type returned is a box plot. For datasets containing continuous and time series data, the result is provided as a line chart.

Table 5.3: Chart Type Prediction

| Data Type | Predicted Chart Type |
|---|---|
| Categorical and Quantitative | Bar Chart |
| Continuous and Univariate | Box Plot |
| Continuous and Time-Series | Line Chart |
| Categorical and Quantitative | Pie Chart |
| Continuous and Bivariate | Scatter Plot |
| Continuous and Univariate | Histogram |
| Continuous and Time-Series | Area Chart |
| Continuous and Multivariate | Bubble Chart |
| Continuous and Multivariate | Radar Chart |
| Continuous and Multivariate | Heatmap |

However, when multivariate columns exist in the dataset, the method determines the chart type based on a subset of columns or dataset. It does not generalize well to provide a chart type for datasets with more than five columns when the chart type preference is not available.

### 5.2.3 Evaluation of Insights generation Model

The study introduces an insight generation model, as discussed in Chapter 3. This model generates insights based on the data obtained from executing the SQL query on the input file.

The required data is then fed into the Llama3 model, which takes this data input and converts it into human-readable natural language. The results of this model entirely depend on the text-to-text generation capabilities of the Llama3 model. According to the release notes of Meta, the base model is pre-trained on 15 trillion tokens of data, indicating the significance and performance of this model. As per their official communication, the performance metrics of the Llama3 model are illustrated in Figure 5.3.

|  | PRE-TRAINED |
|---|---|
|  | Meta Llama 3 400B+ |
| MMLU 5-shot | 84.8 |
| AGIEval English 3-5-shot | 69.9 |
| BIG-Bench Hard 3-shot, CoT | 85.3 |
| ARC-Challenge 25-shot | 96.0 |
| DROP 3-shot, F1 | 83.5 |

Figure 5.3: Performance Matrix for Llama 3 (as per Meta's official page)

## 5.3 Results Generated from Text2Insights

An end-to-end evaluation of the proposed model Text2Insights on cricket data is performed. The cricket dataset was fed into the model using a range of natural language queries, from simple to complex, to assess the model's performance on different types of queries. Additionally, error validation was conducted to evaluate the model's error and exception handling capabilities.

When the data is selected as ODI_data.csv and various natural language queries are fed into the Text2Insights model, the resultant output appears as follows:

**Input Query:** Who are the top 10 run getters?
**SQL Output:** SELECT player_name, runs FROM table_name ORDER BY runs DESC LIMIT 10;
**Chart Type:** Bar Chart
**Generated Insights:** The top 10 run getters in the dataset are led by SR Tendulkar with 18426 runs.
The top 5 run getters are all above 12650 runs, with DPMD Jayawardene being the 4th highest with 12650 runs.
The gap between the top 2 run getters (SR Tendulkar and KC Sangakkara) is 4182 runs, indicating a significant lead by SR Tendulkar.
The average runs scored by the top 10 run getters is 12575.6 runs.

The list includes players from different countries, with SR Tendulkar, V Kohli, and R Dravid from India, KC Sangakkara from Sri Lanka, RT Ponting from Australia, ST Jayasuriya from Sri Lanka, DPMD Jayawardene from Sri Lanka, Inzamam-ul-Haq from Pakistan, JH Kallis from South Africa, and SC Ganguly from India.

**Generated Chart:**

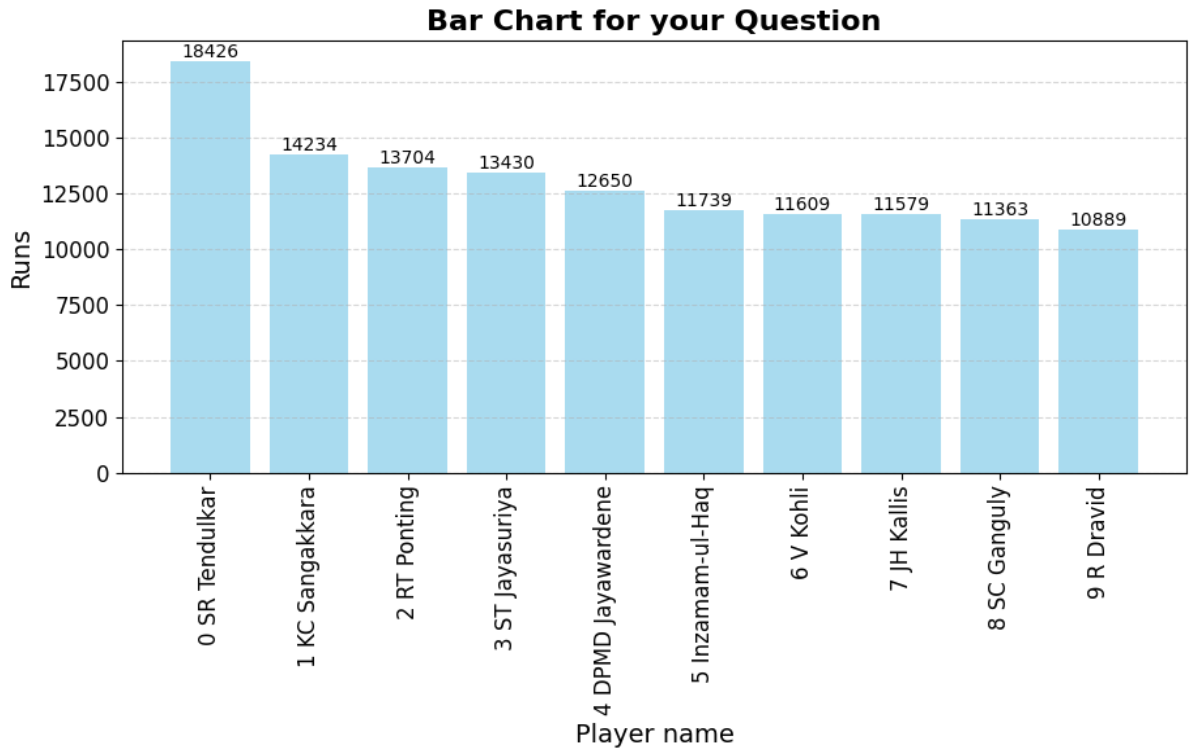

Figure 5.4: Text2Insights Generated Output - 1

**Input Query:** plot line chart for the top 5 players who has best average with their strike rate?
**SQL Output:** SELECT player_name, average, strike_rate FROM table_name ORDER BY average DESC LIMIT 5;
**Chart Type:** Line Chart
**Generated Insights:** The top 5 players with the best average are MP O'Dowd, AL Kandappah, Aamer Yamin, KJ Barnett, and CT Radley.
MP O'Dowd has the highest average of 145.00, which is significantly higher than the rest of the players.

Aamer Yamin has the highest strike rate of 98.95, which is significantly higher than the rest of the players.

There is a positive correlation between average and strike rate, as the top 2 players with the highest average also have high strike rates.

The average strike rate of the top 5 players is around 70, which indicates that they are all aggressive players.

CT Radley has the lowest strike rate among the top 5 players, but still has a high average, indicating that he is a consistent player.

There is a significant gap between the average of the top player (MP O'Dowd) and the rest of the players, indicating that he is an exceptional player.

The data suggests that a high strike rate is not the only factor that contributes to a high average, as KJ Barnett has a relatively low strike rate but still has a high average.

**Generated Chart:**

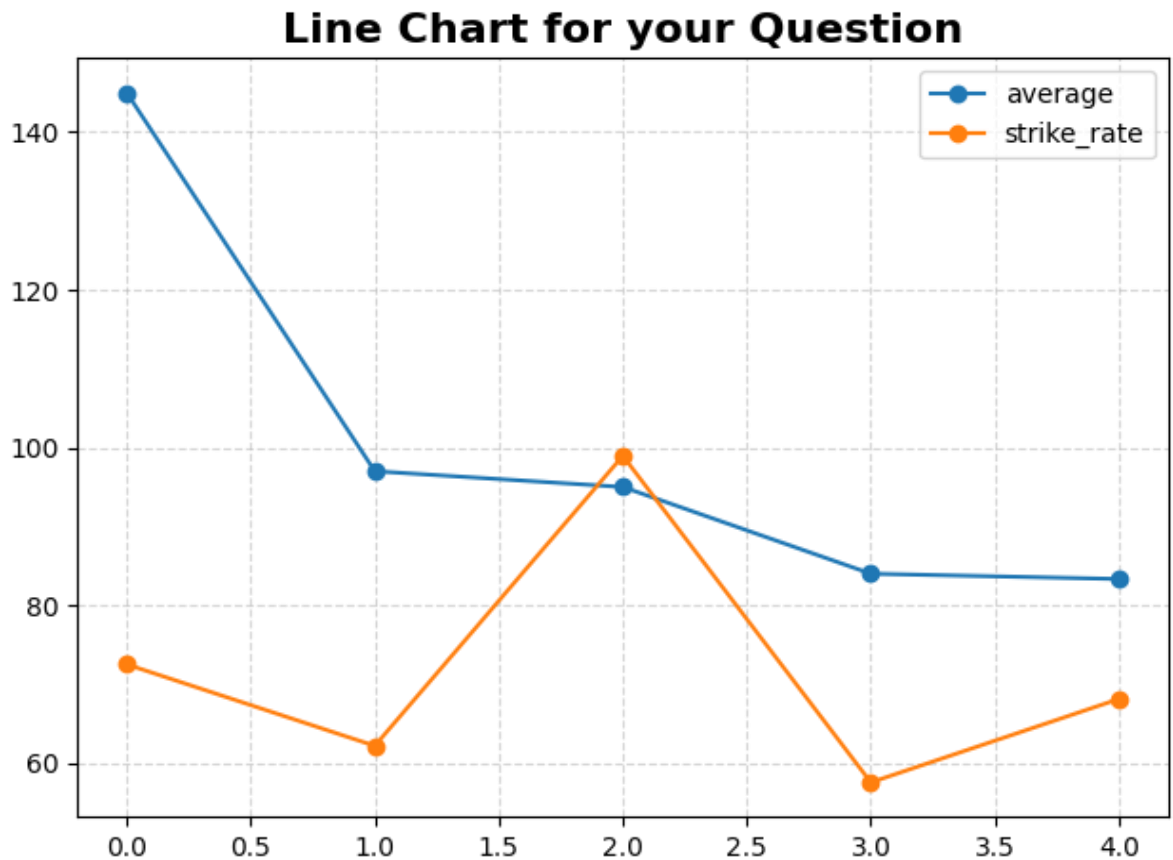

Figure 5.5: Text2Insights Generated Output - 2

**Input Query:** Show me heatmap for the top 10 run getter with their strike rate and average and total run and 100 scored?

**SQL Output:** SELECT player_name, runs, average, strike_rate, one_hundred FROM table_name ORDER BY runs DESC LIMIT 10;

**Chart Type:** Heatmap

**Generated Insights:** The top 10 run getters have scored a total of 115,132 runs, with SR Tendulkar leading the pack with 18,426 runs.

The top 10 run getters have an average strike rate of 81.45, with ST Jayasuriya having the highest strike rate of 91.20.

The top 10 run getters have scored a total of 225 centuries, with SR Tendulkar and V Kohli scoring the most centuries with 49 and 43 centuries respectively.

The top 10 run getters have an average of 41.43, with V Kohli having the highest average of 59.84.

The top 5 run getters are from the following countries: India (2), Sri Lanka (2), Australia (1), Pakistan (1), and South Africa (1).

The top 10 run getters have played a total of 541 matches, with SR Tendulkar playing the most matches (248).

The top 10 run getters have an average of 34.5 centuries per player, with SR Tendulkar and V Kohli having the most centuries.

**Generated Chart:**

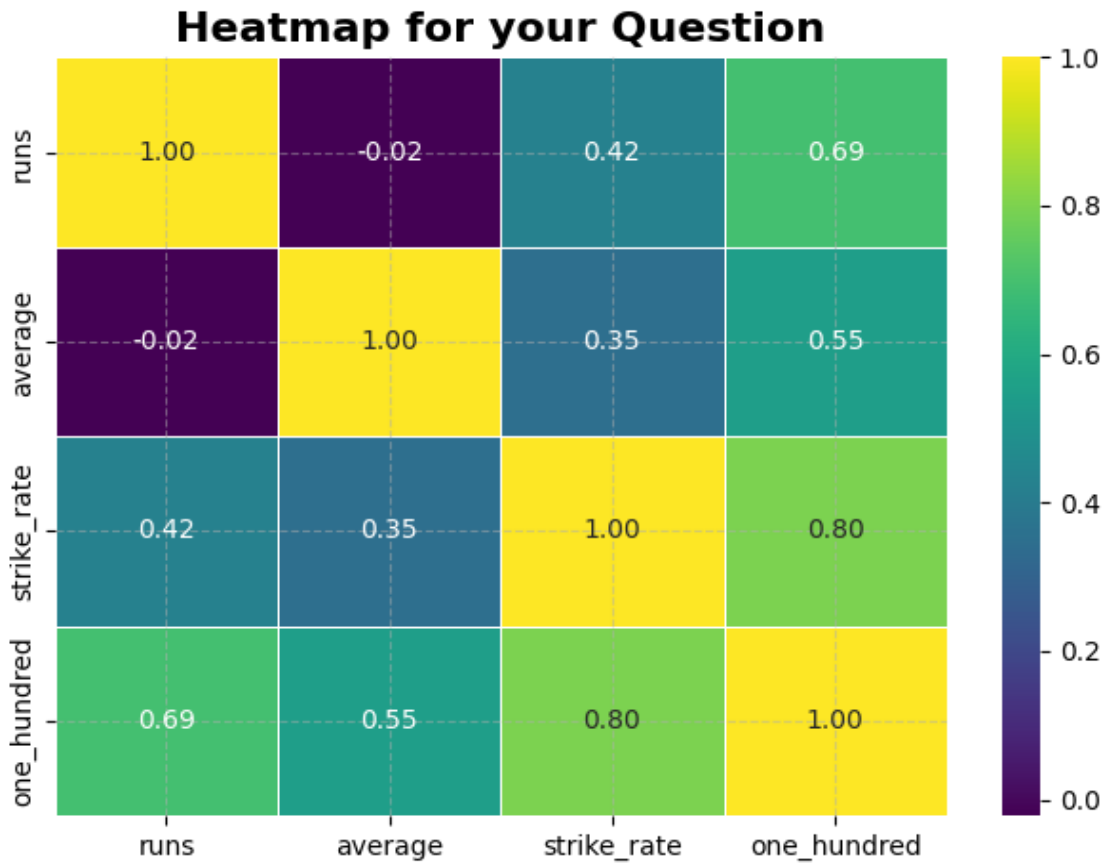

Figure 5.6: Text2Insights Generated Output - 3

**Input Query:** compare top 10 run getter with their strike rate and average?

**SQL Output:** SELECT player_name, runs, strike_rate, average FROM table_name ORDER BY runs DESC LIMIT 10;

**Chart Type:** Scatter Plot

**Generated Insights:** The top run getter is SR Tendulkar with 18426 runs, having a strike rate of 86.23 and an average of 44.83.

The top 3 run getters are SR Tendulkar, KC Sangakkara, and RT Ponting, all of whom have scored over 13700 runs.

The highest strike rate among the top 10 run getters is 93.28, achieved by V Kohli, who is also the highest average holder with 59.84.

The lowest average among the top 10 run getters is 32.36, achieved by ST Jayasuriya, who still managed to score over 13400 runs.

The top 10 run getters have an average strike rate of 81.44 and an average of 42.35.

Only two players, V Kohli and SR Tendulkar, have an average above 44, while three players, ST Jayasuriya, DPMD Jayawardene, and Inzamam-ul-Haq, have an average below 40.

The top 10 run getters have scored a combined total of 114,123 runs, with an average of 11412.3 runs per player.

**Generated Chart:**

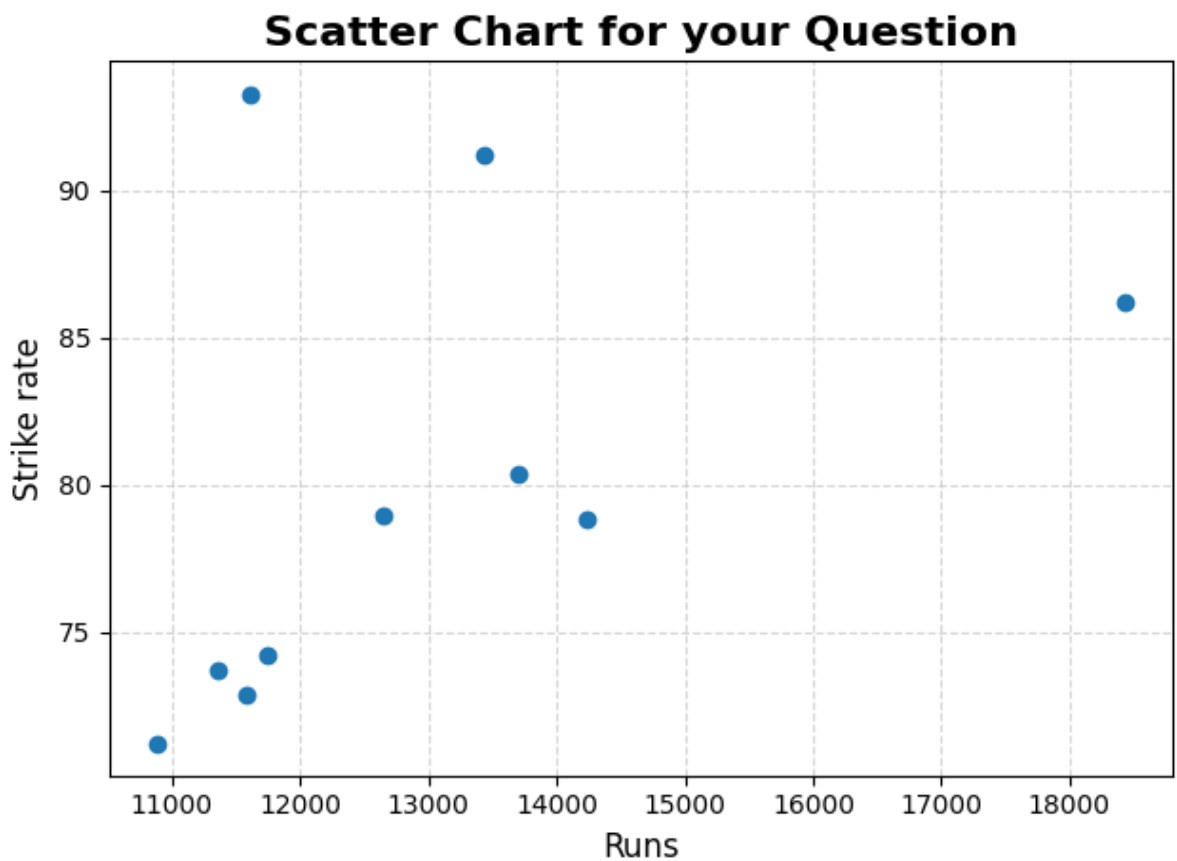

Figure 5.7: Text2Insights Generated Output - 4

**Input Query:** Provide me pie chart for top 5 player with their 100?

**SQL Output:** SELECT player_name, one_hundred FROM table_name ORDER BY one_hundred DESC LIMIT 5;

**Chart Type:** Pie Chart

**Generated Insights:** The top 5 players with the most number of centuries (100s) are:

SR Tendulkar with 49 centuries

V Kohli with 43 centuries

RT Ponting with 30 centuries

ST Jayasuriya and RG Sharma, both with 28 centuries

The data suggests that SR Tendulkar is the leading player in terms of centuries scored, with a significant margin of 6 centuries over the second-placed V Kohli.

The top 5 players are from different countries, indicating a diverse representation of cricketing talent.

The data does not provide information about the time period or the format of the matches, but it gives a general idea of the top performers in terms of centuries scored.

**Generated Chart:**

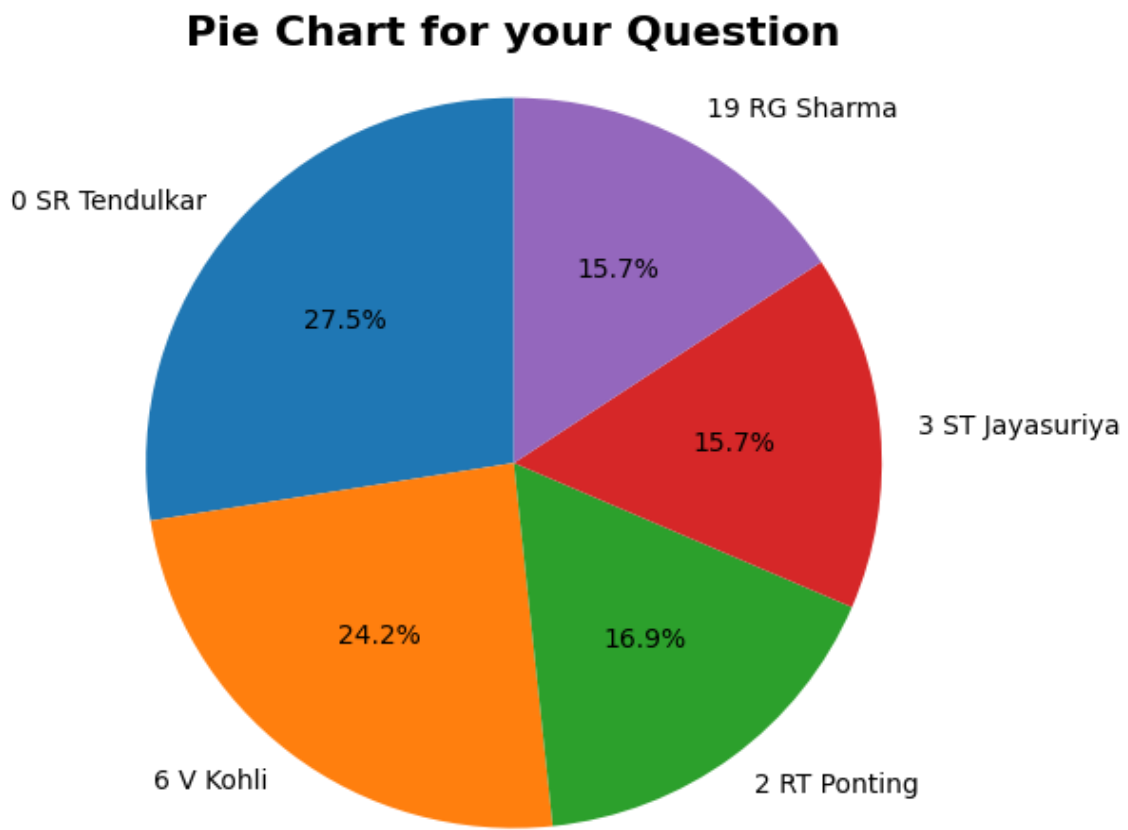

Figure 5.8: Text2Insights Generated Output - 5

**Input Query:** Who is the prime minister of India?

**SQL Output:** Wrong question for the data.

**Chart Type:** Invalid Data Provided.

**Generated Insights:** Looks like you are providing incorrect question, which is not related to your dataset. Please provide valid question for your dataset!

**Generated Chart:** Looks like you are providing incorrect question, which is not related to your dataset. Please provide valid question for your dataset!

Figures 5.4 to 5.8 display the resultant output from the developed model. These figures demonstrate the model's performance with different input queries, ranging from simple to complex queries, both when the chart type is provided and when it is not provided.

## 5.4 Evaluation of Question-Answering Model

The study developed a question-answering model for IPL cricket matches using the BERT model. This model provides natural language answers to user questions by analysing contextual historical data, including details such as the teams playing, match winners, toss winners, players of the match, and individual scores for all players involved. To evaluate the results of the question-answering model, the study divided the dataset into 80% training data and 20% testing data. The model was developed using the training dataset and evaluated on the testing dataset.

Table 5.4: Performance Matrix for Question-Answering Model

| Measure | Value |
| --- | --- |
| Loss | 1.5875 |
| Accuracy | 0.8883 |
| Precision | 0.7235 |
| Recall | 0.6945 |
| F1-Score | 0.6894 |
| Runtime | 14.2854 |
| Samples Per Second | 66.782 |
| Eval Steps Per Second | 8.4 |
| Epoch | 5.0 |

As indicated by the evaluation results shown in Table 5.4, It can be seen that with an evaluation loss of 1.58, an accuracy of 88.83 %, precision of 72.35%, recall of 69.45%, and an F1 score of 68.94%. Moreover, achieving these results within an evaluation runtime of 14.28 minutes, with an impressive throughput of 66.782 samples per second and 8.4 steps per second, shows the efficiency of the model. These results, achieved after fine-tuning and optimizing hyperparameters, shows the model's robustness and potential for practical application in answering the user's query based on input document.

## 5.5 Evaluation of Predictive Model

The study developed a prediction model for IPL cricket matches using the BERT classification model. This model predicts the winner from natural language text by analyzing historical data available for the teams playing, their records against the opposition, the provided venue, and the playing teams.

To evaluate the results of the predictive model, the study divided the dataset into 80% training data and 20% testing data. The model was developed using the training dataset and evaluated on the testing dataset.

### 5.5.1 Evaluation of Binary Classification Model

A binary classification model using BERT is developed to predict the outcome between Team 1 and Team 2, using two labels: Team 1 and Team 2. The model is fine-tuned on the historical data of IPL matches. Through hyperparameter tuning, the best parameters are identified, and the final model is developed using these optimal parameters. When evaluating this model on the test dataset, the performance metrics are presented in Table 5.5.

Table 5.5: Performance Matrix for Binary Predictive Model

| Measure | Value |
|---|---|
| Loss | 0.4008 |
| Accuracy | 0.6976 |
| Precision | 0.7040 |
| Recall | 0.6571 |
| F1-Score | 0.6798 |
| Runtime | 8.8143 |
| Samples Per Second | 24.392 |
| Eval Steps Per Second | 3.063 |
| Epoch | 5 |

As indicated by the evaluation results shown in Table 5.5, It can be seen that with an evaluation loss of 0.4008, an accuracy of 69.77%, precision of 70.41%, recall of 65.71%, and an F1 score of 67.98%, the model demonstrates a considerable capability in distinguishing between Team 1 and Team 2 outcomes. Moreover, achieving these results within an evaluation runtime of 8.8143 seconds, with an impressive throughput of 24.392 samples per second and 3.063 steps per second, shows the efficiency of the model. These results, achieved after fine-tuning and optimizing hyperparameters, shows the model's robustness and potential for practical application in predicting IPL match winners.

### 5.5.2 Evaluation of Multi Class Classification Model

A multi-class classification model using BERT is developed to predict the outcome from the ten teams currently participating in the IPL. The model employs ten labels, one for each team, to classify the match outcomes accurately. The model is fine-tuned on the historical data of IPL matches. Through hyperparameter tuning, the best parameters are identified, and the final model is developed using these optimal parameters. When evaluating this model on the test dataset, the performance metrics are presented in Table 5.6.

Table 5.6: Performance Matrix for Multi-Class Predictive Model

| Measure | Value |
|---|---|
| Loss | 4.5578 |
| Accuracy | 0.49 |
| Precision | 0.4969 |
| Recall | 0.4909 |
| F1-Score | 0.4897 |
| Runtime | 0.7801 |
| Samples Per Second | 256.379 |
| Eval Steps Per Second | 32.047 |
| Epoch | 4 |

As indicated by the evaluation results shown in Table 5.6, It can be seen that despite a relatively high evaluation loss of 4.5578, the model demonstrates an accuracy of 49%, with a precision of 49.70%, recall of 49.10%, and an F1 score of 48.98%. Despite these relatively lower metrics compared to binary classification model's performance, achieving them within an evaluation runtime of 0.7801 seconds. Moreover, with a throughput of 256.379 samples per second and 32.047 steps per second, the model showcases efficiency in processing data. While there's room for improvement in the model's performance, these results provide insights into its capabilities and areas for further refinement, contributing to advancements in IPL match outcome prediction for multi class classification model.

## 5.6 Summary

This chapter presents a results analysis of Text2Insights model, question-answering model and predictive models. Results for Text2Insights model, which converts natural language text into insightful visualizations, utilizing a multi-model architecture, reveal strong syntactical accuracy and BLEU scores, indicating the model's robust performance in generating SQL queries. Furthermore, the model accurately predicts chart types based on dataset characteristics. However, challenges arise when handling negative input queries or datasets with multiple columns.

Evaluating the question-answering model, the fine-tuned model on historical IPL data exhibits promising accuracy, precision, and recall metrics. Evaluating the predictive model for winner prediction, a binary classification model fine-tuned on historical IPL data demonstrates promising accuracy, precision, and recall metrics. Additionally, a multi-class classification model using BERT is developed to predict outcomes from the ten IPL teams. Despite the efficiency in processing data, there's room for improvement in model performance.

# CHAPTER 6

# CONCLUSIONS AND RECOMMENDATIONS

## 6.1 Introduction

The conclusions and recommendations are derived from the research methodology developed in this study. Building upon the development and evaluation of the Text2Insights model, question-answering model and predictive models for IPL datasets, this chapter highlights key findings, insights, and areas for future research. Detailed discussion on the performance and limitations of the models, along with suggestions for enhancing their effectiveness and applicability are discussed.

Furthermore, recommendations for future research directions and potential extensions of the Text2Insights framework is outlined, offering valuable insights for researchers, practitioners, and stakeholders interested in leveraging dynamic data visualization and analysis techniques for data exploration and decision-making. Through a comprehensive review of the study's contributions and implications, this chapter aims to provide actionable insights and guidance for advancing the field of data analysis and visualization.

## 6.2 Discussion and Conclusions

Data visualization systems is characterized by a static nature, where visualizations are generated based on predefined data and chart types. This becomes a significant drawback when user requirements deviate from the predetermined visualization techniques, making it difficult to obtain customized visual representations from the existing systems. The problem at hand is the lack of adaptability and flexibility in current data visualization tools.

To understand the problem this study has examined existing research on visualizing data from natural language text. It focuses on available methods, models, and datasets, as well as how large language models can enhance data visualization performance. Survey papers also reviewed to understand the current state of research in this field.

It has been observed from literature review is that the domain of converting natural language text into analysis and visualizations remains relatively underexplored within current academic research. Despite its potential significance, this area has received limited attention, leaving substantial room for further exploration and improvement.

To overcome the identified problem, study proposes the Text2Insights model, which converts natural language text into insightful charts from the provided data. Text2Insights follows a series of steps to achieve the goal of converting text to visualization. First, it converts natural language into an SQL query using Llama3. This SQL query is then refined based on the available schema of the input dataset using NER's similarity index library. Next, the SQL query is executed on the dataset to obtain the resultant subset of data using pandasql library. The subsequent step predicts the chart type based on the input query or the subset of the dataset using developed method. Finally, the chart and insights are generated based on the chart type and the subset of data obtained from the previous steps using matplotlib and seaborn libraries.

The Text2SQL step, responsible for translating natural language into SQL, is evaluated based on syntactical accuracy and correctness of the SQL query. The 'spider-clean-text-to-sql' development dataset is used for this evaluation. The developed model generates SQL queries for all 665 natural language inputs provided in the development dataset, and these queries are assessed for correct SQL syntax using a SQL parser. Evaluation results indicate that the Text2SQL model achieves an accuracy of 99.25%, with a precision of 100%, a recall of 99.25%, and an F1-score of 99.62%.

The BLEU score is then calculated for all the model-generated SQL queries. A threshold value of 0.5 is established to determine the correctness of the generated SQL query compared to the actual SQL query provided in the dev dataset. It is identified that the model achieves an accuracy, precision, recall, and F1-score of one for the generated queries.

The SQL query is refined using the generated schema of the provided input file. This refinement process involves checking for the similarity between dataset columns and the keywords in the generated SQL query using NER's word similarity index. If the SQL query contains a field name that is not present in the dataset schema, the field name is replaced with the most similar field name available in the schema.

The refined SQL query is then executed on the dataset using the pandasql library, obtaining the required subset of data from the input file. The resultant output is subsequently used to generate the chart and derive insights.

The study developed a method to determine the chart type based on the user's input. Initially, the method checks the user's preference. If user has specified a particular chart type, that preference is prioritized. If no specific chart type is indicated, the method examines the resultant subset of the dataset and analyzes the data types of all columns. Depending on the data type - such as univariate, multivariate, continuous, categorical, or time series - the appropriate chart type is determined. However, the model performs best when the resultant subset has fewer columns, and it supports only ten different types of charts.

However, it has been observed that when the input query contains negative preferences for chart types, the method is unable to provide the correct chart type. Additionally, when the dataset subset contains more than five columns, the method struggles to determine the exact chart type. In such cases, the method defaults to a bar chart if there is one categorical and one numeric column, and to a line chart when all columns are numeric or continuous.

From the resultant subset of the dataset, insights are then generated using the Llama 3 model. The study leverages Llama 3 to utilize its capability for text-to-text generation. The latest Llama 3 model, pre-trained on 15 trillion tokens, demonstrates the model's effectiveness in generating text from the provided input.

From the resultant subset of the dataset and the determined chart type, the model generates charts according to the user's requirements. Utilizing matplotlib and seaborn, the model can produce ten different chart types. Evaluation of the Text2Insights end-to-end model indicates its ability to handle both simple and complex natural language queries effectively. However, limitations include its dependency on a fixed set of chart types and optimal performance with smaller dataset subsets. Overall, Text2Insights represents a promising approach to facilitating data exploration and visualization through natural language interfaces.

Additionally, this study developed a question-answering model to provide natural language responses for user queries based on contextual historical data. Leveraging the BERT model, the model underwent fine-tuning on IPL datasets, optimizing its performance through hyperparameter tuning. Evaluation on the test dataset revealed an accuracy of 88%, precision of 72%, recall of 69%, and an F1-score of 69%.

Lastly, this study developed a predictive model to forecast IPL match outcomes based on natural language queries and contextual historical data. Leveraging the BERT classification model, the predictive model underwent fine-tuning on IPL datasets, optimizing its performance through hyperparameter tuning. Two types of predictive models were developed: binary classification and multi-class classification, both utilizing the same dataset.

The binary classification model predicts match outcomes based on two labels: Team 1 and Team 2. Evaluation on the test dataset revealed that accuracy, precision, recall, and F1-score were approximately 70%. However, accurately identifying Team 1 and Team 2 from the input text is crucial. To address this, the study developed a multi-class classification model with different labels for each team.

Evaluation of the multi-class classification model on the test dataset indicated that accuracy, precision, recall, and F1-score were close to 50%, attributed to the increased number of labels. Nevertheless, results demonstrate that the output labels remain consistent with those present in the input text, indicating the model's fidelity to the provided information.

Also, from the data analysis, it has been found that the nature of IPL is highly competitive, with all teams having around a 50% chance of winning each match. Therefore, the achieved results indicate promising performance.

In conclusion, this study presents, Text2Insights, designed to convert natural language queries into insightful visualizations, enhancing data visualization capabilities. Leveraging a multi-model architecture and techniques such as BERT classification, Text2Insights demonstrates proficiency in generating SQL queries, predicting chart types, and providing valuable insights from resultant dataset. Despite challenges such as negative chart preferences and complex dataset structures, the model shows robustness in handling various queries, although it performs optimally with resultant dataset with fewer columns and a fixed set of chart types.

Furthermore, the study extends its contributions by developing question-answering model and predictive models for IPL match dataset. The binary classification model predicts match results based on Team 1 and Team 2 labels, achieving commendable accuracy, precision, recall, and F1-score. However, accurately identifying teams from the input text proves crucial, leading to the development of a multi-class classification model with individual labels for each team. Although this model yields slightly lower performance due to the increased number of labels, it ensures consistency by only providing output labels present in the input text.

Overall, Text2Insights, question-answering and the predictive models represent significant advancements in natural language processing and data analysis, offering valuable tools for researchers, analysts, and decision-makers to extract insights and make informed decisions from textual data and historical records.

## 6.3 Contributions

This study contributes significantly to the field of data analysis and visualization by introducing Text2Insights, a novel approach that converts natural language queries into insightful visualizations. By leveraging a multi-model architecture and advanced models like Lallam3, Text2Insights demonstrates proficiency in generating SQL queries, predicting chart types, and providing valuable insights from resultant dataset.

Furthermore, the development of question-answering and predictive models for IPL match extends the study's contributions, providing tools for analyzing existing data for past and future outcomes from the contextual natural language. Through rigorous evaluation and experimentation, this research enhances the understanding of natural language interfaces, data analysis and visualization techniques, and predictive modelling, offering practical solutions for data visualization and decision-making.

## 6.4 Future Recommendations

For future researchers, it is recommended to focus on enhancing the flexibility of the Text2Insights model to provide a broader spectrum of chart types and adapt to any kind of dataset structures. This could involve exploring alternative architectures or integrating additional data preprocessing techniques to accommodate diverse user preferences and demands.

Additionally, there is potential to enhance Text2Insights by accommodating multiple datasets with diverse schemas and incorporating join-related queries to provide more comprehensive results for users. By integrating data from multiple interrelated datasets, Text2Insights can offer insights that span across various domains. Moreover, considering that the model currently supports only the English language, a similar approach can be extended to other languages to enable multilingual data visualization. It allows users from diverse linguistic backgrounds to leverage its capabilities for data analysis and visualization effectively.

Integration of real-time data streams into the Text2Insights framework presents a promising way for future research, enabling dynamic updates and on-the-fly analysis. This would enhance the utility of Text2Insights in evolving environments such as financial markets or social media analytics, empowering users to make informed decisions based on the most current information available.

Moreover, refining the question-answering and predictive models could significantly benefit from incorporating additional features or refining the labeling process to enhance accuracy and generalization across various scenarios.

Furthermore, exploring the application of predictive models in domain-specific contexts beyond IPL match predictions holds significant potential. For instance, these techniques could be applied in healthcare analytics for predicting patient outcomes, financial forecasting for market trend analysis, or sentiment analysis in social media data. By using into these domain-specific applications, future research can unlock new ways for leveraging natural language processing, data visualization, analysis, and predictive modeling techniques to address real-world challenges and drive innovation across diverse domains.

# Appendix A – Research Plan

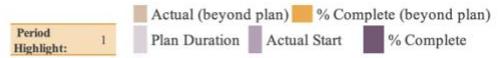

**Text2Insight: Research Plan**

*Each Period is equal to one week.*

Period Highlight: 1

Legend: Actual (beyond plan) | % Complete (beyond plan) | Plan Duration | Actual Start | % Complete

| ACTIVITY | PLAN START | PLAN DURATION | ACTUAL START | ACTUAL DURATION | PERCENT COMPLETE |
|---|---|---|---|---|---|
| Literature Review | 1 | 4 | 1 | 3 | 100 |
| Write first draft for Research Paper (Chapter 1-3) | 1 | 2 | 1 | 3 | 100 |
| Investigate and Analyse BERT, NER and Random Forest | 3 | 1 | 4 | 2 | 100 |
| Design Text2Insight: Multi Model Architecture | 4 | 1 | 4 | 2 | 100 |
| Develop Text2Insight: Multi Model Architecture | 5 | 1 | 6 | 1 | 100 |
| Evaluate Text2Insight: Multi Model Architecture | 5 | 1 | 6 | 1 | 100 |
| Write second draft for Research Paper (Chapter 4-6) | 6 | 2 | 7 | 2 | 100 |
| New Literature Review | 8 | 3 | 9 | 1 | 100 |
| Fine-Tune the architecture from the feedback of Interim | 10 | 1 | 10 | 1 | 100 |
| Evalute the updated model | 10 | 1 | 11 | 1 | 100 |
| Final Research Paper Writing | 11 | 4 | 12 | 1 | 100 |
| Video Presentation | 15 | 1 | 13 | 1 | 100 |
| PPT Documentation | 15 | 1 | 15 | 1 | 100 |

# Appendix B – Research Proposal

**Abstract**

The increasing need for dynamic and user-centric data visualization solutions is evident in various domain including healthcare, finance, research and other. Traditional data visualization systems, while valuable, often fall short in meeting user expectations as they are static and predefined by the system, not aligning with the unique requirements of individual users.

This research introduces Text2Insight, a novel solution designed to address this limitation by delivering data visualizations tailored to users' specific needs. The proposed approach employs a multi-model architecture that takes two key inputs from users: a CSV file containing the data for visualization and natural language text specifying the user's visualization requirements.

The methodology involves an initial data analysis of the input CSV file to extract relevant information such as shape, columns, and related values. Subsequently, a pre-trained BERT model is employed to convert the user-provided natural language text into an SQL query. This generated query is refined with the provided CSV data to formulate an accurate SQL query using NER model, ensuring precise results. The system then utilizes a pre-trained random forest classifier model to determine the most fitting chart type based on the inputted text. Ultimately, the approach generates visually informative charts as the output for user-friendly data visualization.

To validate the efficacy of the proposed approach, a comprehensive evaluation will be conducted at multiple levels. The performance of each inner model will be assessed individually, followed by an overall evaluation of the complete model. This rigorous evaluation by using precision, recall, F1-score and ROC, AUC curve, aims to determine the viability and effectiveness of Text2Insight as a promising solution for transforming natural language text into dynamic and user-specific data visualizations.

**Table of Contents**

**List of Figures**

**List of Tables**

# List of Abbreviations

AUC………………… Area Under the Curve
BERT………………. Bidirectional Encoder Representations from Transformers
CSV…………………. Comma Separated Values
GPT…………………. Generative Pre-trained Transformer
JSON………………... JavaScript Object Notation
LLM………………… Large Language Model
LSTM………………. Long Short-Term Memory
NER…………………. Named Entity Recognition
NLP…………………. Natural Language Processing.
OCR………………… Optical Character Recognition
POS…………………. Parts of Speech
RDBMS……………... Relational Database Management System
ROC…………………. Receiver Operating Characteristic
SQL…………………. Structured Query Language
SVM………………… Support Vector Machine

# 1. Background

In today's world, we're flooded with data from all directions, making it essential to find ways to make sense of it all. Data visualization steps in as a handy tool that transforms complex information into easy-to-understand pictures like charts or graphs. Imagine trying to read a giant spreadsheet versus seeing a clear graph - that's the power of data visualization. It's not just about making things look pretty, it helps us spot trends, connections, and important details in a way our brains can grasp effortlessly.

In recent years, the intersection of natural language processing (NLP) and data visualization has emerged as a dynamic field with the potential to revolutionize information communication and comprehension. One notable application within this domain is the generation charts from textual input, fostering a more dynamic and user-friendly presentation of data. The synthesis of text and charts holds promise across diverse domains, including business analytics, scientific research, healthcare, research, and journalism.

# 2. Problem Statement and Related Work

## 2.1   Problem Statement

The current landscape of data visualization systems is characterized by a static nature, wherein visualizations are generated based on predefined data and chart types. This rigidity becomes a significant drawback when user requirements deviate from the predetermined visualization techniques, creating difficulties in obtaining custom visual representations from the existing systems. The problem at hand is the lack of adaptability and flexibility in current data visualization tools, hindering their ability to cater effectively to varying user needs and preferences.

This research aims to address this limitation by exploring solutions that promote dynamic data visualization, ensuring a more responsive and user-centric approach for data visualization from natural language text.

## 2.2 Related Work

Several research efforts have delved into the domain of automated chart generation from natural language text.

A notable contribution in this field is the Text2Chart (Rashid et al., 2021) system, a multi-staged chart generator in the realm of natural language processing. This system excels in transforming analytical text into visual representations by adeptly identifying axis elements, mapping x-entities with y-entities, and generating appropriate chart types. This study has explored innovative approaches to tackle the challenges associated with automated chart generation. One approach involved integrating BERT with Long LSTM architectures. This integration aimed to enhance language understanding and improve contextual relevance in the conversion process. Additionally, machine learning techniques, including the application of the Random Forest classifier, were employed to discern patterns in textual data.

However, the task of mapping x and y entities is noted to be highly imbalanced, resulting in a bias towards negative mappings in the model. Furthermore, the probability distribution in the Random Forest Classifier has been noted to surpass that of the Support Vector Machine (SVM).

Another research for chart generation have been made with the introduction of ChartLlama (Han et al., 2023), a multimodal large language model. The model is specifically designed to comprehend and generate charts, leveraging a diverse dataset tailored for this purpose. A multimodal LLM has been trained on the dataset for image-text pair, encompassing a broader spectrum of chart types and associated tasks. A noteworthy aspect of ChartLlama's development is its integration with GPT-4, a powerful language model, for the interpretation and narration of chart content. This approach enhances the model's capacity for holistic understanding, providing a comprehensive analysis of visual data.

However, it is important to acknowledge a limitation in the vision encoder of ChartLlama – the absence of multilingual Optical Character Recognition (OCR) capability. This signifies a potential avenue for future research and development to augment the model's language understanding across diverse linguistic contexts.

Another notable research is Chat2VIS (Maddigan and Susnjak, 2023) for Natural Language to Visualization (NL2VIS) solutions, with a particular focus on addressing challenges in generating data visualizations. Chat2VIS leverages the functionalities of pre-trained, large-scale language models like GPT-3, Codex, and ChatGPT, demonstrate a robust integration of natural language processing with visualization generation. Also, Chat2VIS has ability to effectively interpret and respond to diverse, multilingual user queries.

Despite its advancements, Chat2VIS acknowledges and attempts to tackle specific challenges encountered in NL2VIS systems. One notable challenge addressed by Chat2VIS is the handling of situations where multiple x-axis values share the same y-axis value, posing potential complications in visual representation. Moreover, Chat2VIS confronts issues related to missing data values, which are represented as empty strings, potentially leading to null values in the visual output. The system recognizes the significance of addressing this concern to ensure the accuracy and completeness of the generated visualizations. Another aspect considered by Chat2VIS is the potential mismatch in data points when grouping results by category. This issue is crucial in maintaining the coherence and relevance of visual representations, and Chat2VIS makes strides in mitigating such discrepancies.

Another research for text to graph generation is CycleGT (Guo et al., 2020), that addresses the challenge of training on non-parallel graph and text data. CycleGT adopts an iterative back-translation strategy between graphs and text, aiming to bridge the gap between these modalities without the need for labelled data. One key aspect of CycleGT's contribution lies in its ability to achieve performance levels comparable to supervised models, particularly demonstrated on the challenging WebNLG datasets. This speaks to the effectiveness of the unsupervised training paradigm employed by CycleGT in capturing complex relationships between graphs and textual representations.

However, it is important to note that the model faces challenges stemming from the non-differentiability introduced by the discrete decoding of text. This characteristic poses inherent difficulties in the optimization process, requiring careful consideration in the training pipeline. Furthermore, CycleGT's exclusive reliance on cycle training and inductive biases is a distinctive feature. This approach suggests a deliberate design choice, emphasizing the model's capacity to learn and generalize from the underlying data distribution. Understanding the implications and trade-offs associated with this design decision is crucial for contextualizing CycleGT within the broader landscape of unsupervised text-to-graph generation methods.

Another research is ChartQA (Masry et al., 2022), this research introduces two transformer-based models specifically designed for the generation of charts and question, answers from charts. A novel pipeline approach is employed, integrating visual features with extracted data from charts to enhance overall performance.

It is essential to note that, despite their capabilities, the output of these models is not infallible, and there exists the possibility of inaccuracies in their predictions. Furthermore, a noteworthy consideration is the potential misuse of these models, leading to the dissemination of misleading information about the content depicted in charts. This aspect underscores the importance of responsible and ethical deployment of such models to mitigate the risk of misinformation and maintain the integrity of visual data interpretation.

Another research NL4DV (Narechania et al., 2021), which is focused on the intersection of natural language processing and data visualization. NL4DV serves as a powerful tool for facilitating natural language-driven data visualization by proficiently interpreting natural language queries and subsequently generating JSON-based analytic specifications. Within the generated JSON object, NL4DV encapsulates essential elements, including data attributes, analytic tasks, and Vega-Lite specifications.

**3. Research Questions**

The following research questions are proposed for the study, as highlighted as follows:

4. How can the implementation of user-friendly data visualization be effectively achieved?
5. How can Natural Language Processing (NLP) techniques be applied to enhance user-friendly data visualization?

## 4. Aim and Objectives

This research aims to propose a conceptual framework for the generation of charts dedicated to data visualization derived from natural language inputs, with a specific emphasis on refining the accuracy of data utilized in the analysis process. This study endeavors to augment the efficiency of data visualization by facilitating the seamless conversion of textual descriptions into visually elucidating charts.

The research aim is divided into objectives, which are as follows:

- To obtain all necessary information from the input CSV.
- To convert natural language text into SQL statements is the central focus of this academic investigation.
- To refine generated SQL statements into valid SQL with the help of input CSV's headers.
- To define what kind of chart will be preferred to user from user's input text.
- To evaluate the performance of model using accuracy of generated chart from natural language input.

## 5. Significance of the Study

The contemporary landscape of data visualization mandates a nuanced and adaptable approach to cater to the diverse analytical needs across various industries. Presently, the provision of data insights is predominantly characterized by pre-defined charts and static analyses, limiting user flexibility. Organizations tend to offer predetermined visualizations exclusively on predefined datasets. Consequently, users encountering scenarios necessitating alternative analyses on diverse inputs encounter substantial challenges.

The significance of this research lies in its potential to address this inherent limitation. By developing a model facilitating the transformation of natural language input, into personalized and context-specific data visualizations, this study seeks to empower users with the capability to conduct data visualization according to their distinct preferences. This novel approach aims to enhance the adaptability and user-friendliness of data visualization, contributing to a more dynamic and intuitive exploration of diverse analytical scenarios.

## 6. Scope of the Study

This research paper is delimited to the analysis and visualization of data within the context of cricket dataset, primarily due to time constraints. The chosen scope ensures focused and in-depth exploration of the proposed methodology within the specified timeframe. The dataset used for experimentation and validation is exclusively related to cricket.

Furthermore, the research predominantly concentrates on the English language, considering the challenges and resource-intensive nature of training models in multiple languages. Given the complexity and time requirements associated with multilingual model training, the decision to narrow the scope to English facilitates a more efficient and targeted investigation.

However, it is important to note that the designed model, once established and validated on the cricket dataset in English, holds the potential for broader applicability. Future endeavors can extend the use of the developed approach to diverse datasets beyond cricket and languages other than English. This adaptability allows for the scalability and versatility of the proposed model, opening avenues for cross-domain and multilingual applications beyond the initial cricket-focused and English-centric scope.

## 7. Research Methodology

The research methodology employed in this study encompasses several crucial processes. These include the analysis of data from input CSV, the transformation of natural language input text into SQL queries, refinement of the SQL query using the header information from the input CSV, identification of chart types based on either the input text or the generated output of the SQL query, followed by the visualization of the chart, and ultimately, the assessment of the machine learning model.

## 7.1 Methodology Flowchart

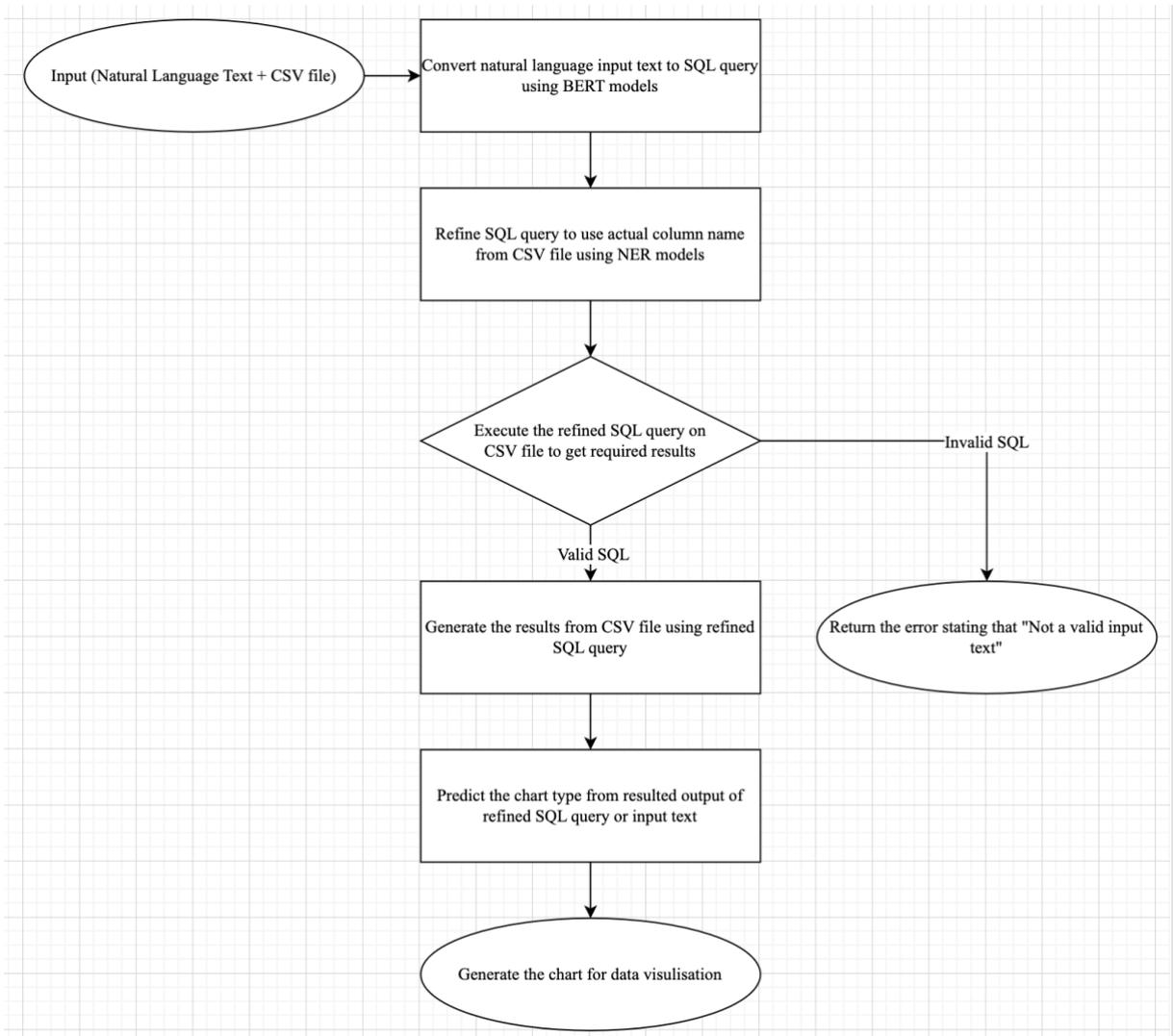

Figure 7.1: Multi-Model Architecture for Data Visualization

## 7.2 Comprehensive Data Analysis of Input CSV

The proposed methodology is designed to accept two inputs: firstly, a CSV file designated by the user for data visualization, and secondly, natural language text specifying the criteria for the desired data visualization. In the initial phase of CSV data analysis, upon receiving the CSV file, the model initiates an exhaustive examination, extracting pertinent details such as the total number of columns, rows, numeric and string columns, column names, primary key columns, and other relevant attributes associated with the CSV file. These values are systematically stored utilizing the Pandas and NumPy libraries in Python. Subsequently, the CSV is transformed into a relational database management system (RDMS) using the SQLite library in Python.

The shape of the dataset, indicating the number of rows and columns, can be obtained using libraries like pandas in Python. Numeric columns can be identified by performing statistical analyses to assess the distribution and characteristics of numerical data. Extracting column names provides insight into the variables present in the dataset. Identifying the primary key involves examining unique identifiers that uniquely identify each record in the dataset. Lastly, string columns can be identified by assessing the data type of each column and isolating those containing text or categorical information. Utilizing these techniques enables a comprehensive analysis of CSV data.

In this research, Cricket dataset is used. Upon conducting data analysis on Bowling_ODI.csv file, it's observed that output of this step will have information like dataset comprising 2583 rows and 13 columns. These columns include information such as the player's name, the number of matches and innings played, along with various statistical data. These extracted insights serve as crucial input for the subsequent stages of our multi-model architecture.

## 7.3 Transformation: From Natural Language Text to SQL Query

The subsequent phase in the proposed methodology involves the transformation of the second input, consisting of natural language text requesting the desired outcome, into an SQL query utilizing a transformer-based model architecture such as BERT.

This reason behind conversion of input text to SQL is imperative due to the critical nature of ensuring accuracy in the results generated by the proposed model. Given the potential impact of these results on business decision-making based on charts, the model is designed to either furnish accurate outputs or prompt the user to refine their query, thereby ensuring the provision of valid input and preventing the delivery of inaccurate results.

Pre-trained BERT models will be employed to convert natural language text into SQL queries by leveraging their ability to capture contextualized semantic representations. Initially trained on a massive corpus of diverse text, pre-trained BERT models inherently grasp the contextual nuances and relationships between words.

The output of this stage is a SQL query generated from the user's second input, which is a natural language query. For example, if the user inputs "help me with the top 10 players with the highest wickets" the query text is fed into a pre-trained BERT model. The model outputs a SQL query, such as "SELECT player_name, wickets FROM players ORDER BY wickets DESC LIMIT 10".

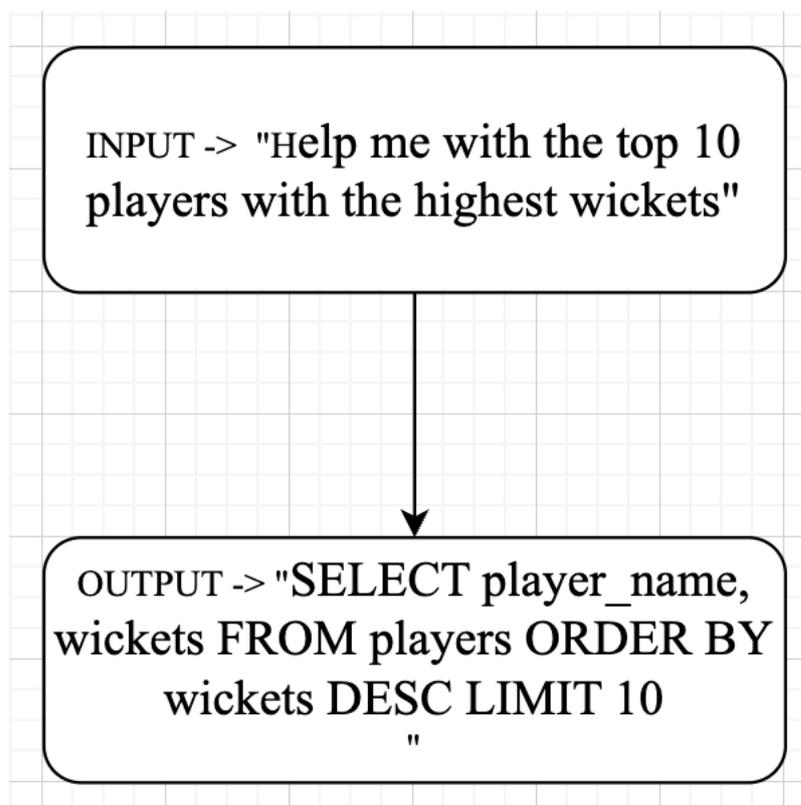

Figure 7.3: Transformation from text to SQL

## 7.4 Fine-Tuning the SQL Query

The subsequent stage in the proposed study involves the refinement of the generated SQL query, utilizing insights derived from the initial step of data analysis applied to the input CSV file. This refinement process incorporates the application of Spacy's Named Entity Recognition (NER) model and Part-of-Speech (POS) tagging.

The rationale behind fine-tuning the SQL query lies in its direct applicability to the converted CSV file turned database, enabling the provision of results. In the event of an invalid query, SQLite would raise an error, prompting the proposed methodology to guide the user towards updating their input text requirements for accuracy and coherence.

By utilizing Spacy's NER capabilities, proposed model can process a given text and extract the relevant keywords, which can help to fine-tune the resulted SQL query.

The output from preceding stages serves as the input for this step, where it undergoes refinement. Previous stages yield CSV column names and a SQL query. In this step, the SQL query is refined. For instance, the initial SQL query may employ keywords such as 'player_name,' 'wickets,' and 'players' to execute the query on a CSV file. However, these cannot be used directly as column names in the CSV file may differ. Consequently, the input SQL query is refined to the following form as "SELECT Player as Player_Name, Wkts as Wickets FROM sql_df_from_csv_file_ ORDER BY Wkts DESC LIMIT 10".

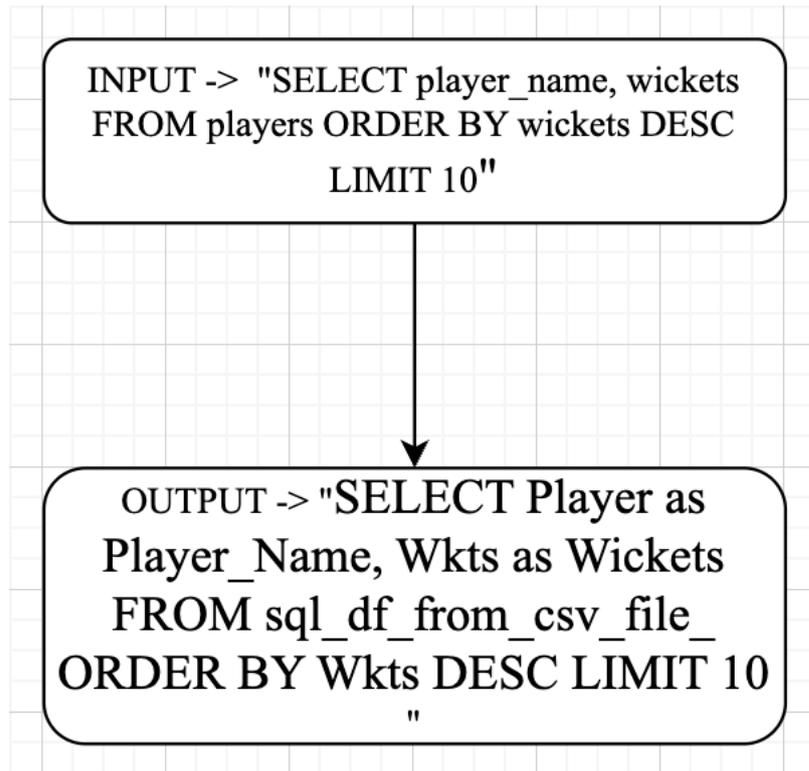

Figure 7.4: Refined SQL Output

## 7.5 Data Subset Generation

In the subsequent phase of the proposed methodology, the refined query is executed on the input CSV file to yield results, forming a distinct subset of the original dataset. This step facilitates the extraction of targeted information tailored to the user's specified criteria.

The outcome of this step will be a subset of the CSV, obtained by executing the refined SQL query received from the preceding step on the CSV file. The resulting subset appears as follows:

Table 7.5: Output of SQL query

| Player Name | Wickets |
|---|---|
| M Muralitharan (Asia/ICC/SL) | 534 |
| Wasim Akram (PAK) | 502 |
| Waqar Younis (PAK) | 416 |
| WPUJC Vaas (Asia/SL) | 400 |
| Shahid Afridi (Asia/ICC/PAK) | 395 |
| SM Pollock (Afr/ICC/SA) | 393 |
| GD McGrath (AUS/ICC) | 381 |
| B Lee (AUS) | 380 |
| SL Malinga (SL) | 338 |
| A Kumble (Asia/INDIA) | 337 |
| ST Jayasuriya (Asia/SL) | 323 |

## 7.6  Chart Type Prediction

In the subsequent phase of the proposed methodology, the focus shifts to predicting the appropriate chart type based on the resulting subset of the dataset obtained from prior steps or through user-specified input text.

Initially, the model checks for any explicit chart preferences conveyed in the user's input text, assigning priority to such articulated preferences. Should the user omit specific chart requirements, the proposed model leverages characteristics of the derived dataset subset. This includes considerations such as the number of numeric and string columns, as well as the presence of specific named columns. For instance, if the subset comprises two numeric columns and one string column, the model strategically determines the most fitting chart type. Furthermore, the model provides insights into the optimal placement of columns on the x-axis and y-axis, guided by the unique attributes of the dataset subset.

The predictive capabilities of the proposed model enhance its adaptability, ensuring that chart types align with user expectations or data characteristics. To achieve this, the proposed methodology will employ a pre-trained random forest classifier. By utilizing this pre-trained model, the methodology gains the ability to predict specific chart types based on given input parameters.

Random Forest Classifier models can be effectively used to predict specific chart types based on input values. The output of this stage designates the chart type. Upon examining the Table 7.6, it becomes evident that a bar chart is more suitable, given its incorporation of one numeric and one string value. Consequently, the Random Forest classifier is anticipated to predict the chart type as a bar chart in a similar fashion.

## 7.7 Chart Generation

In the subsequent phase of the proposed methodology, the generation of charts for user-oriented data visualization is undertaken, leveraging the outputs obtained from preceding steps. This process is accomplished through the utilization of Python's Matplotlib and Seaborn libraries. The integration of these libraries facilitates the creation of visually informative charts, aligning with user-specified criteria and preferences as determined in earlier stages of the methodology. The utilization of established Python libraries ensures a robust and standardized approach to chart generation, enhancing the overall coherence and reliability of the data visualization outcomes within the proposed framework.

Matplotlib and Seaborn are powerful Python libraries widely employed for creating diverse and visually appealing charts and plots. Upon acquiring the complete output from the preceding stages, the chart can be generated using Matplotlib or Seaborn. The resulting chart will manifest as illustrated below for the provided example input text.

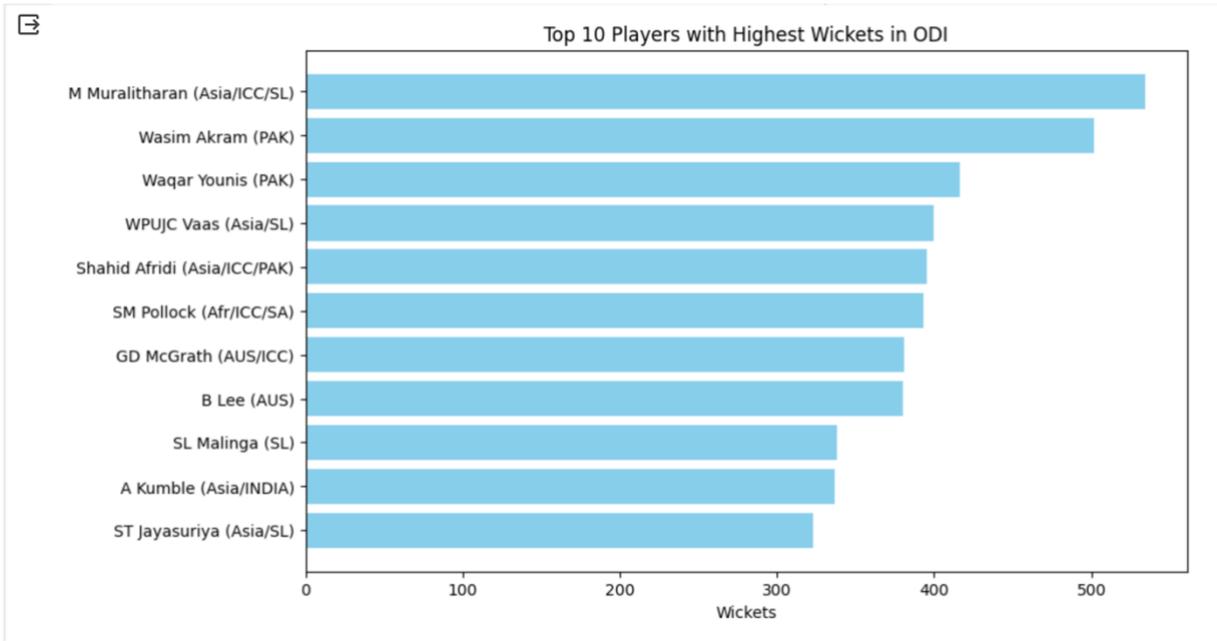

Figure 7.7: Generated Output

## 7.8 Model Evaluation

The concluding step in the proposed study involves the comprehensive evaluation of the employed machine learning models. As delineated in the proposed methodology, various types of models are employed within the study framework. Each of these models undergoes a meticulous evaluation process, scrutinizing diverse parameters to gauge their performance and efficacy. This systematic evaluation ensures a rigorous assessment of the models' capabilities and informs the study's findings with a nuanced understanding of their respective strengths and limitations.

Initially, the BERT model is employed to translate natural language text into SQL queries, a process subjected to evaluation using the BLEU score. The adaptation of BLEU proves instrumental in assessing the quality of SQL query generation, as it involves comparing the n-grams of the generated query with those of the reference query. This methodological approach enhances precision in evaluating the effectiveness of the BERT model's ability to seamlessly transform natural language inputs into coherent and accurate SQL queries.

BLEU score can be calculated with the following formula:

$$BLEU = BP \times \exp(1/n \sum_{i=1}^{n} \log(\text{precision}_i))$$

Subsequently, a second model is employed to refine SQL queries through the application of Spacy's Named Entity Recognition (NER) model. Evaluation of this process will utilize metrics such as recall, precision, and F1-score. These metrics collectively offer a thorough evaluation of the model's proficiency in accurately identifying and classifying entities within the refined SQL queries. This systematic evaluation framework enhances the understanding of the precision and recall capabilities, contributing to a nuanced assessment of the model's performance in fine-tuning SQL queries.

The final model proposed is the Random Forest Classifier, employed for predicting chart types. The evaluation of this model involves the use of a confusion matrix, from which precision and recall metrics will be computed. This systematic evaluation framework provides a detailed analysis of the model's performance in predicting chart types, enhancing our understanding of its precision and recall capabilities within the proposed study.

In addition to the internal evaluation of individual models, the study will conduct a holistic assessment of the entire model. This comprehensive evaluation process involves initially gathering sample inputs from multiple users for cricket dataset. Subsequently, a small dataset is curated using these inputs, and charts are manually generated for each sample input text. This sample dataset, not previously encountered by the model during its training phase, will then be inputted into the developed model for evaluation. The model's performance will be assessed using Receiver Operating Characteristic (ROC) and Area Under the Curve (AUC) metrics, providing a robust evaluation of its predictive capabilities on unseen data.

Confusion Metrix can be used to calculate all the mentioned evaluation metric as follows:

TP = Number of True Positive Predictions
TN = Number of True Negative Predictions
FP = Number of False Positive Predictions
FN = Number of False Negative Predictions

Precision = (TP)/(TP+FP)

Recall = (TP)/(TP+FN)

F1-Score = (2*Precision*Recall)/(Precision+Recall)

## 8. Requirements Resources

### 8.1 Software Requirements

- Google Colab - Cloud-based Jupyter notebook platform for collaborative coding
- Pandas - Python library for data manipulation and analysis
- NumPy - Python library for numerical operations
- SQLite – Python library for compact and serverless database
- Matplotlib - Python library for creating visualizations
- Seaborn - Python library for statistical data visualization
- Pytorch - Library for the development and training of neural networks
- Scikit-learn – ML library in Python to design classification models
- TenserFlow – ML framework for development of deep learning models
- BERT – Pre-trained natural language processing model
- Spacy – Library for understanding textual data

### 8.2 Hardware Requirements

- Laptop with at least 8GB RAM

## 9. Research Plan

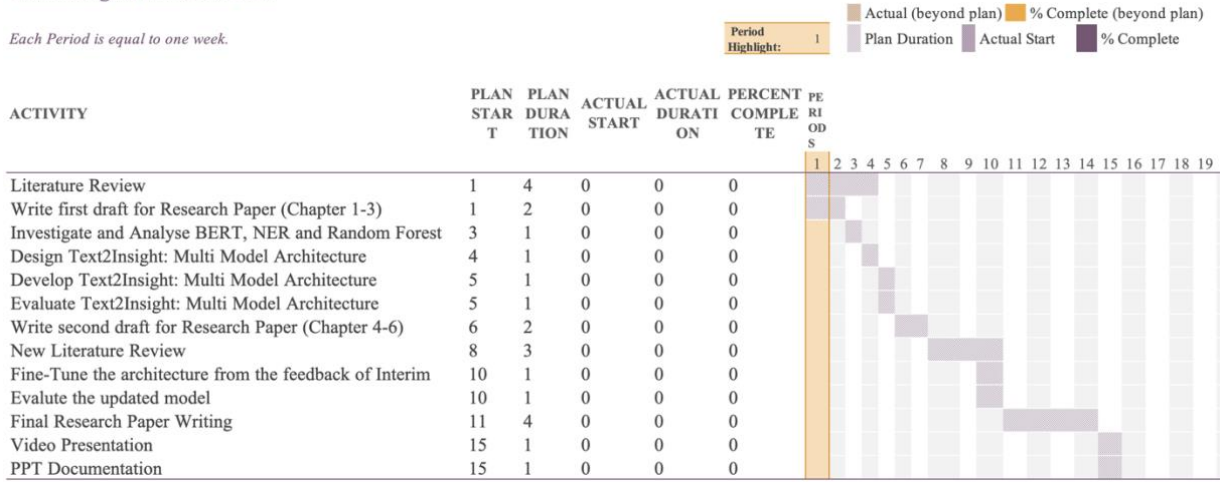